%% file: gan_experiments.tex
\pdfoutput=1
\documentclass[conference]{IEEEtran}
\usepackage{cite}
\usepackage{amsmath,amssymb,amsfonts}
\usepackage{graphicx}
\usepackage{textcomp}
\usepackage{hyperref}
\usepackage{url}
\usepackage{graphicx}
\usepackage{amsfonts,amssymb}
\usepackage{epsfig} 
\usepackage{booktabs}
\usepackage{cleveref}
\usepackage{amsmath}\usepackage{url}
\usepackage{amsfonts,amssymb}
\usepackage{amsthm}
\usepackage{epsfig} 
\usepackage{multirow}
\usepackage{booktabs}
\usepackage{colortbl}
\usepackage[noend]{algpseudocode}
\usepackage{bbm}
\usepackage{lipsum} 
\usepackage{ulem}
\usepackage{lmodern}
\usepackage[T1]{fontenc}
\usepackage[utf8]{inputenc}

\usepackage[caption=true, font=normalsize, labelfont=sf, textfont=sf]{subfig}

\newcommand{\E}{\mathbb{E}}

\newcommand{\eg}{{e.g.,~}}

\newcommand{\ie}{{i.e.,~}}

\newtheorem{prp}{Proposition}

\def\BibTeX{{\rm B\kern-.05em{\sc i\kern-.025em b}\kern-.08em
    T\kern-.1667em\lower.7ex\hbox{E}\kern-.125emX}}
\begin{document}

\title{Lessons Learned from the Training of GANs on Artificial Datasets}
\author{\IEEEauthorblockN{Shichang  Tang\textsuperscript{1, 2, 3}}
\IEEEauthorblockA{\textsuperscript{1}School of Information Science \& Technology, ShanghaiTech University}
\IEEEauthorblockA{\textsuperscript{2}Shanghai Institute of Microsystem and Information Technology, Chinese Academy of Sciences}
\IEEEauthorblockA{\textsuperscript{3}University of Chinese Academy of Sciences}
}

\maketitle


\begin{abstract}
Generative Adversarial Networks (GANs) have made great progress in synthesizing realistic images in recent years. However, they are often trained on image datasets with either too few samples or too many classes belonging to different data distributions. Consequently, GANs are prone to underfitting or overfitting, making the analysis of them difficult and constrained. 
Therefore, in order to conduct a thorough study on GANs while obviating unnecessary interferences introduced by the datasets, we train them on artificial datasets where there are infinitely many samples and the real data distributions are simple, high-dimensional and have structured manifolds.
Moreover, the generators are designed such that optimal sets of parameters exist.
Empirically, we find that under various distance measures, the generator fails to learn such parameters with the GAN training procedure.
We also find that training mixtures of GANs leads to more performance gain compared to increasing the network depth or width when the model complexity is high enough. Our experimental results demonstrate that a mixture of generators can discover different modes or different classes automatically in an unsupervised setting, which we attribute to the distribution of the generation and discrimination tasks across multiple generators and discriminators. 
As an example of the generalizability of our conclusions to realistic datasets, we train a mixture of GANs on the CIFAR-10 dataset and our method significantly outperforms the state-of-the-art in terms of popular metrics, \ie Inception Score (IS) and Fréchet Inception Distance (FID).
\end{abstract}

\input{1intro}

\input{2related}
\input{3model}
\input{4exp}

\section{Conclusions and Future Work}
In this work, we explore different distance measures to investigate whether GAN training succeeds in learning the distribution. Our empirical results show that even when the distances between $\mathbb{P}_g$ and $\mathbb{P}_{r}$ are short, there exists a simple classifier with a model complexity similar to the discriminator that can distinguish $\mathbb{P}_g$ and $\mathbb{P}_{r}$ accurately. It suggests that $\mathbb{P}_g$ and $\mathbb{P}_{r}$ have little non-negligible overlapping manifold\cite{towardsprincipled}.
Empirically, we also find that even when an optimal set of generator parameters exists, GAN training fails to find it. Therefore, it remains an open question whether GANs should be replaced by non-adversarial generative models (\eg \cite{glow,VQVAE2,VAE,NCSN,residualflow}).

In our experiments on the synthetic datasets, increasing the size of the training set can improve the performance of GANs, even when it is already very large. On the other hand, a small training set can negatively affect GANs. Therefore, current datasets might not be large enough to make GANs learn the real data distribution or even result in overfitting.

Our experimental results show that training a mixture of GANs is more beneficial than simply increasing the complexity of standalone networks (that are sufficiently complex) for modeling multi-modal data. It is an interesting topic to devise different ways to combine models in the mixtures. It is also promising to measure and promote the diversity of the ensemble\cite{krogh1995neural,diversity} of discriminators. 

Finally, while current state-of-the-art GAN models such as BigGAN\cite{biggan}, CR-BigGAN\cite{cr-biggan} and LOGAN\cite{logan} use a number of TPU cores that is the same as the height or width of the images, we are not able to conduct such large-scale experiments. But we believe that with more computing power, a large mixture of GANs can be trained on datasets such as ImageNet $128\times 128$ and improve current state-of-the-arts.

\appendices
\section{Proof of Proposition 1}\label{proof1}
\textbf{Proposition 1.}
\textit{
	Let $J$ be a deterministic classifier for samples from two distributions $\mathbb{P}_{r}$ and $\mathbb{P}_{g}$ with equal prior probabilities. Let $\delta(\mathbb{P}_{r},\mathbb{P}_{g})$ be the total variation distance between $\mathbb{P}_{r}$ and $\mathbb{P}_{g}$, then
}
\begin{align}
\delta(\mathbb{P}_{r},\mathbb{P}_{g})\ge 2\E[J_{acc}]-1
\end{align}
\begin{proof}
	Without loss of generality, assume that the label $y$ of a sample point $x$ equals 1 if $x$ is from $\mathbb{P}_{r}$ and 0 otherwise. Let $J_{opt}$ be an optimal classifier with the highest expected accuracy. For all $x$ such that $p_r(x)+p_g(x)>0$, we have
	$P(y=1|x)=\frac{p_r(x)}{p_r(x)+p_g(x)}$ and $P(y=0|x)=\frac{p_g(x)}{p_r(x)+p_g(x)}$.
	Then there exists a $J_{opt}$ that predicts $J(x)=1$ if $p_r(x)\ge p_g(x)$ and $J(x)=0$ if $p_r(x)< p_g(x)$. Therefore,
	\begin{align}
	\E[J_{acc}]
	&\le \E[J_{{opt}_{acc}}]\\
	&=\frac{1}{2}\int p_r(x)\mathbbm{1}(J(x)=1)dx\\
	&+\frac{1}{2}\int p_g(x)\mathbbm{1}(J(x)=0)dx\\
	&=\frac{1}{2}\int_{p_r(x)\ge p_g(x)} p_r(x)dx\\
	&+\frac{1}{2}\int_{p_r(x)< p_g(x)} p_g(x)dx\\
	&=\frac{1}{2} \int max\{p_r(x),p_g(x)\}dx\\
	&=\frac{1}{2}\Big(\int max\{p_r(x),p_g(x)\}dx-1+1\Big)\\
	&=\frac{1}{2}\Big(\int max\{p_r(x),p_g(x)\}dx\\
	&-\frac{1}{2}\int (p_r(x)+p_g(x))dx+1\Big)\\
	&=\frac{1}{4}\int \Big(max\{p_r(x),p_g(x)\}\\
	&-min\{p_r(x),p_g(x)\}\Big)dx+\frac{1}{2}\\
	&=\frac{1}{4}\int |p_r(x)-p_g(x)|dx+\frac{1}{2}\\
	&=\frac{1}{2}\delta(\mathbb{P}_{r},\mathbb{P}_{g})+\frac{1}{2}.
	\end{align}
	It follows that $\delta(\mathbb{P}_{r},\mathbb{P}_{g})\ge 2\E[J_{acc}]-1$.
\end{proof}

\section{Network architectures}\label{architectures}
In Table \ref{CIFAR_G} and \ref{CIFAR_D}, we list the network architectures we use for CIFAR-10.
\begin{table}[!h]
	\caption{MHingeGAN Generator for $32\times 32$ images}\label{CIFAR_G}
	\centering
	\footnotesize
	\setlength{\tabcolsep}{4pt}
	\renewcommand{\arraystretch}{1.2}
	\centering
	\begin{tabular}[h]{p{0.45\linewidth}p{0.45\linewidth}}
		\toprule
		\midrule
		{Block or layer(s)} &{Output shape}\\
		\midrule
		$z\in \mathbb{R}^{80}\sim \mathcal{N}(0,I)$  &$20+20+20+20$    \\
		\midrule 
		Embed($y$)$\in\mathbb{R}^{128}$  & {$128$}\\
        \midrule
        Linear$(20+128\rightarrow 4 \times 4 \times 256)$ & $4 \times 4 \times 256$\\
        \midrule
        Resblock up $256\rightarrow 256$ & $8 \times 8 \times 256$\\
        \midrule
        Resblock up $256\rightarrow 256$ & $16 \times 16 \times 256$\\
        \midrule
        Resblock up $256\rightarrow 256$ & $32 \times 32 \times 256$\\
       \midrule
        BN, ReLU, conv $3\times3$, $tanh$ & $32 \times 32 \times 3$\\
		\midrule
		\bottomrule
	\end{tabular}
\end{table}

\begin{table}[!h]
	\caption{MHingeGAN Discriminator for $32\times 32$ images}\label{CIFAR_D}
	\centering
	\footnotesize
	\setlength{\tabcolsep}{4pt}
	\renewcommand{\arraystretch}{1.2}
	\centering
	\begin{tabular}[h]{p{0.45\linewidth}p{0.45\linewidth}}
		\toprule
		\midrule
		{Block or layer(s)} &{Output shape}\\
		\midrule
		input & $32 \times 32 \times 3$\\
		\midrule
        Resblock down $3\rightarrow 256$ & $16 \times 16 \times 256$\\
        \midrule
        Resblock down $256\rightarrow 256$ & $8 \times 8 \times 256$\\
        \midrule
        Resblock  $256\rightarrow 256$ & $8 \times 8 \times 256$\\
        \midrule
        Resblock  $256\rightarrow 256$ & $8 \times 8 \times 256$\\
       \midrule
        ReLU, global sum pooling & $256$\\
        \midrule
        Linear$(256\rightarrow K+1)$& $ K+1$\\
		\midrule
		\bottomrule
	\end{tabular}
\end{table}

\section*{Acknowledgement}
This work was supported with Cloud TPUs from Google's TensorFlow Research Cloud (TFRC). Special thanks go to Dr. Kun Huang who helped revise an early draft.
\bibliographystyle{ieee_tran}
\bibliography{iclr2017_conference}

~\\
~\\

\end{document}

%% file: 1intro.tex
\section{Introduction}\label{sec:introduction}

The past few years have witnessed the arising popularity of generative models. As can be seen, image processing (\eg image super-resolution and editing) and machine learning (\eg reinforcement learning and semi-supervised learning) tasks are infused strong energy by generative models\cite{goodfellow2016nips}. Typically, a generative model learns a distribution $\mathbb{P}_g$ to approximate the true distribution $\mathbb{P}_r$, given a set of observed samples. 



Generative Adversarial Network \cite{GAN}, with no doubt, is the most prevailing generative model. It is composed of a generator $G$ that maps random noise to synthesized data points, and a discriminator $D$ which aims to tell whether its input comes from the real data distribution $\mathbb{P}_r$ or generative distribution $\mathbb{P}_g$. During training, $D$ and $G$ are updated simultaneously or alternatingly. In a vanilla GAN, $D$ gives an estimate of the Jensen--Shannon divergence between $\mathbb{P}_r$ and $\mathbb{P}_g$ while $G$ tries to minimize it \cite{GAN}.

Unfortunately, the objective of $G$ can get saturated when $\mathbb{P}_g$ and $\mathbb{P}_{r}$ do not have an non-negligible overlapping manifold, causing vanishing gradients to the generator\cite{towardsprincipled}. Let $\mathcal{Z}$ and $\mathcal{X}$ be the domain and codomain of $G$ respectively. $G(\mathcal{Z})$ is contained in a countable union of manifolds of dimension at most $\text{dim}\ \mathcal{Z}$. Then, according to \cite{towardsprincipled}, if the dimension of $\mathcal{Z}$ is less than that of $\mathcal{X}$, $G(\mathcal{Z})$ will be a set of measure $0$ in $\mathcal{X}$, $\mathbb{P}_r$ and $\mathbb{P}_g$ can be distinguished with accuracy $1$ by $D$ and thus no gradient is provided to $G$.
Besides, GANs suffer from \textit{mode collapse}. Mode collapse refers to the phenomenon that the samples of the generator lacks the diversity exhibited in $\mathbb{P}_{r}$. \cite{generalization} prove that the generator can fool the discriminator by generating a limited number of images from the training set. In other cases of mode collapse, the generated samples are even meaningless as $G$ needs only to fool $D$ in the current iteration. When mode collapse happens, the model fails to generate diverse and realistic data.



To cope with these challenges, variants of GAN were proposed (\eg \cite{roth2017stabilizing, lsgan,wgan,improvedwgan,spectralnorm,mescheder2018training}). 
Limited by the fact that these methods are applied to high-dimensional realistic datasets with inadequate samples from each class, the behavior of GANs remains not completely understood. 
Another problem with realistic datasets is that the performance of GANs can degrade simply due to data scarcity or insufficient model complexity \cite{generalization,biggan}. 

Considering that we aim to study the behavior of GANs, conventional image datasets might not be good choices.
Hence we train GANs on artificially constructed datasets (\eg mixtures of Gaussians in high dimensional space), applying neural networks with sufficiently high capacity. In this way, we can avoid the influence of the aforementioned factors and focus on the inherent problems of GAN training.

The contributions of this work can be summarized as follows:
\begin{itemize}
	\item We propose a set of metrics for evaluating GANs trained on the artificial datasets.
	\item We designed controlled experiments where we can adjust the network width/depth, the mixture of networks, and the training set size, and then relate them to the performance of GANs.
	\item Our empirical study suggests that GANs may fail to learn the real data distribution, even if at least one set of optimal parameters exists for the generator by design. 
	\item In terms of model complexity, our experimental result demonstrates that when the networks are already reasonably large, training a mixture of GANs is more beneficial than increasing the complexity of stand-alone networks, as the generation task can be divided by multiple generators and the variance of the discrimination model is reduced when using an ensemble of discriminators. We further validate this conclusion on the CIFAR-10 dataset and achieve the state-of-the-art Inception Score and Fréchet Inception Distance.
\end{itemize}

%% file: 2related.tex

\section{Related Work}
There are attempts to make a GAN converge to an equilibrium \cite{generalization,FID,mescheder2018training}. However, even if a GAN reaches an equilibrium, it might fail to learn the desired real data distribution. To support this conjecture, \cite{dogans} adopt the \textit{Birthday Paradox} to measure the diversity of the generative distribution. They present empirical evidence that $\mathbb{P}_{g}$ has lower support than $\mathbb{P}_{r}$. 
However, it should be noted that this problem also might be due to the dimension of the manifold of the latent distribution being lower than the dimension of the manifold of $\mathbb{P}_{r}$\cite{towardsprincipled}.
In order to rule out this possibility, we set the dimension of $z$ to be no lower than the dimension of $x$ in our experiments on the artificial datasets.
Basically, we share a similar goal with \cite{dogans}, but we conduct experiments on artificial constructed datasets with infinite data samples. 


Consistent with \cite{dogans}, our experiments reveal that even when a GAN converges to a diverse distribution, it still differs from the true distribution. Considering that the \textit{birthday paradox} test in \cite{dogans} is rather restrictive on continuous data, we propose to use some other measures for validating whether GANs can learn the real data distribution.

Recently, large scale GAN training(\eg \cite{biggan,bigbigan,logan,cr-biggan}) has proven effective on the ImageNet \cite{imagenet} dataset. Their superiority over previous models is mainly due to high model complexity and large batch sizes.
While current state-of-the-art GAN models on ImageNet are still subject to model complexity and batch size, our work focus on synthetic datasets that allows the batch size and model complexity to be sufficiently high, which enables us to explore the properties of GANs in ideal cases.

Some previous work has studied the feasibility of using multiple discrimiantors \cite{GMAN}, multiple generators\cite{MADGAN, MGAN}, or both \cite{generalization} to improve the performance of GANs. Our experiments on artificial datasets is based on MIX+GAN \cite{generalization} and we find it beneficial to use multiple generators and discriminators.
Further, our experimental results unveil the relations between the number of generators and discriminators and the performance of GANs. As the computation of MIX+GAN is expensive or even infeasible, we modify it to allow larger mixtures and achieve stat-of-the-art results on CIFAR-10.
In our work, we also explore how factors such as network depth, network width and training set size affect the performance of GANs.

%% file: 3model.tex
\section{Models}


WGAN-GP\cite{improvedwgan} has been gaining popularity (e.g., \cite{PGGAN,stargan,styleGAN}) for its stability, while MIX+GAN\cite{generalization} guarantees the existence of approximate equilibrium using a mixture of generators and discriminators. MIX+GAN is also effective in modeling multi-modal data which is common in realistic datasets. Therefore, we combine WGAN-GP and MIX+GAN for our experiments on the artificial datasets. 

In the following, we will first introduce Wasserstein GAN (WGAN)\cite{wgan} and WGAN-GP\cite{improvedwgan}, then introduce MIX+GAN\cite{generalization}.

\subsection{WGAN-GP}
In a vanilla GAN, the generator tries to minimize the approximate Jensen-Shannon divergence defined by the discriminator. Different from vanilla GAN, the discriminator in WGAN calculates an approximate Wasserstein distance between the real and fake data distributions. The discriminator in WGAN is also referred to as the "critic". We will use both terms interchangably in this paper.  

The minimax game for WGAN is formulated as
\begin{equation}
W({P}_{r},\mathbb{P}_g)=\mathop {\min }\limits_G \mathop {\max }\limits_D 
{E}_{x\sim \mathbb{P}_r}[D(x)]
-\mathbb{E}_{\tilde{x}\sim \mathbb{P}_g}[D(\tilde{x})]
\label{wganobj}
\end{equation}
where $D$ is in the set of all 1-Lipschitz functions and $\mathbb{P}_g$ is the model distribution implicitly defined by  $z \sim p(z)$, $\tilde{x} = G(z)$. Note that Eq. \ref{wganobj} can be reformulated as
\begin{equation}
W(\mathbb{P}_{r},\mathbb{P}_g)\!\!=\!\!
\frac{1}{k}
\{\mathop {\min }\limits_G 
\mathop {\max}\limits_D 
\mathbb{E}_{x\sim \mathbb{P}_r}[D(x)]
-\mathbb{E}_{\tilde{x}\sim \mathbb{P}_g}[D(\tilde{x})]
\label{k_lip}
\}\!\!
\end{equation}
where $D$ is in the set of all k-Lipschitz functions.

WGAN  \cite{wgan} adopts a weight clipping approach to enforce the Lipschitz constraint. However, it can lead to optimization problems and pathological behaviors. To overcome these problems, An improved version of WGAN was proposed in \cite{improvedwgan}, introducing a new objective for the critic:
\begin{align}
&\mathbb{E}_{\tilde{x}\sim \mathbb{P}_g}[D(\tilde{x})]
\!-\!\mathbb{E}_{x\sim \mathbb{P}_r}[D(x)]
\!+\!\lambda \mathbb{E}_{\hat{x} \sim \mathbb{P}_{\hat{x}}}[( \left\|\nabla _{\hat{x}} D(\hat{x})\right \|_2 \!-\!1)^2]\!\!
\label{wgan-gp}
\end{align}
where $\hat{x}$ comes from the distribution $\mathbb{P}_{\hat{x}}$ whose samples are interpolated between samples from $\mathbb{P}_g$ and $\mathbb{P}_{data}$. This choice is based upon the fact that the L2-norm of the gradient of the optimal D is 1 between the manifolds of $\mathbb{P}_g$ and $\mathbb{P}_{data}$ \cite{improvedwgan}.


The last term can be interpreted as a regularizer that forces the gradient between the real and fake datasets to be at a moderate scale, so that $\mathbb{P}_g$ is moved smoothly to the real data distribution.

\subsection{MIX+GAN}
A group of datasets that the GANs in this paper are tasked with is mixtures of Gaussians. A generator can learn any n-dimensional Gaussian distribution with n-dimensional isotropic Gaussian input noise by simply learning an affine transformation. However, this problem becomes less straightforward if $\mathbb{P}_r$ is a mixture of Gaussians. Another problem is that different modes in the dataset can be discontinuous (which is common in realistic datasets) and thus cannot be learned by a continuous generator network, posing another challenge to GAN training. Therefore, we use MIX+GAN \cite{generalization} to model mixtures of Gaussians. In MIX+GAN, there are $n_G$ generators and $n_D$ discriminators. Each $G_i$ and each $D_j$ has a weight $w_{i}$ and $v_{j}$ respectively to indicate their relative importance. The weights are produced by the softmax function on the learnable log-probabilities, therefore $\sum_{i=1}^{n_D}w_{i}=1$ and $\sum_{j=1}^{n_D}v_{j}=1$.
In a MIX+GAN, both players play mixed-strategies: The generators' weighted probability density at point $x$ is
\begin{align}
p_g(x)&=\sum_{i=1}^{n_G}w_ip_{gi}(x),
\end{align}
and the discriminators' weighted output at point $x$ is
\begin{align}
D(x)&=\sum_{j=1}^{n_D}v_{j}D_{j}(x).
\end{align}
To encourage the weights to get close to the discrete uniform distribution, entropy regularization terms, $-\frac{1}{n_D}\sum_{j=1}^{n_D}\log(v_{j})$ and $-\frac{1}{n_G}\sum_{i=1}^{n_G}\log(w_{i})$, are added to the loss of the generators and the loss of discriminators respectively.
Therefore, the overall loss for the discriminators is
\begin{align}\label{D_loss1}
&L_D=\sum_{j=1}^{n_D}\sum_{i=1}^{n_G}v_{j}w_{i}\mathbb{E}_{x\sim \mathbb{P}_{gi}}[L_{D,fake}(D_j(x))]\\
&+\sum_{j=1}^{n_D}v_{j}\mathbb{E}_{{x}\sim \mathbb{P}_r}L_{D,real}(D_j(x))]-\frac{1}{n_D}\sum_{j=1}^{n_D}\log(v_{j})\!\label{D_loss2}
\end{align}
and the  overall loss for the generators is
\begin{align}\label{G_loss}
\!L_G\!\!=\!\!\sum_{j=1}^{n_D}\sum_{i=1}^{n_G}v_{j}w_{i}\mathbb{E}_{x\sim \mathbb{P}_{gi}}[L_{G}(D_j(x))]\!-\!\frac{1}{n_G}\!\sum_{i=1}^{n_G}\log(w_{i})\!\!\!
\end{align}
where $L_{D,real}$, $L_{D,fake}$ and $L_{G}$ are functions of the outputs of the discriminators. For example, in a vanilla GAN, $L_{D,real}(D_j(x))\!=\!-log(D_j(x))$, $L_{D,fake}(D_j(x))\!=\!-log(1-D_j(x))$ and $L_{G}(D_j(x))=-log(D_j(x))$.

%% file: 4exp.tex
\section{Experiments on the artificial datasets}

\subsection{Datasets}
Popular GANs usually focus on learning high-dimensional and complicated datasets (\eg realistic images and natural languages). As a result, the GAN models are sensitive to almost every hyperparameter. Furthermore, these datasets contain either too many categories or too few samples in each category, thus GAN models can easily underfit or overfit \cite{comparison_of_ml_and_gan}, \ie generating samples of low visual quality or encountering mode collapse. In this paper, we conduct experiments on the following simple artificial datasets with infinite samples:
\begin{enumerate}	
	\item \textit{Mixture of Gaussians}. In our experiments, a dataset of a mixture of Gaussians consists of samples from independent high-dimensional Gaussian components with equal prior probabilities. The Gaussian components have their centers lying on axes of a Cartesian coordinate system in 1024-dimensional space. Specifically, the coordinate of the $i$'th center is $e_i=(0,...,0,1,0,...0)$ whose $i$'th entry is 1 and the other entries are 0; the covariance matrices are all $0.09I$. 
	
	\item \textit{Output of a randomly initialized network}. This dataset is from the output of a network $R$ that has the same input noise as a generator. In the case of a single generator $G$, if $R$ and $G$ have the same arhitecture and $G$ has learned the parameters of $R$, then no classifier can distinguish $\mathbb{P}_r$ and $\mathbb{P}_g$.
\end{enumerate}

\subsection{Design of Model Architectures}
In order to compare across different experimental settings, we 
design our model architectures following the rules below to reduce unnecessary interference and maintain simplicity:
\begin{itemize}
	\item All neural networks consist of affine layers and LeakyReLU non-linearities only.
	\item Each hidden affine layer is followed by a LeakyReLU activation layer.
	\item The input dimension and output dimension of each generator is 1024. In addition to having LeakyReLU activations, each generator has no less than 1024 neurons in each of its hidden layers, so that it can be learned to be injective.
	\item If a network has hidden layers, then all of its hidden layers contain the same number of neurons.
	\item Throughout this session, the number of layers refers to the number of hidden affine layers plus one input layer and one output layer, excluding LeakyReLU layers, e.g., a 2-layer network is an affine transformation from the input space to the output space; a 5-layer network has 3 hidden layers.
	\item Unless stated otherwise, each network has 5 layers and has 1024 neurons in each hidden layer.
\end{itemize}

\subsection{Evaluation metrics}
Evaluating different generative models accurately and objectively remains challenging. Currently, there are some reasonable and widely accepted metrics, \ie Turing test, Inception Score \cite{ImprovedTechniquesforGANS,note_on_IS}, Fréchet Inception Distance \cite{FID} and approximate Wasserstein distance \cite{comparison_of_ml_and_gan}. The evaluation metrics we use are detailed as follows:

\subsubsection {Visualization and Turing test} 
\begin{figure}[h]
	\centering
		\includegraphics[width=0.7\linewidth]{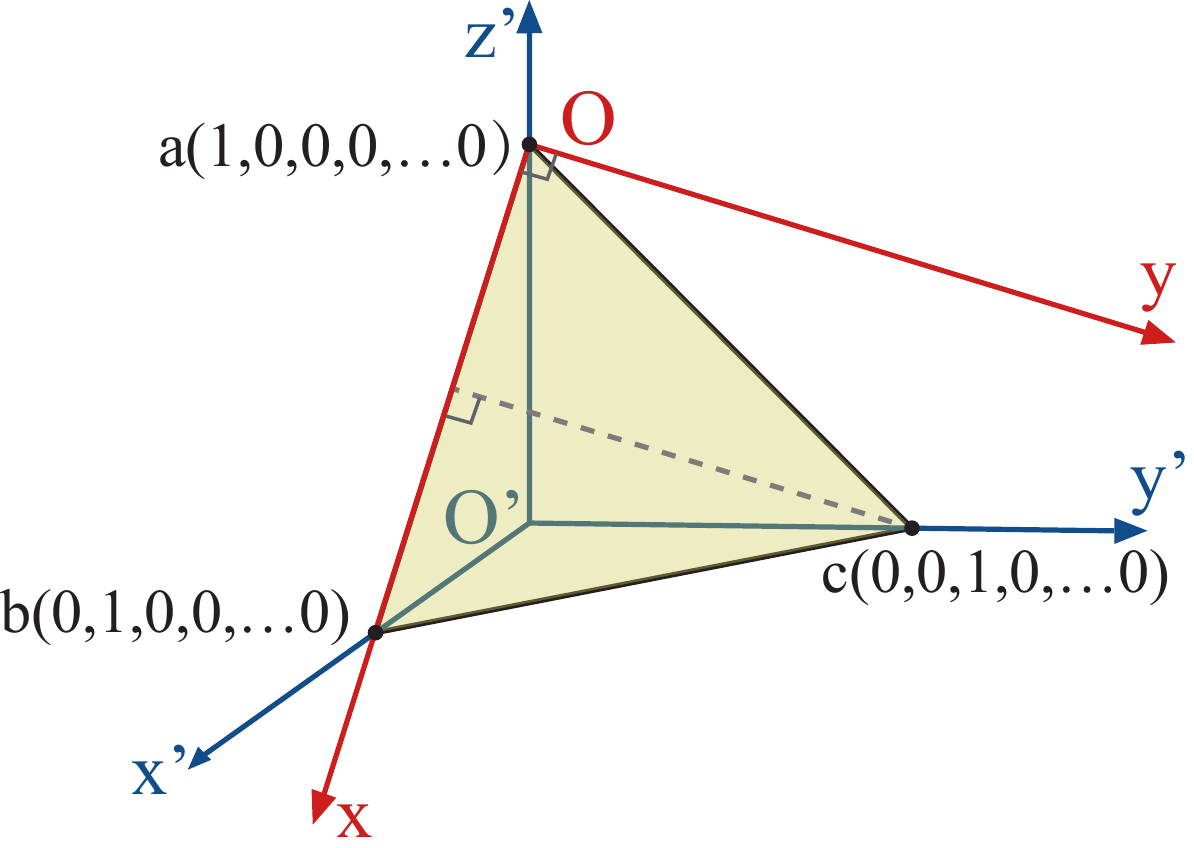}
	\caption{In our projection method, all the data points are projected from the 1024-dimensional space onto the plane that a, b and c lie in.}
	\label{coordinates}
\end{figure}

This is perhaps the simplest way to evaluate a generative model. It is done by using human inspectors to check the quality of (the projection of) the generated data. If a human inspector cannot distinguish whether the generated data are real or fake, then one can conclude that the generative model is very successful. If the inspectors say that samples generated by one model are significantly better that those generated by another model, then it can also be concluded that one model is better than another. On the other hand, if a human inspector cannot tell a significant difference, then one may want to resort to more objective and more accurate metrics. To inspect the generated synthetic high dimensional samples manually, we project the generated samples onto a plane determined by $a=(1, 0, 0, 0, ..., 0)$, $b=(0, 1, 0, 0, ..., 0)$ and $c=(0, 0, 1, 0, ..., 0)$. In the projection plane, the origin is $a$, The x-axis is in the same direction as $\overrightarrow {ab}$, and the y-axis is in a direction that is perpendicular to the x-axis, as is shown in Figure \ref{coordinates}. The projections of sample data can be seen in Figure \ref{comp_quali} and some other figures in this paper, where real data points are indicated by red dots, fake generated data points are indicated by blue dots. Background colors show the contour plot of the output of the discriminator(s): red corresponds to high values while blue corresponds to low values.

\subsubsection {Fréchet Distance} 
\cite{FID} proposed to use the Fréchet Inception Distance (FID) as a metric for evaluating generative models. The Fréchet Distance (FD, also known as the Wasserstein-2 distance) for two Gaussian distributions $N(m_1, C_1)$ and $N(m_2, C_2)$ is given by $\|m_1-m_2\|_2^2+tr(C_1+C_2-2(C_1C_2)^{1/2})$ \cite{FD}. During the computation of FID, images from the real and fake distributions are fed into the Inception model \cite{inception} to get their activations in the last pooling layer. The distributions of the activations are approximately treated as Gaussian so that their means and covariances can be used to compute the FID. In this paper, since we are dealing with artificial data and the Inception model was intended for realistic data, we use the means and covariances of the artificial data directly without passing them through the Inception Network. 50,000 data points are sampled from $\mathbb{P}_{r}$ and $\mathbb{P}_{g}$ respectively for computing the Fréchet Distance.

\subsubsection{Critic output}
As is noted in \cite{wgan}, the loss of the critic provides a meaningful estimate of the Wasserstein distance between $\mathbb{P}_r$ and $\mathbb{P}_g$. We can log the average value of $D(x_r)-D(x_g)$ during each iteration with almost no additional computation cost. If it is positive, then it tells us that $\mathbb{P}_r$ is different from $\mathbb{P}_g$. Moreover, it is an indicator of the training dynamics of WGAN.

\subsubsection {Wasserstein distance} 
The above estimate of the Wasserstein distance can be inaccurate due to adversarial training. Alternatively, one can train an independent critic to approximate the Wasserstein distance after training a GAN \cite{comparison_of_ml_and_gan}. Note that the gradient penalty term might be large than 0 and the critic may be a k-Lipschitz function, thus we normalized the estimated Wasserstein distance using Eq. \ref{k_lip}.

For fair comparison, we train an independent critic with the same architecture across different experiments. Specifically, it has 5 layers and 1024 neurons in each hidden layer.
In our experiments, we estimate the approximate Wasserstein distance $W(\mathbb{P}_{r},\mathbb{P}_{g})$ with 25,600 sample points from $\mathbb{P}_{r}$ and $\mathbb{P}_{g}$ respectively. 

\subsubsection {"Judge" accuracy} 
In all our experiments, an independent classifier called "Judge" is trained to distinguish samples from $\mathbb{P}_g$ and $\mathbb{P}_{r}$. The accuracy of the Judge is an objective metric for evaluating all of our GANs. After the Judge is fully trained, its classification accuracy is expected to range between 0.5 and 1. If the generator(s) has learned the distribution, then the Judge should have an accuracy of around 0.5. Conversely, if the generator(s) produces a distribution different from $\mathbb{P}_{r}$, the Judge is expected to have an accuracy higher than 0.5. 
Following Theorem 2.2 of \cite{towardsprincipled}, given two distributions $\mathbb{P}_g$ and $\mathbb{P}_{r}$ that have support contained in two closed manifolds $M$ and $P$ that don’t perfectly align and don’t have full dimension, and assume that $\mathbb{P}_g$ and $\mathbb{P}_{r}$ are continuous in their respective manifolds, then there exists a perfect classifier that has accuracy 1.

One can show that the expected Judge accuracy is related to the total variation distance:
\begin{prp}
		Let $J$ be a deterministic classifier for samples from two distributions $\mathbb{P}_{r}$ and $\mathbb{P}_{g}$ with equal prior probabilities. Let $\delta(\mathbb{P}_{r},\mathbb{P}_{g})$ be the total variation distance between $\mathbb{P}_{r}$ and $\mathbb{P}_{g}$, then
	\begin{align}
	\delta(\mathbb{P}_{r},\mathbb{P}_{g})\ge 2\E[J_{acc}]-1
	\end{align}
	\label{acc_and_tv}
\end{prp}
The proof of Proposition \ref{acc_and_tv} is provided in Appendix \ref{proof1}.

Proposition \ref{acc_and_tv} is intuitive: If the total variation distance between two distributions is very low, then it is hard for any classifier to tell them apart and the accuracy of a classifier can hardly get above 0.5; if a classifier has an accuracy of 1, then the total variation distance between them is high. One can in turn show that the total variation distance is related to the Kullback–Leibler Divergence \cite{tvtokl}.

For fair comparison, we train an independent Judge with the same architecture across different experiments. Specifically, it has 5 layers and 1024 neurons in each layer.
In our experiments, we estimate $J_{acc}$ with 25,600 sample points from $\mathbb{P}_{r}$ and $\mathbb{P}_{g}$ respectively. 

\subsection{Training}
We follow some experimental setups of \cite{improvedwgan} for toy data: The batch size is 256; there are 100,000 GAN iterations, each of which includes 1 generator update and 5 discriminator updates; After the training of GAN, we train the Judge and the independent critic for another 100,000 iterations respectively. Adam optimizers\cite{adam} are used for optimizing all models.

Our experiments differ from \cite{improvedwgan} in the following ways:
In order to imitate the training of GANs on
high-dimensional realistic data, the data points lie in 1024-dimensional space; motivated by \cite{towardsprincipled}, in order to allow the manifold of $\mathbb{P}_g$ to
have the same dimension as that of $\mathbb{P}_r$, the input noise $z$ follows a 1024-dimensional Gaussian distribution and the activation
layers are chosen to be LeakyReLU layers that are injective;
in all the experiments, $\lambda$ in WGAN-GP is set to $10$ to
improve stability; we adopt a "two time-scale update rule"
(TTUR)\cite{FID}: the learning rate of
the Discriminators(s) is set to $1e-4$ and the learning rate of
the Generator(s) is set to $1e-5$ after some hyperparameter
searching.

During each iteration of the training of GAN, the generator(s) and the discriminator(s) are updated in the following order:
\begin{enumerate}
	\item Draw a batch of real data from the real data distribution.
	\item Each generator generates $n_D$ batch(es) of fake data, which are distributed to the $n_D$ discriminator(s).
	\item Compute the loss for the generator(s) as described in Eq. \ref{G_loss} and take an optimization step on the generator(s).
	\item Compute the loss for the discriminators(s) as described in Eq. \ref{D_loss1}-\ref{D_loss2} and take an optimization step on the discriminator(s).
	\item Repeat step $1)$, $2)$ and $4)$ for another 4 times.
\end{enumerate}

To stabilize GAN training, some tricks were proposed in \cite{GANtips}. However, most of them are not necessary under the WGAN-GP's setting. Besides, we do not incorporate into our models other potentially beneficial techniques such as normalization techniques \cite{batchnorm,layernormalization,weightnormalization,spectralnorm} as they may limit the capacity of the models.

\subsection{Results}
In this part, we will report the experimental results on the artificial datasets.

\subsubsection{Generation of Mixtures of Gaussians}

\begin{figure}[!h]
	\begin{center}
		\subfloat[1G1D]{\includegraphics[width=0.24\linewidth] {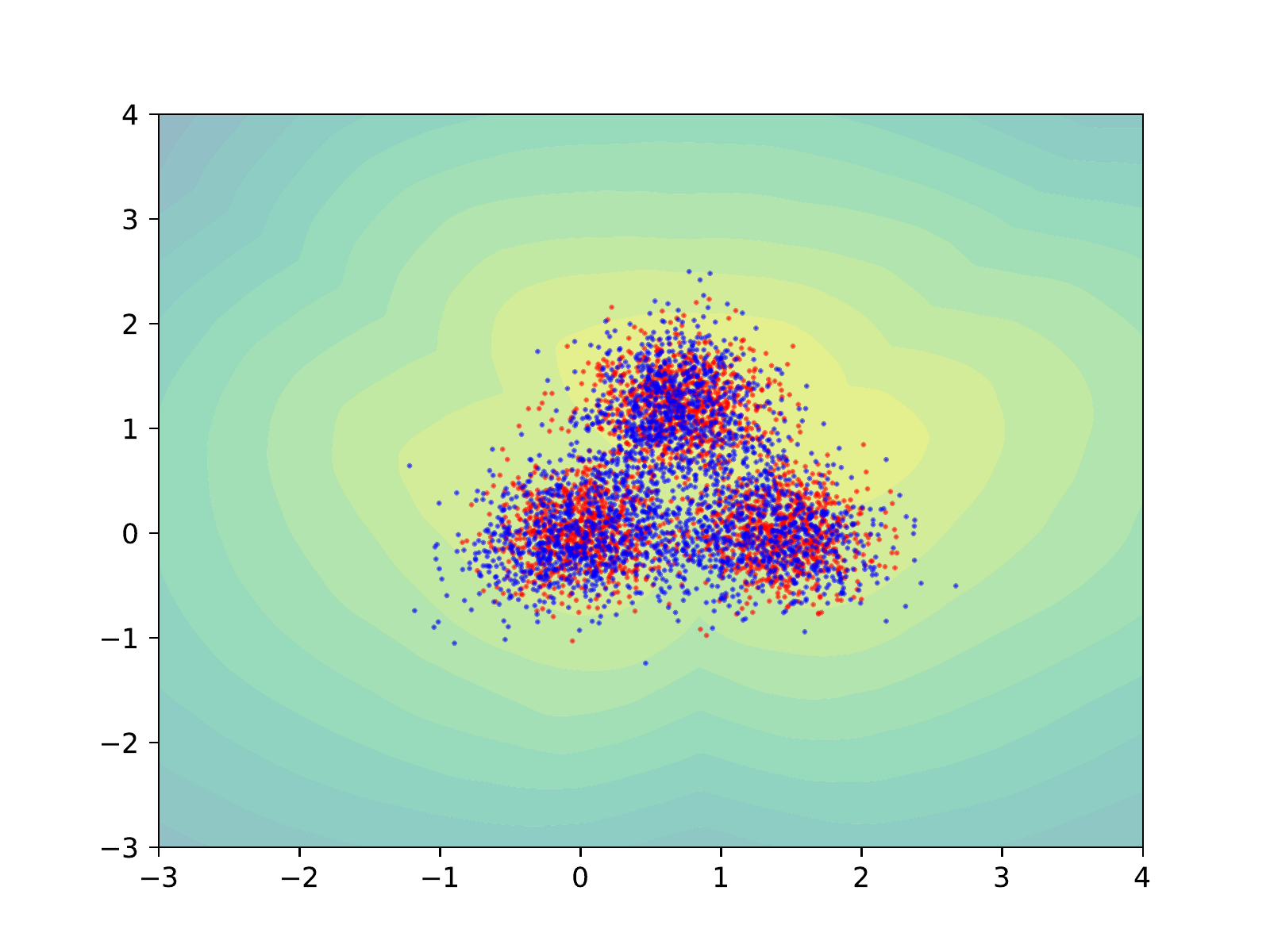}
		}
		\subfloat[3G3D]{\includegraphics[width=0.24\linewidth]{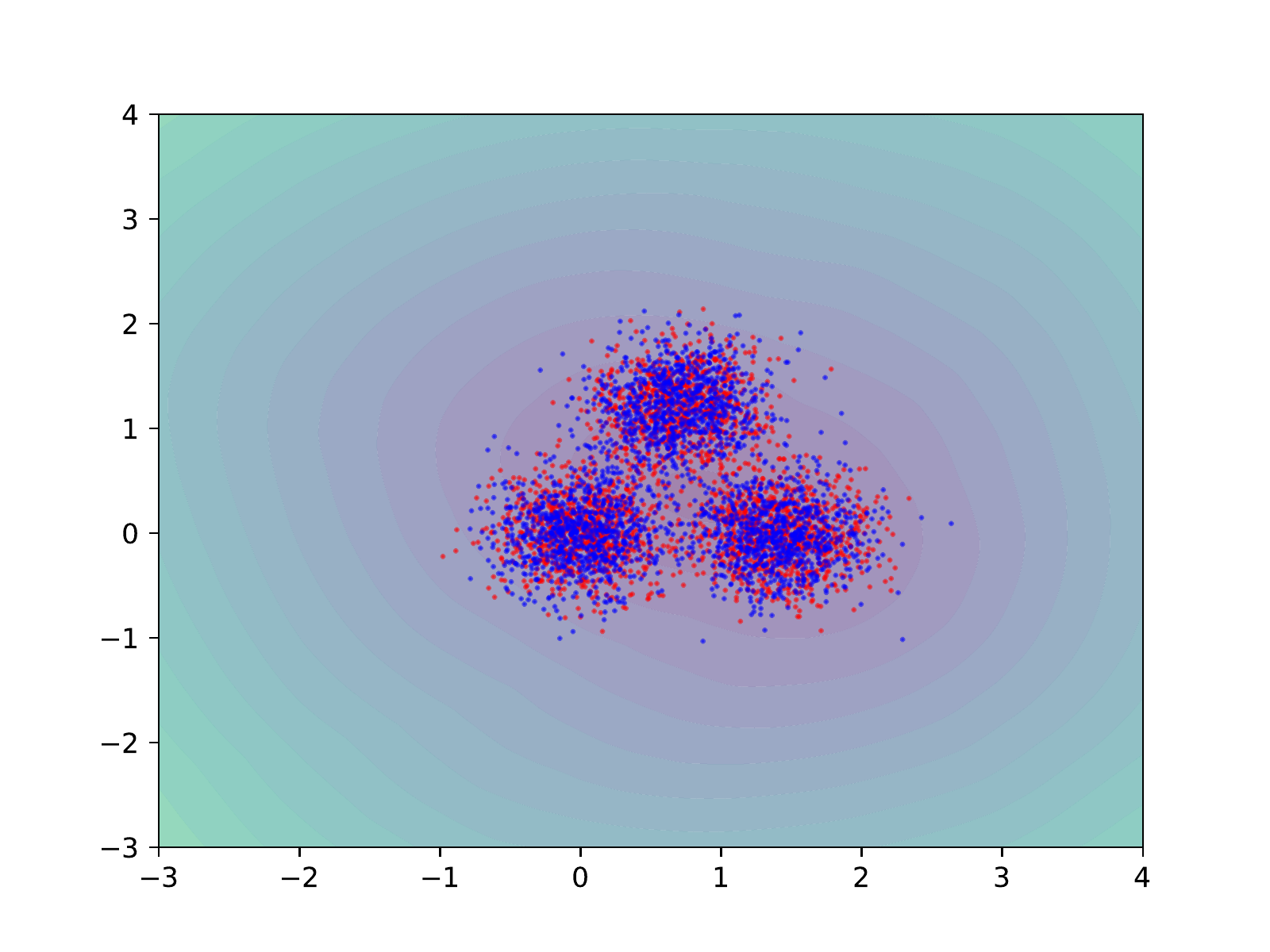}}
		\subfloat[5G5D]{\includegraphics[width=0.24\linewidth]{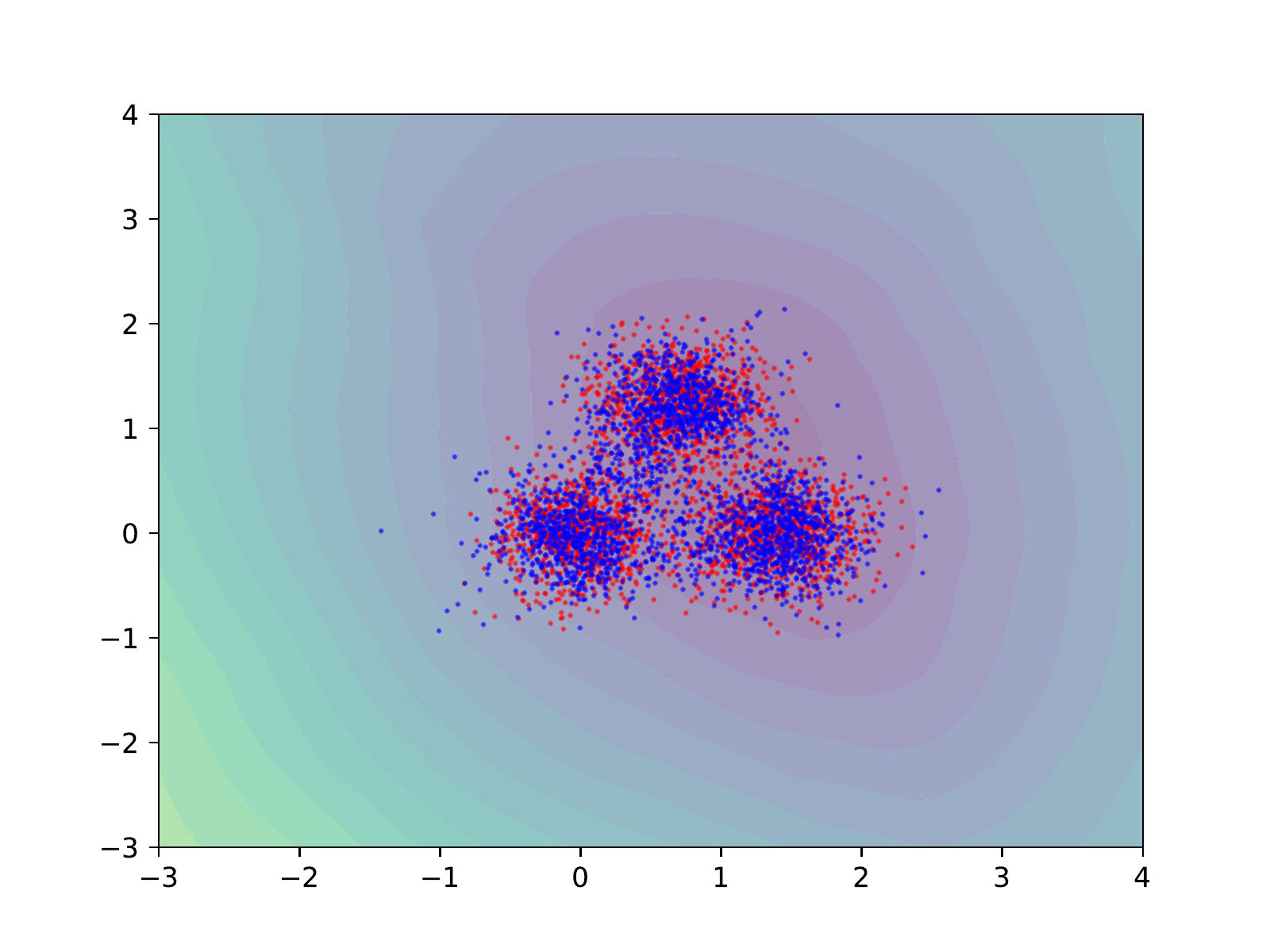}}
		\subfloat[10G10D]{\includegraphics[width=0.24\linewidth]{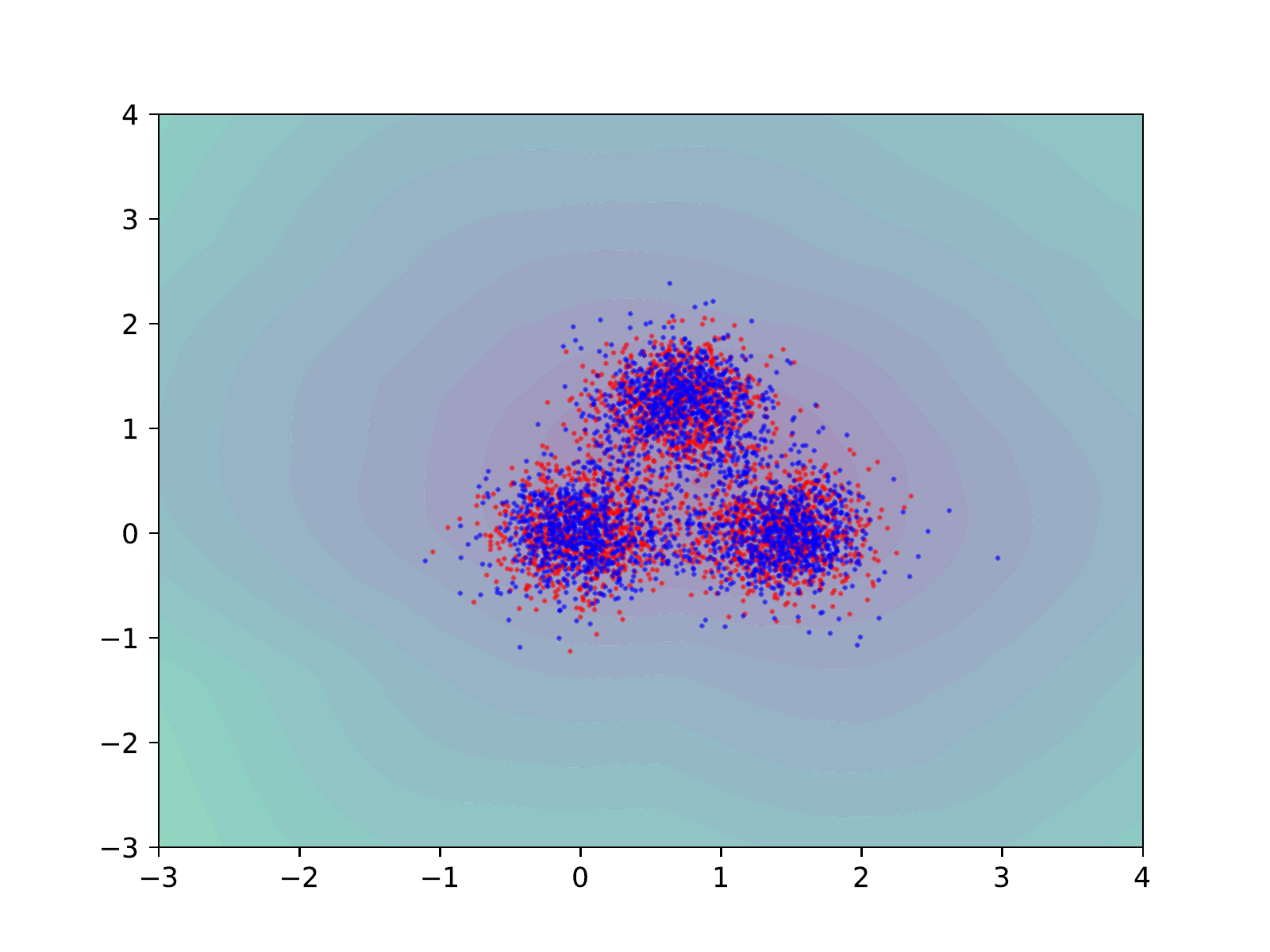}}
	\end{center}
	\caption{Projections of real data (red dots) and samples (blue dots) generated by "MIX+GAN" with different mixtures of models. "$nGmD$" indicates that there are $n$ generators and $m$ discriminators.}
	\label{comp_quali}
\end{figure}

\begin{figure}[!h]
	\begin{center}
		\subfloat[Fréchet distance]{\includegraphics[width=0.45\linewidth]{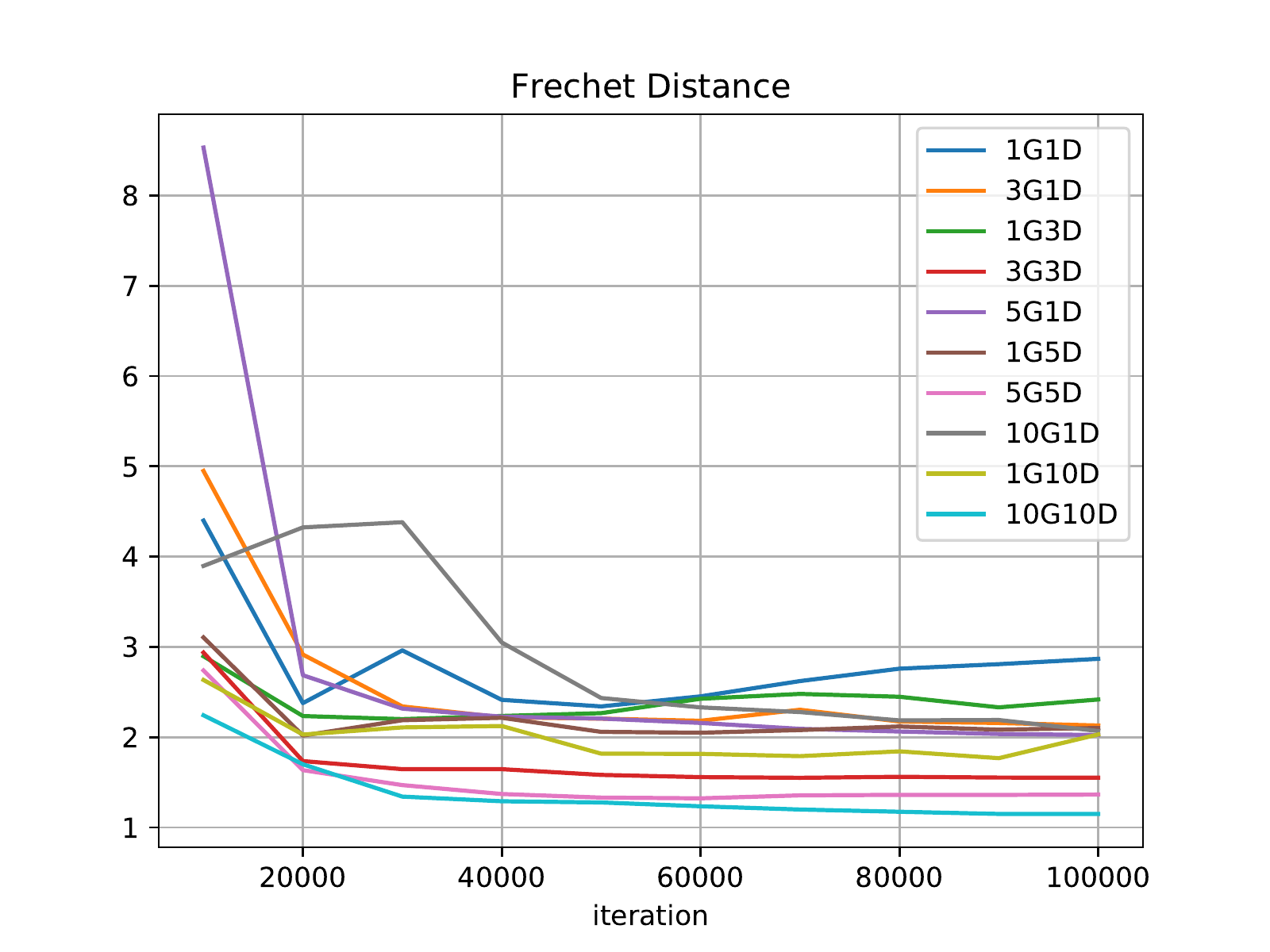}}
		\subfloat[$D(x_r)-D(x_g)$]{\includegraphics[width=0.45\linewidth]{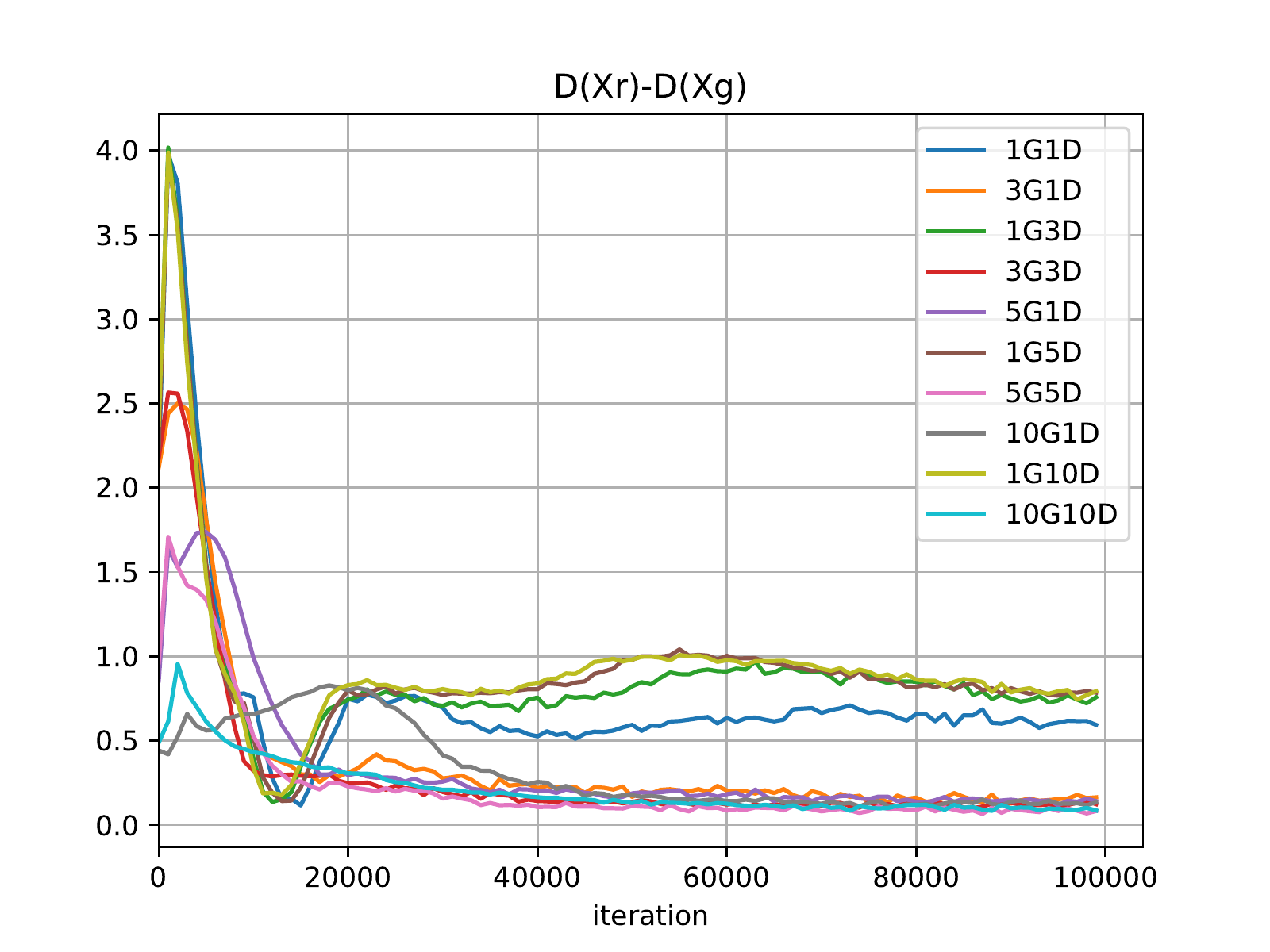}\label{D-D}}\\
		\subfloat[Judge accuracy]{\includegraphics[width=0.45\linewidth] {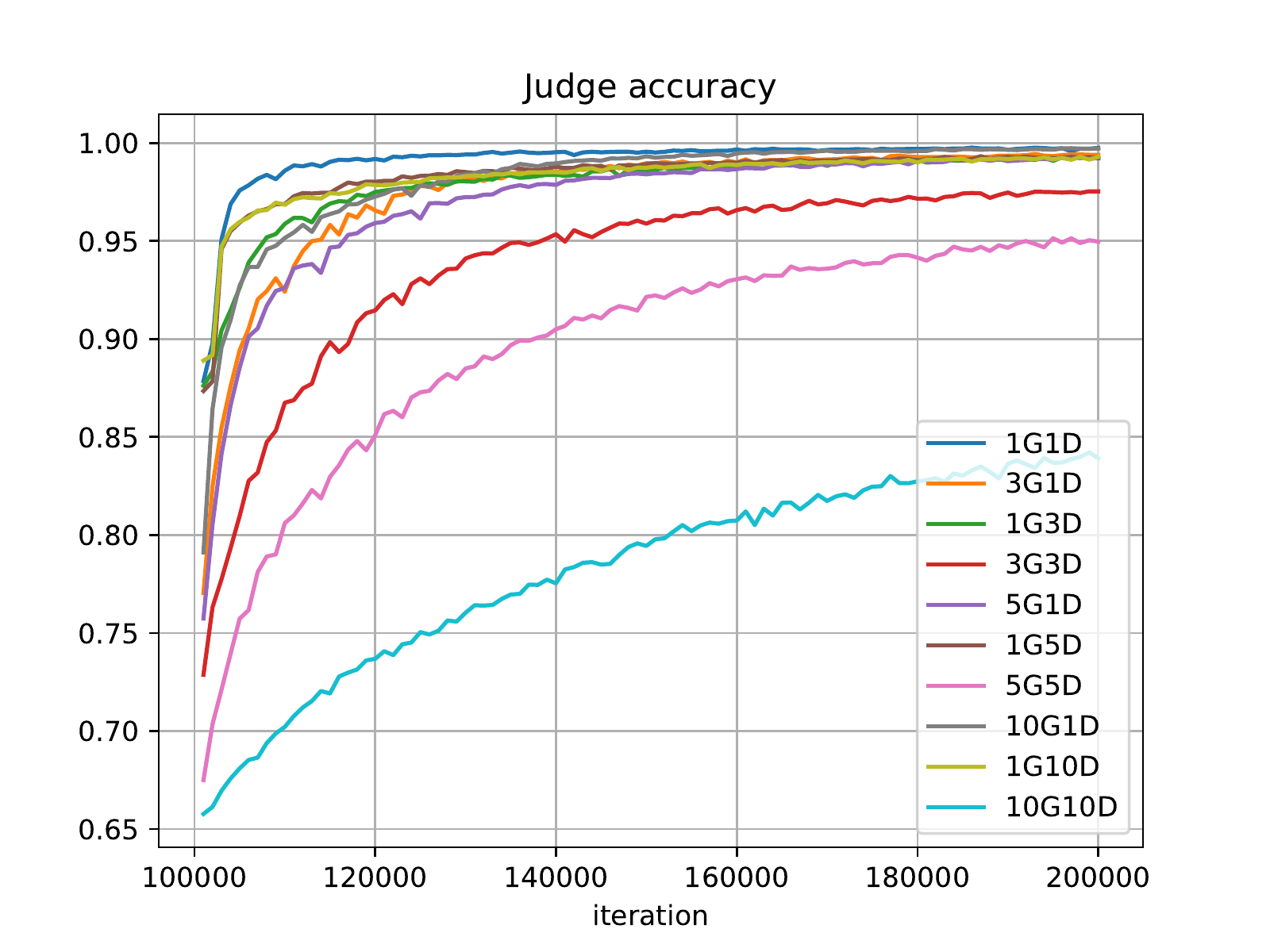}
		}
		\subfloat[Wasserstein distance]{\includegraphics[width=0.45\linewidth]{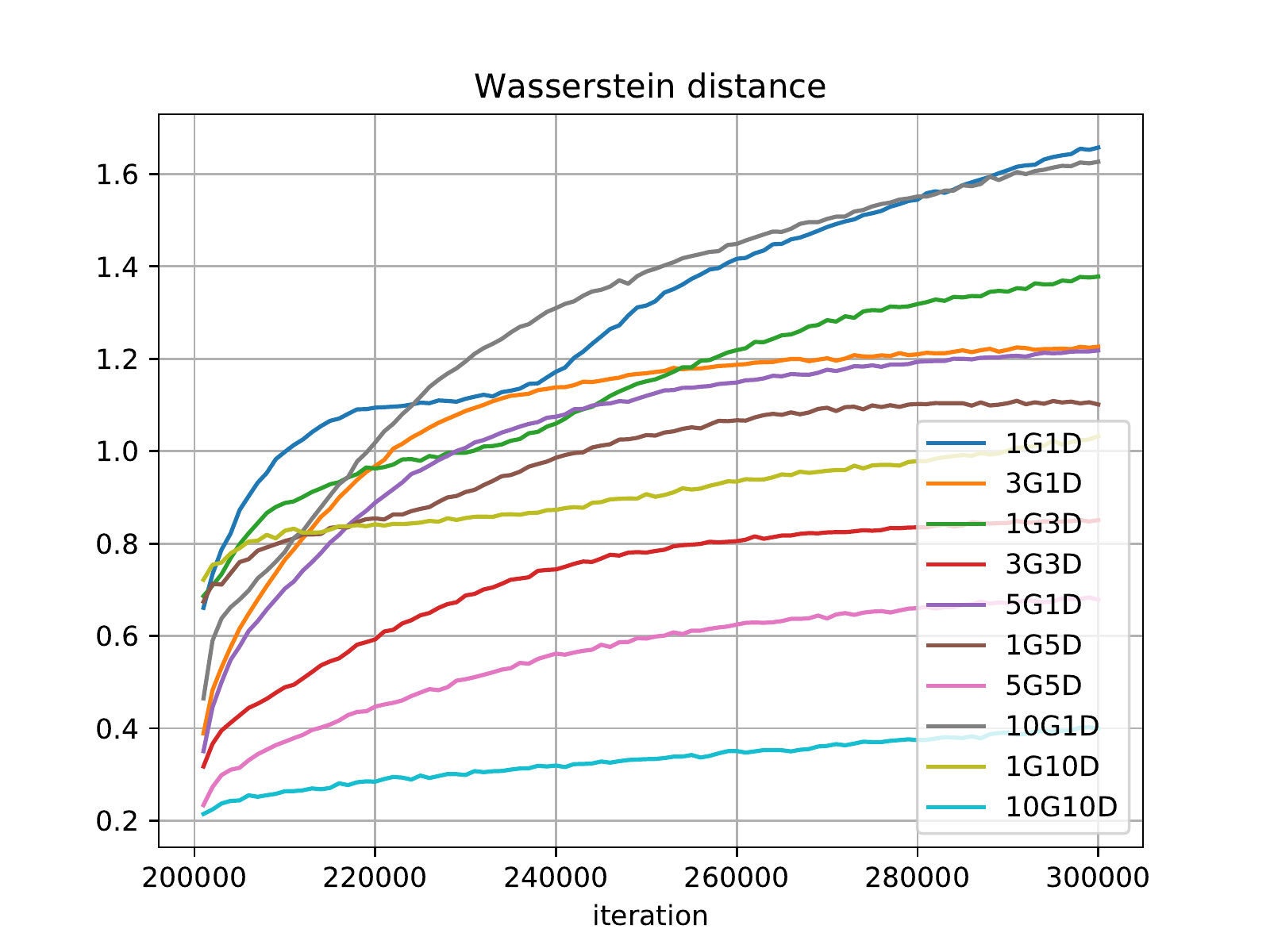}}
	\end{center}
	\caption{Comparisons of MIX+GANs with different components for generating mixtures of 3 Gaussians. We evalate the Fréchet distance during training and train the Judge and the independent critic after the training of WGAN-GP. For all the metrics, lower is better.}
	\label{comp_quant}
\end{figure}

In Figure \ref{comp_quali}, we present the qualitative results on the 3-Gaussians dataset. The projections of real data points and generated data points are indicated by red and blue dots, respectively. The contour plot shows that the output of the critic of WGAN-GP is quite smooth. In Figure \ref{comp_quant}, we compare MIX+GANs with different combinations of mixtures quantitatively using the aforementioned metrics. 

In each experiment, MIX+GAN successfully learns a 3-modal mixture, but it differs from the real distribution. Nevertheless, the generative distribution $\mathbb{P}_g$ is closer to $\mathbb{P}_r$ with larger mixtures. 

There are at least two ways for the generator(s) to win the game. For one thing, Corollary 3.2 in \cite{generalization} states that low-capacity discriminators are unable detect lack of diversity, thus the generator(s) can memorize a large quantity of training data to win the game. For another, since the generator(s) can be learned to be injective with all the hidden dimensions being 1024, which is the same as the input dimension and the output dimension, a mixture of 3 generators can learn 3 individual Gaussian components perfectly. But in GAN training, the generator(s) does not win, as Figure \ref{D-D} shows that the discriminator(s) can distinguish the real and generative data distributions.

\begin{figure}[!h]
	\begin{center}
		\subfloat[$G_1$]{\includegraphics[width=0.3\linewidth]{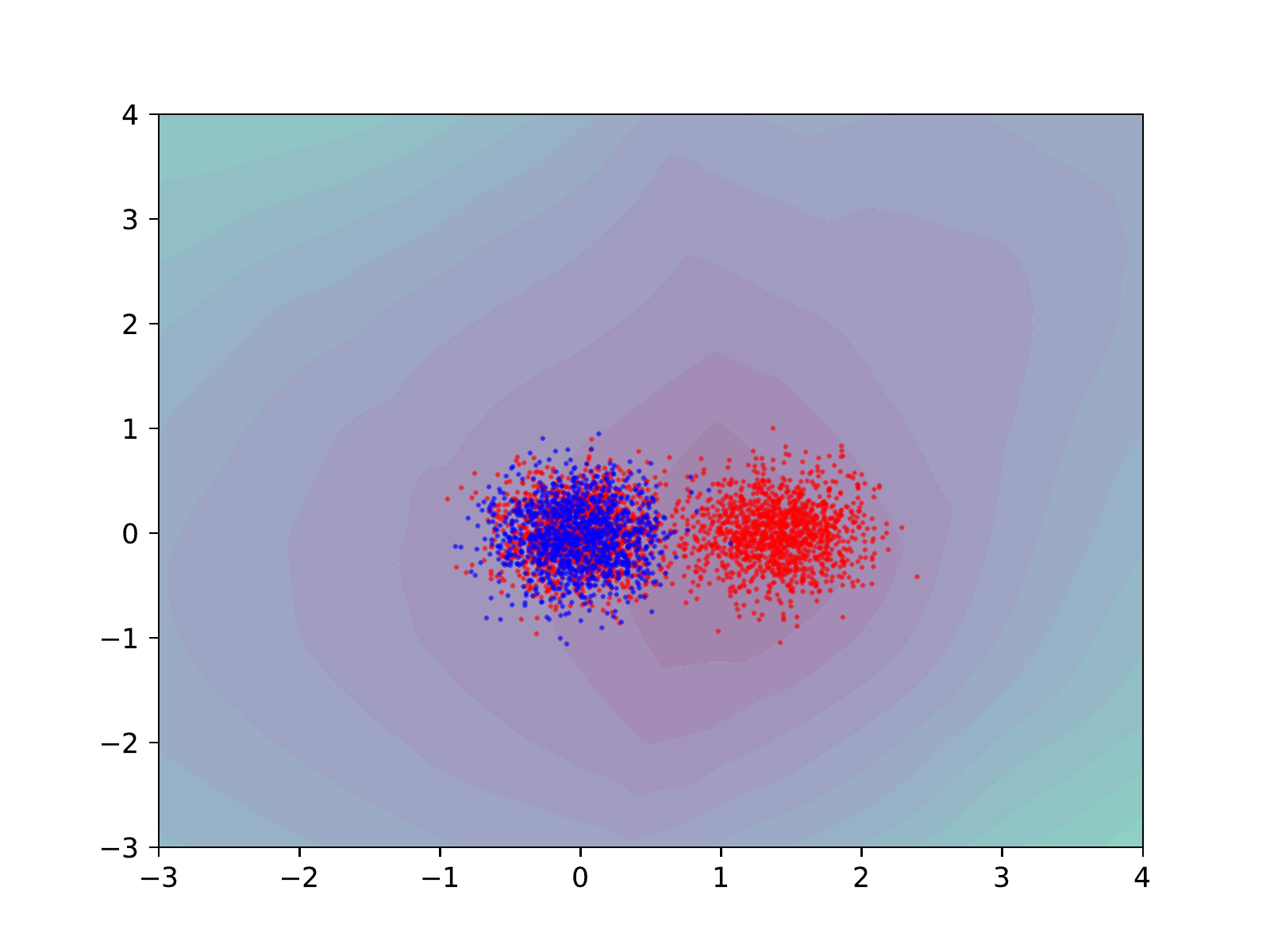}}
		\subfloat[$G_2$]{\includegraphics[width=0.3\linewidth]{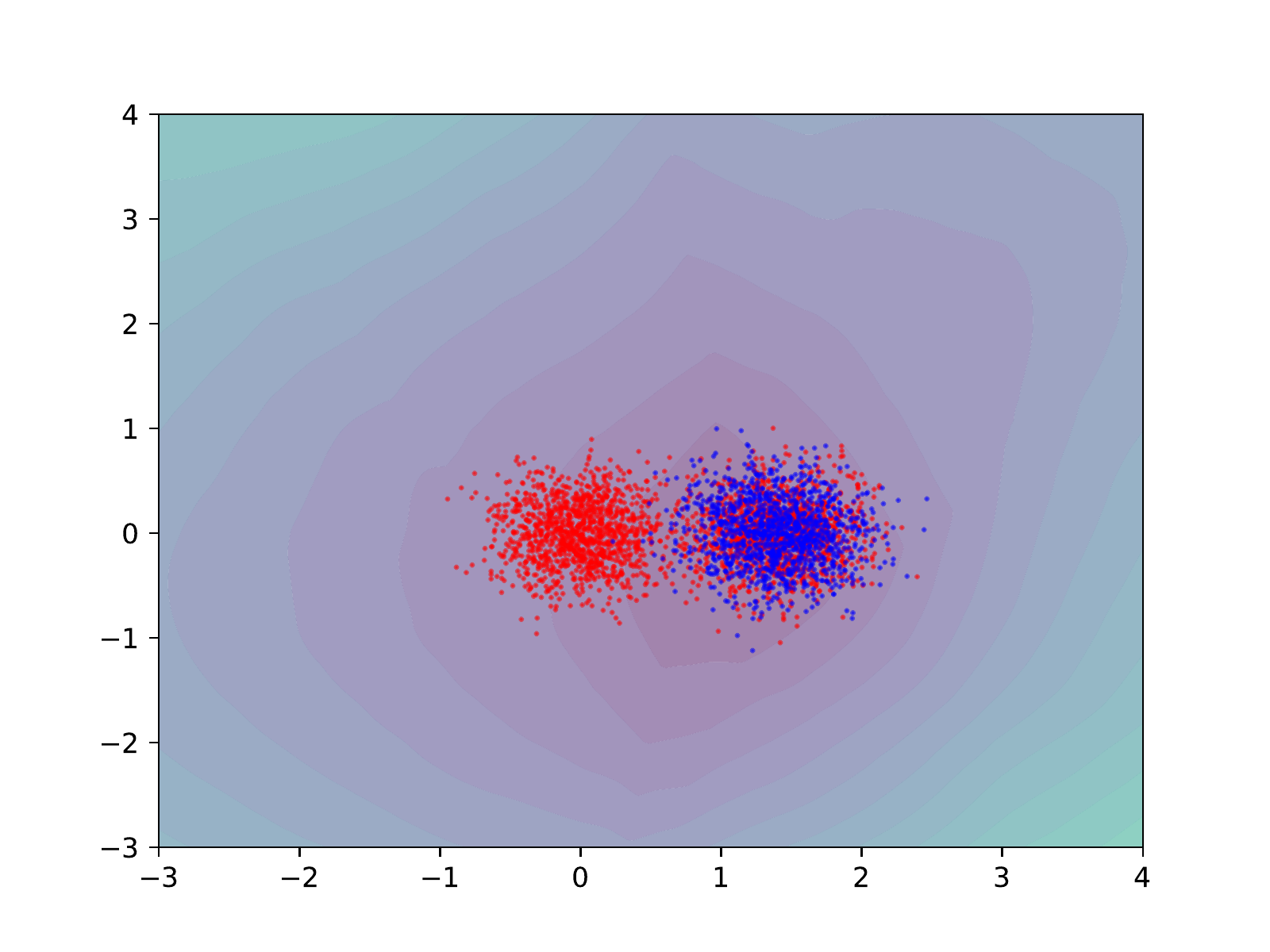}}
		\subfloat[$\bigcup_{i=1}^{2}G_i$]{\includegraphics[width=0.3\linewidth]{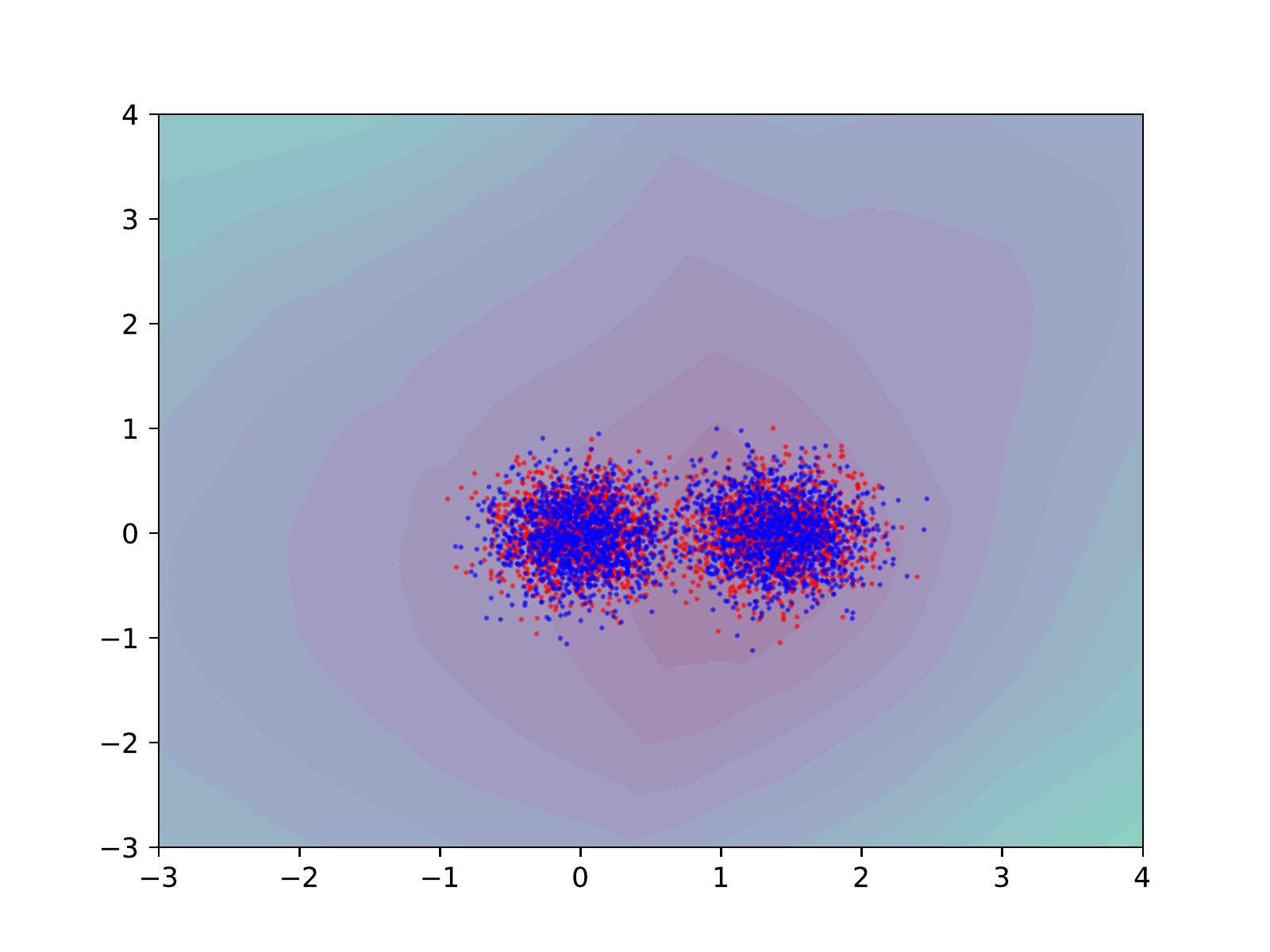}}
	\end{center}
	\caption{Projections of real data (red dots) and samples (blue dots) generated by different generators of a MIX+GAN with 2 generators and 2 discriminators trained on the 2-Gaussians dataset.}
	\label{2G2D}
\end{figure}

\begin{figure}[!h]
	\begin{center}
		\subfloat[$G_1$]{\includegraphics[width=0.24\linewidth]{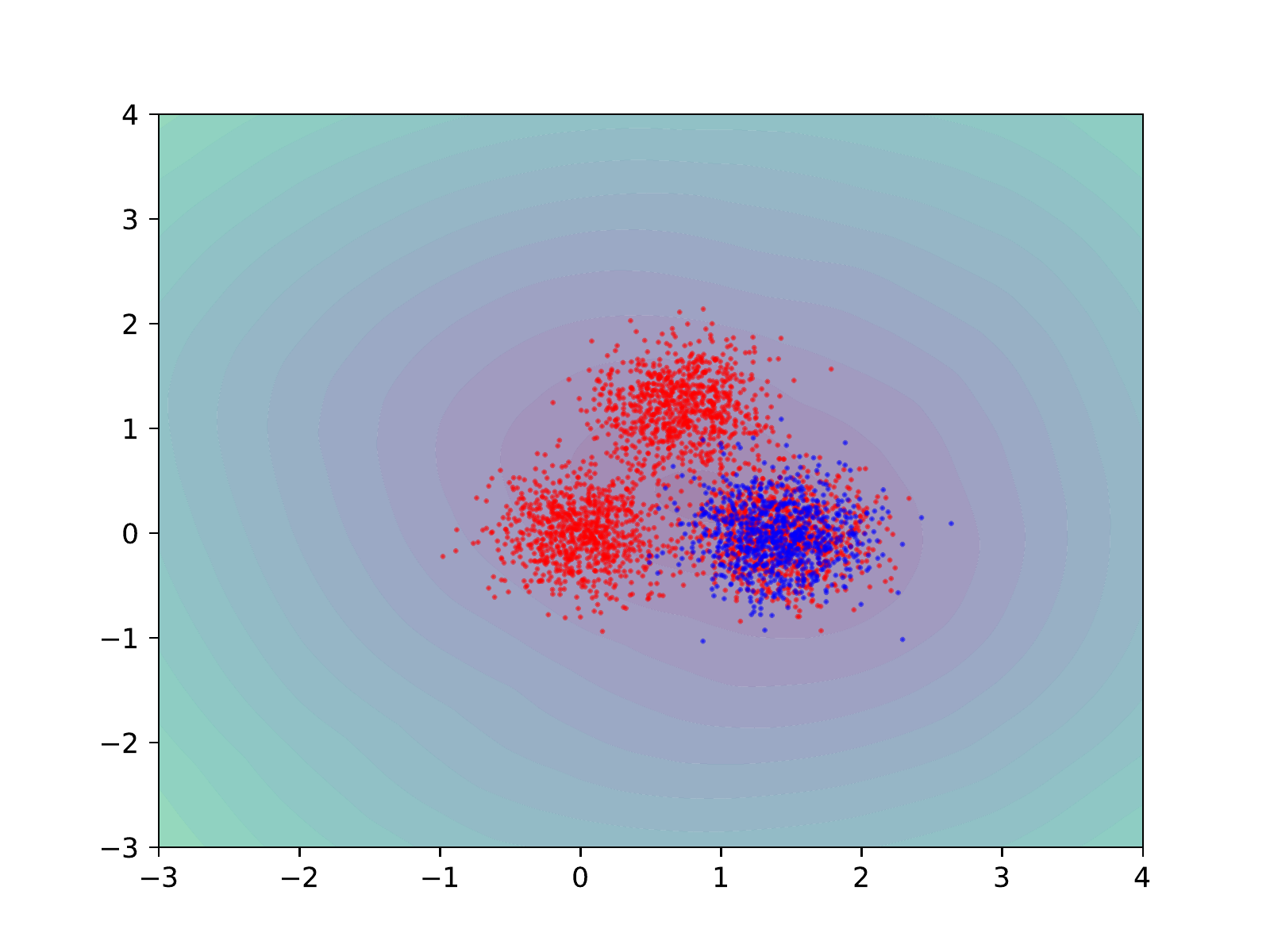}
		}
		\subfloat[$G_2$]{\includegraphics[width=0.24\linewidth]{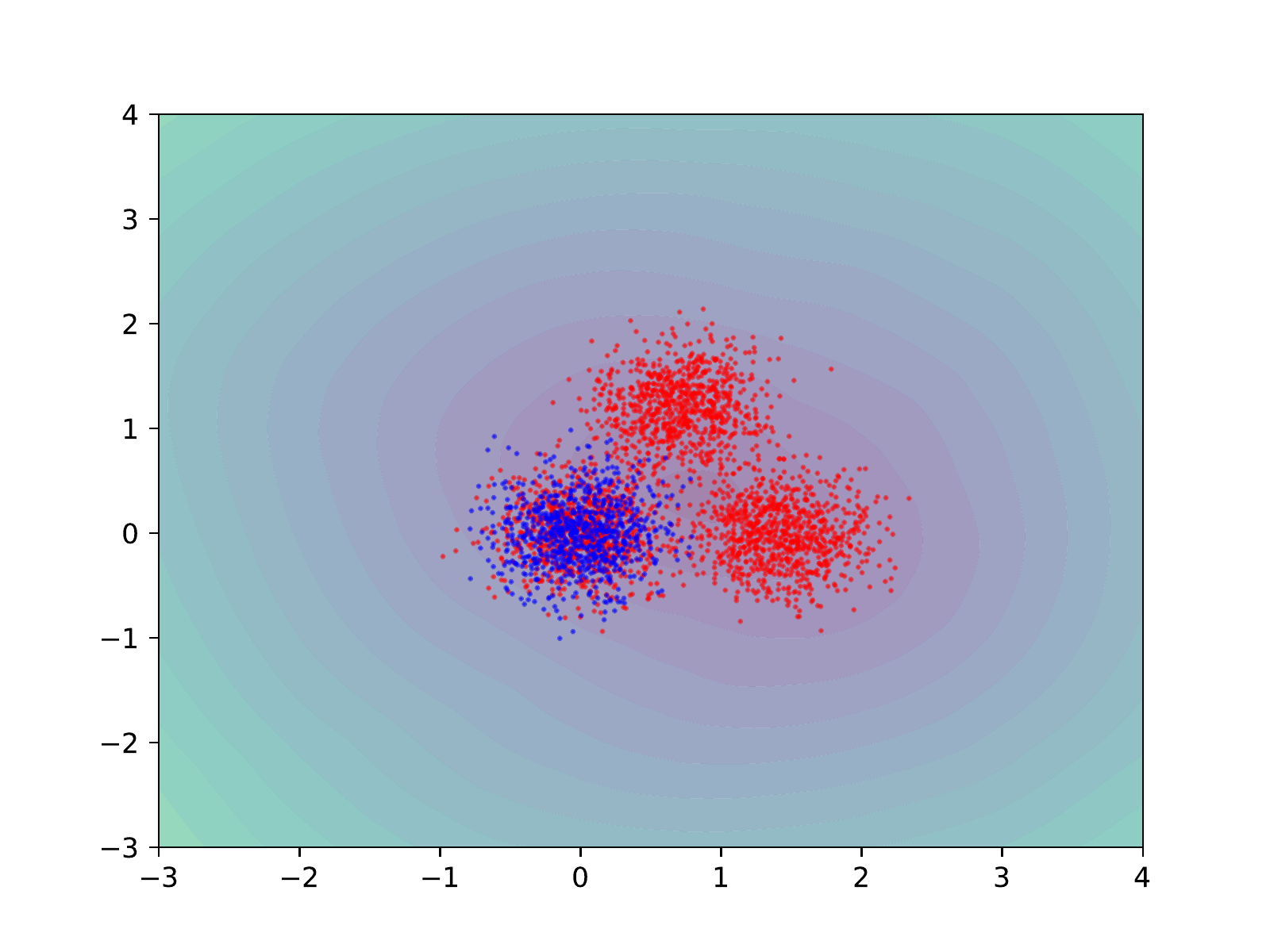}}
		\subfloat[$G_3$]{\includegraphics[width=0.24\linewidth]{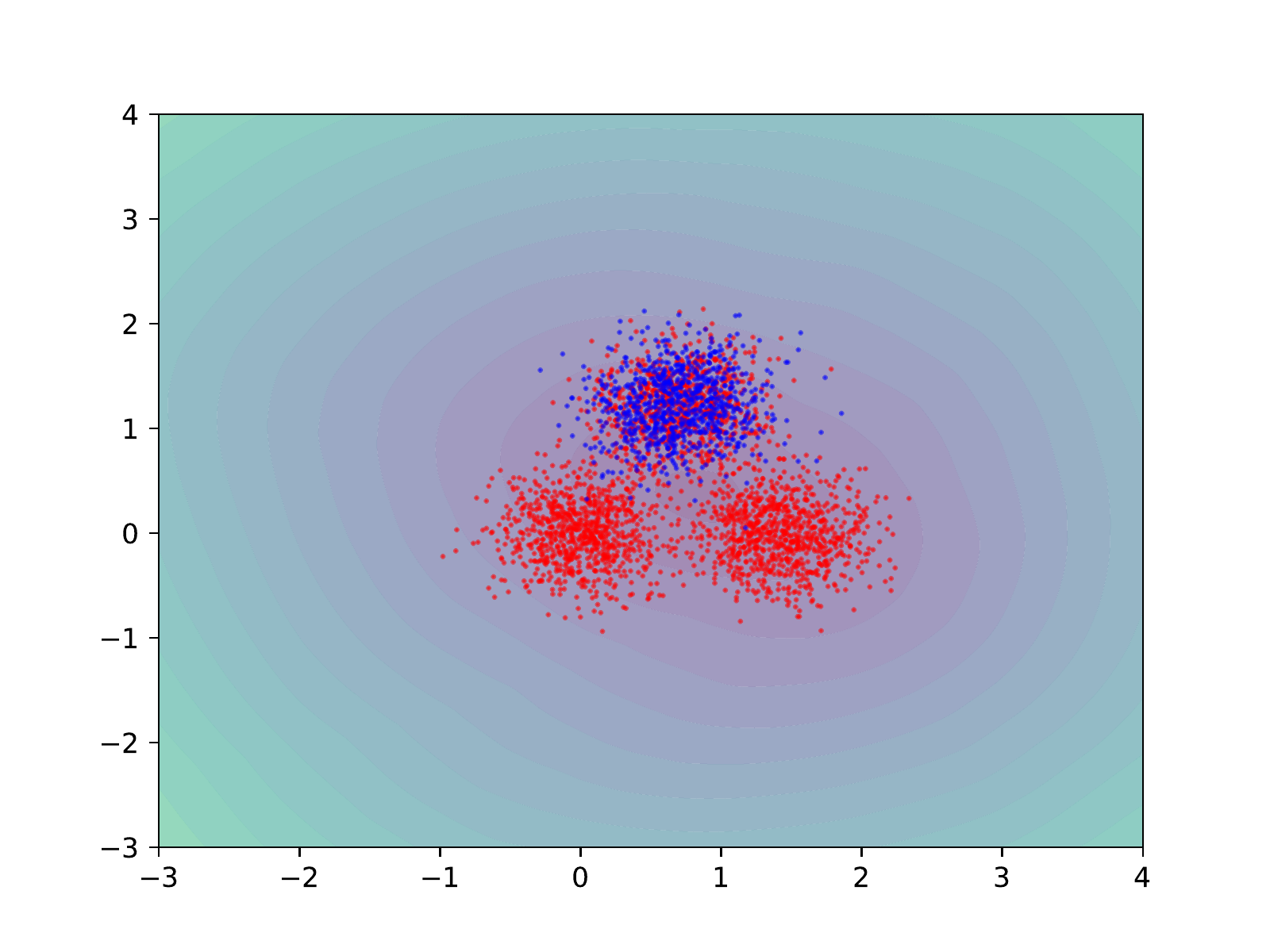}}
		\subfloat[$\bigcup_{i=1}^{3}G_i$]{\includegraphics[width=0.24\linewidth]{MIX+GAN/33.pdf}}
	\end{center}
	\caption{Projections of real data (red dots) and samples (blue dots) generated by different generators of a MIX+GAN with 3 generators and 3 discriminators trained on the 3-Gaussians dataset.}
	\label{3G3D}
\end{figure}

\begin{figure}[!h]
	\begin{center}
		\subfloat[$G_1$]{\includegraphics[width=0.3\linewidth]      {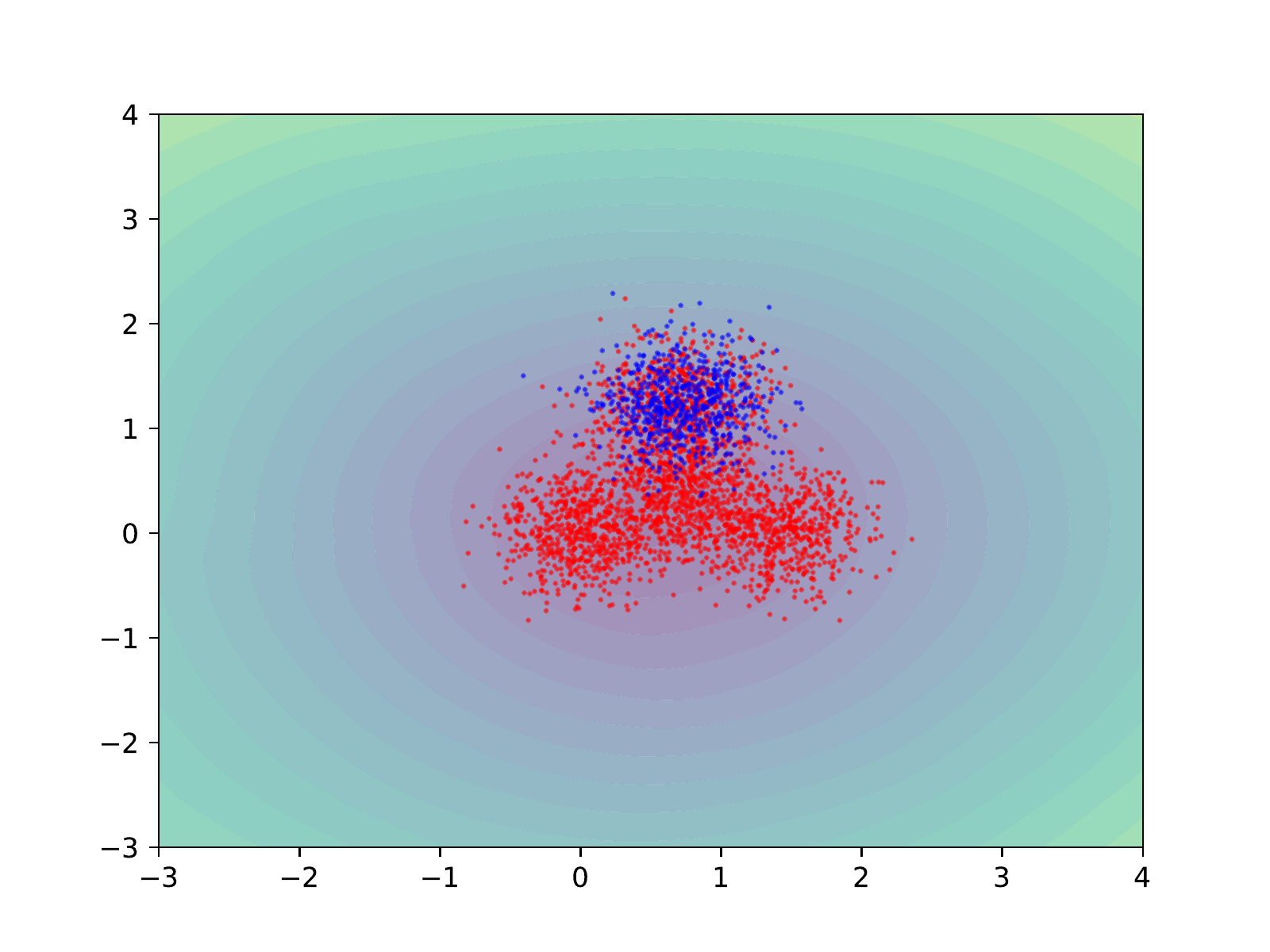}
		}
		\subfloat[$G_2$]{\includegraphics[width=0.3\linewidth]{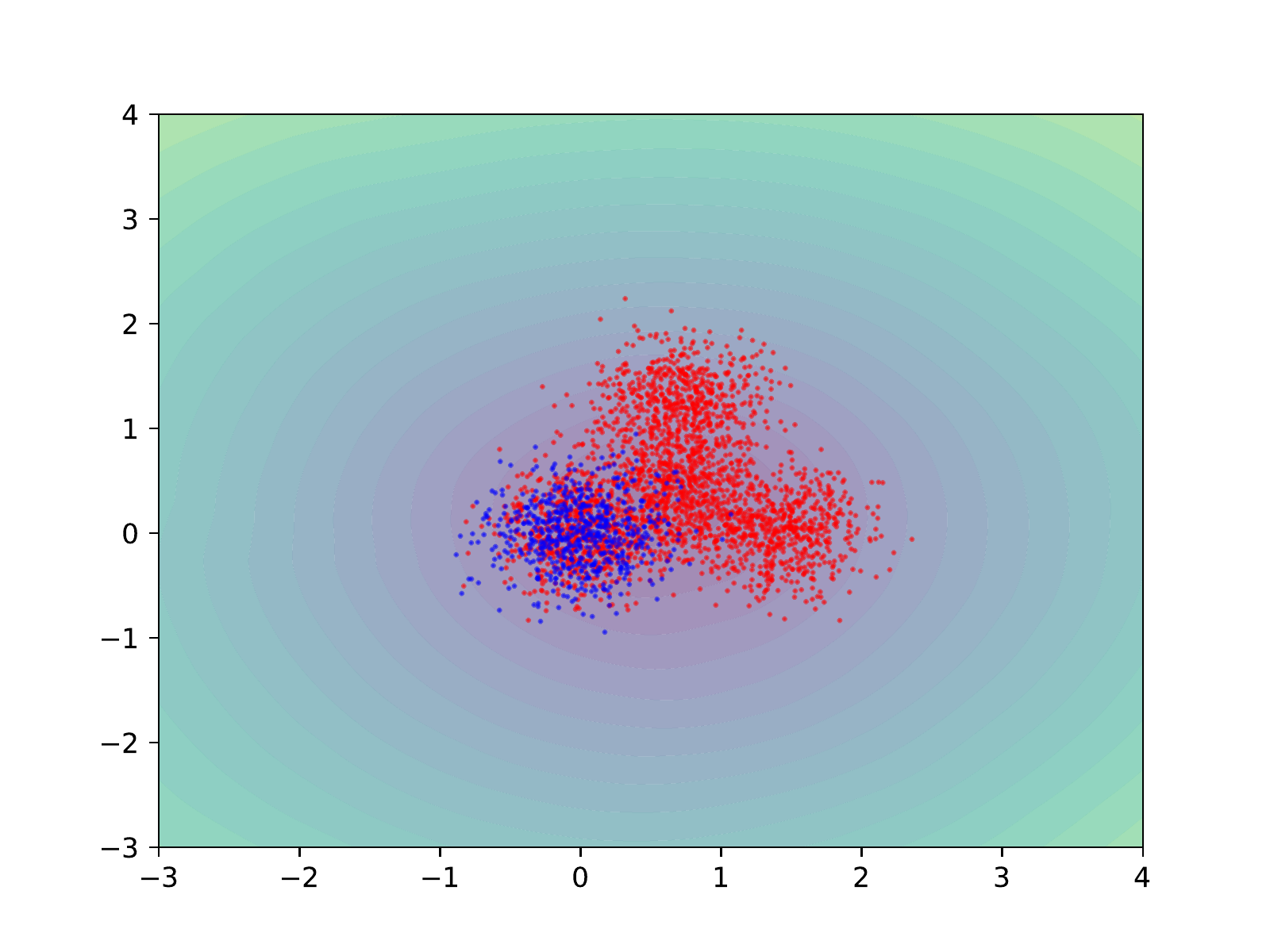}}
		\subfloat[$G_3$]{\includegraphics[width=0.3\linewidth]{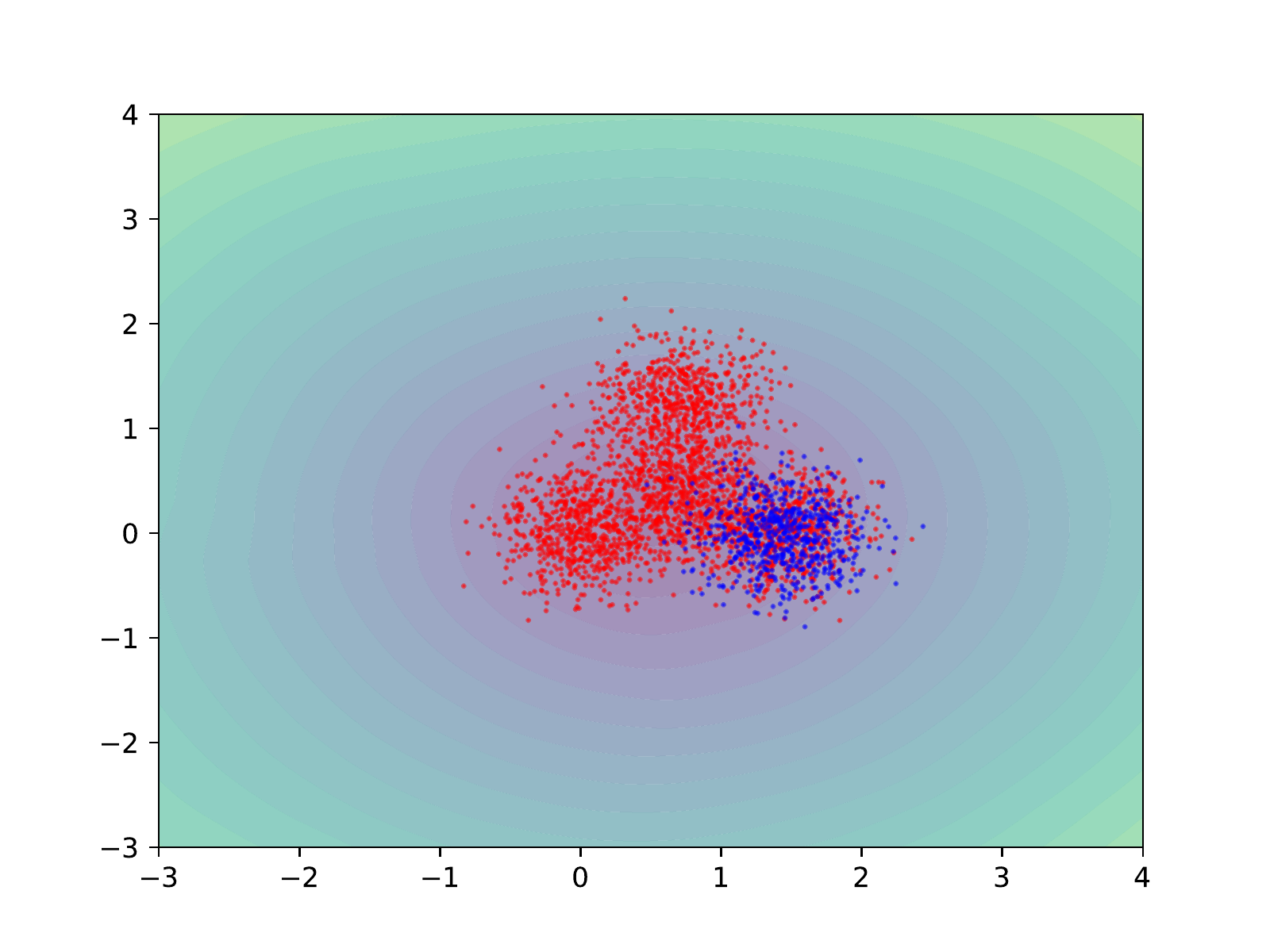}}\\
		\subfloat[$G_4$]{\includegraphics[width=0.3\linewidth]{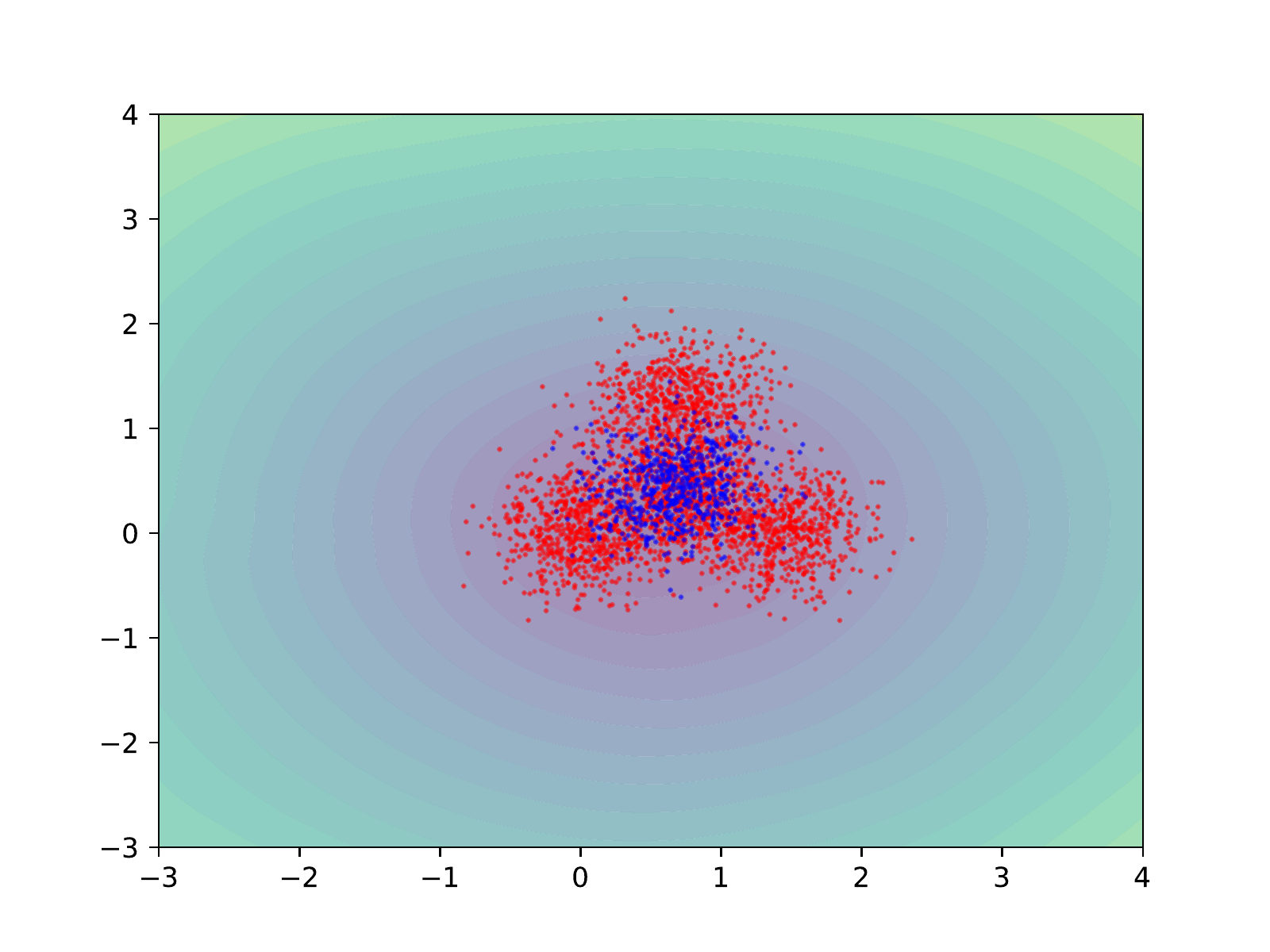}}
		\subfloat[$\bigcup_{i=1}^{4}G_i$]{\includegraphics[width=0.3\linewidth]{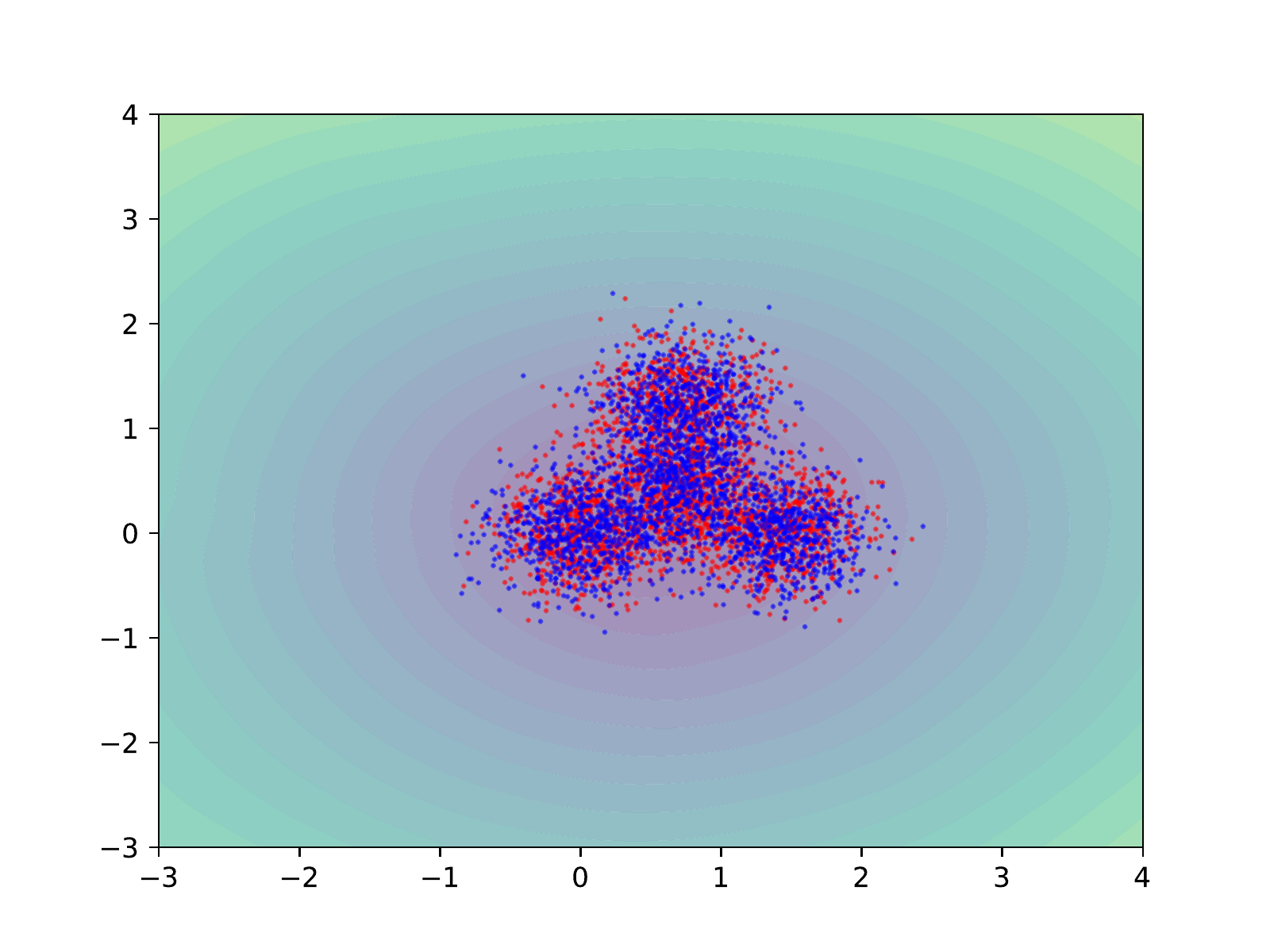}}
	\end{center}
	\caption{Projections of real data (red dots) and samples (blue dots) generated by different generators of a MIX+GAN with 4 generators and 4 discriminators trained on the 4-Gaussians dataset.}
	\label{4G4D}
\end{figure}

\begin{figure}[!h]
	\begin{center}
		\subfloat[$G_1$]{\includegraphics[width=0.3\linewidth]      {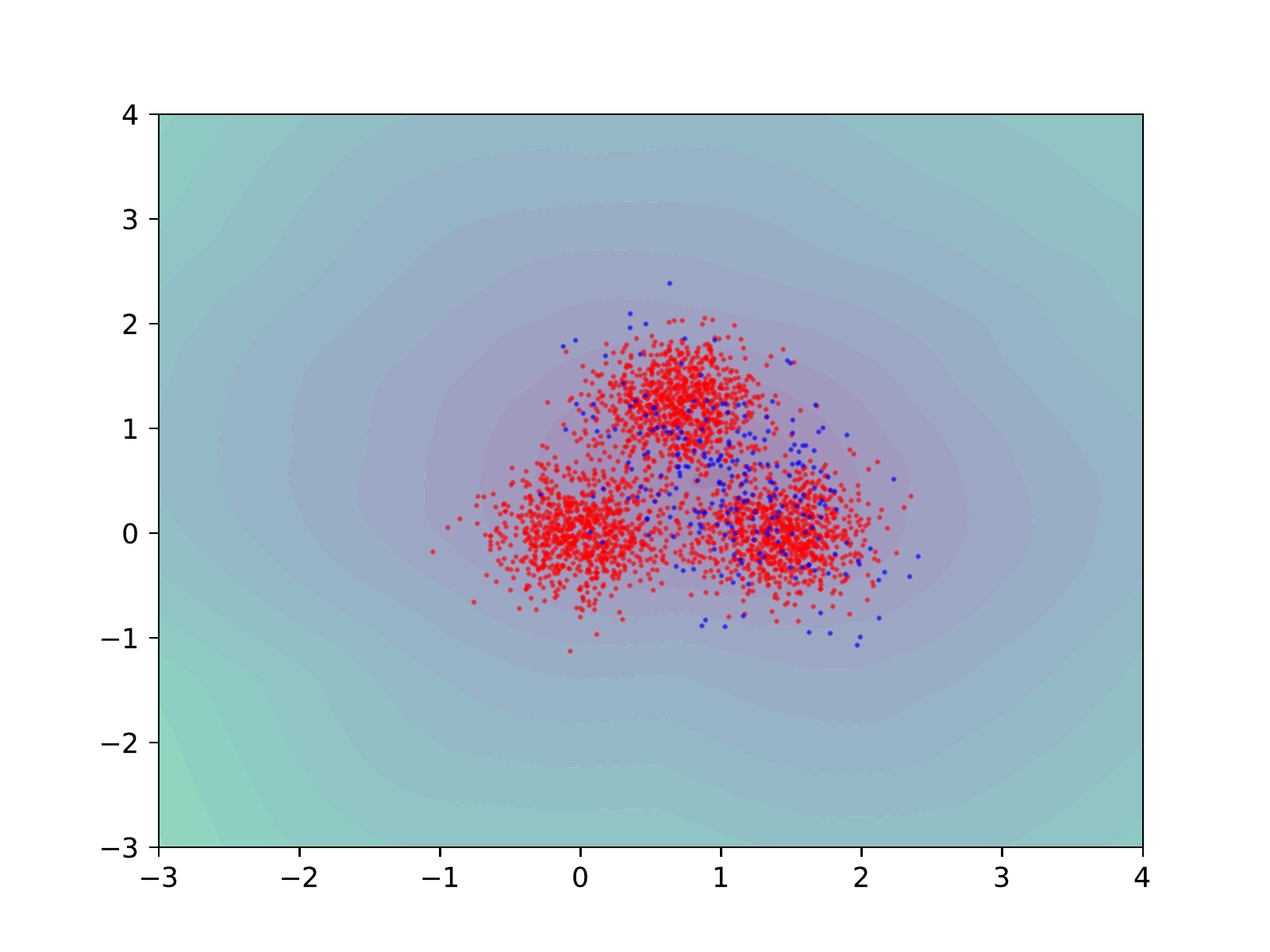}
		}
		\subfloat[$G_2$]{\includegraphics[width=0.3\linewidth]{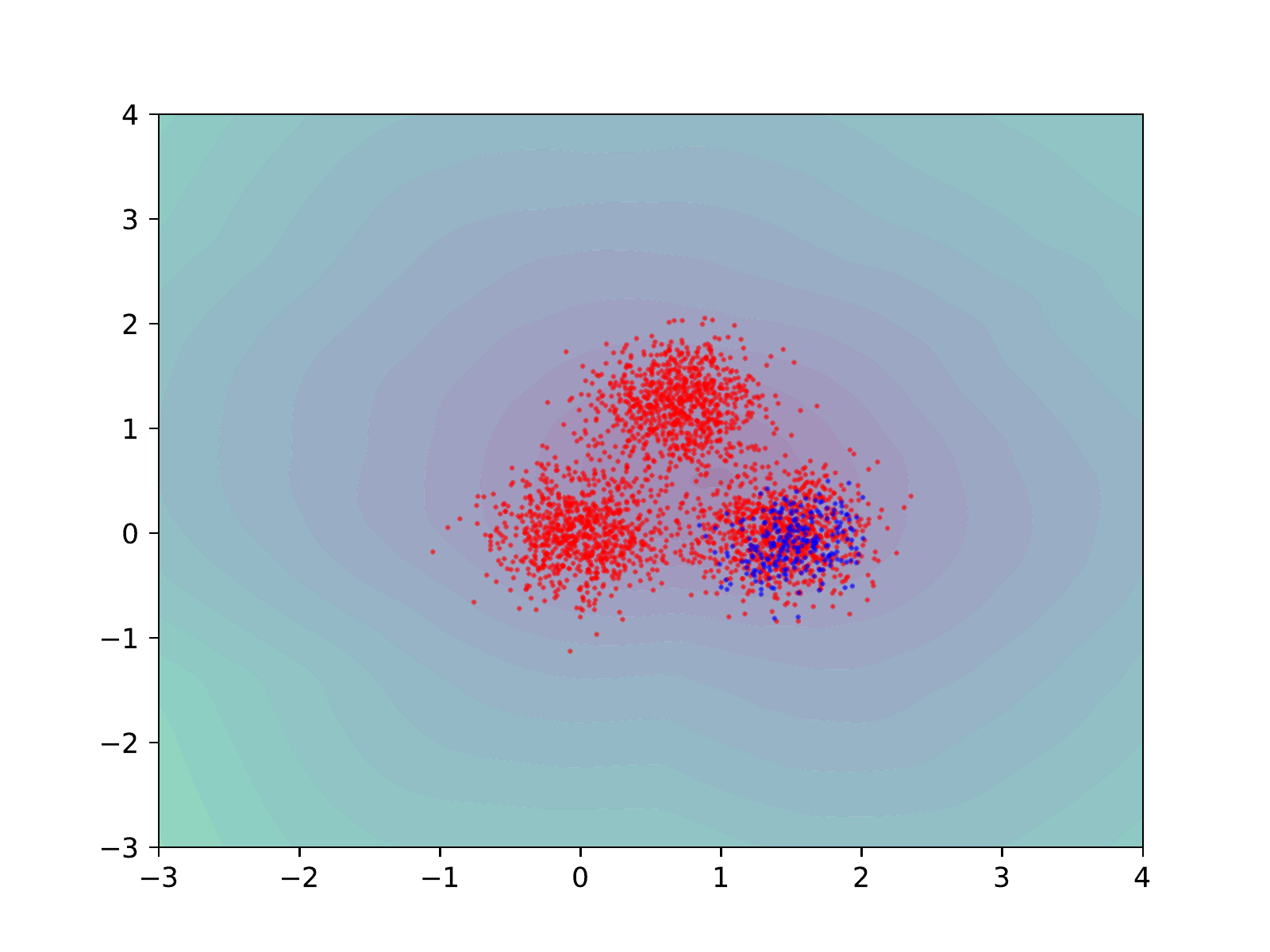}}
		\subfloat[$G_3$]{\includegraphics[width=0.3\linewidth]{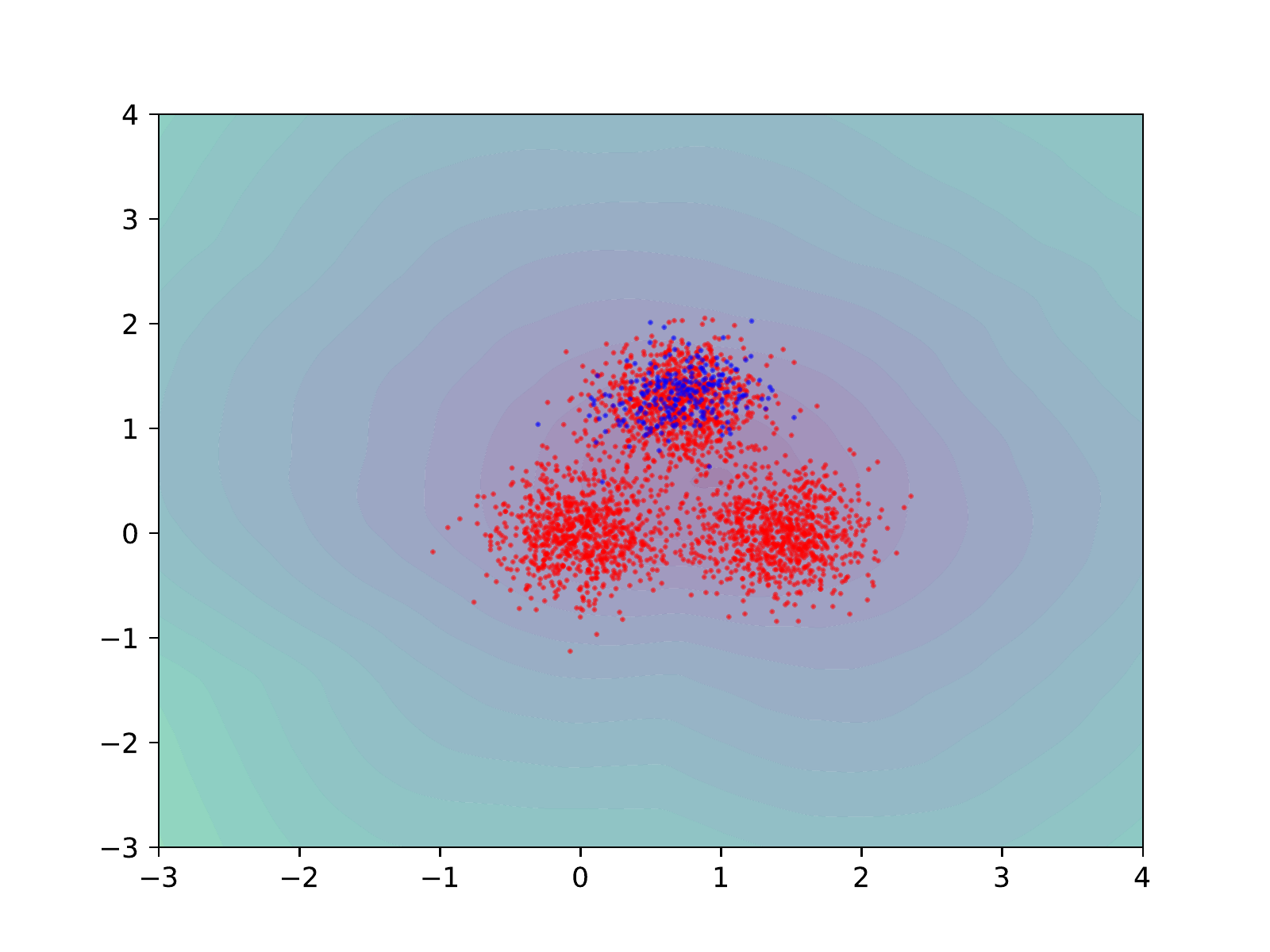}}\\
		\subfloat[$G_4$]{\includegraphics[width=0.3\linewidth]{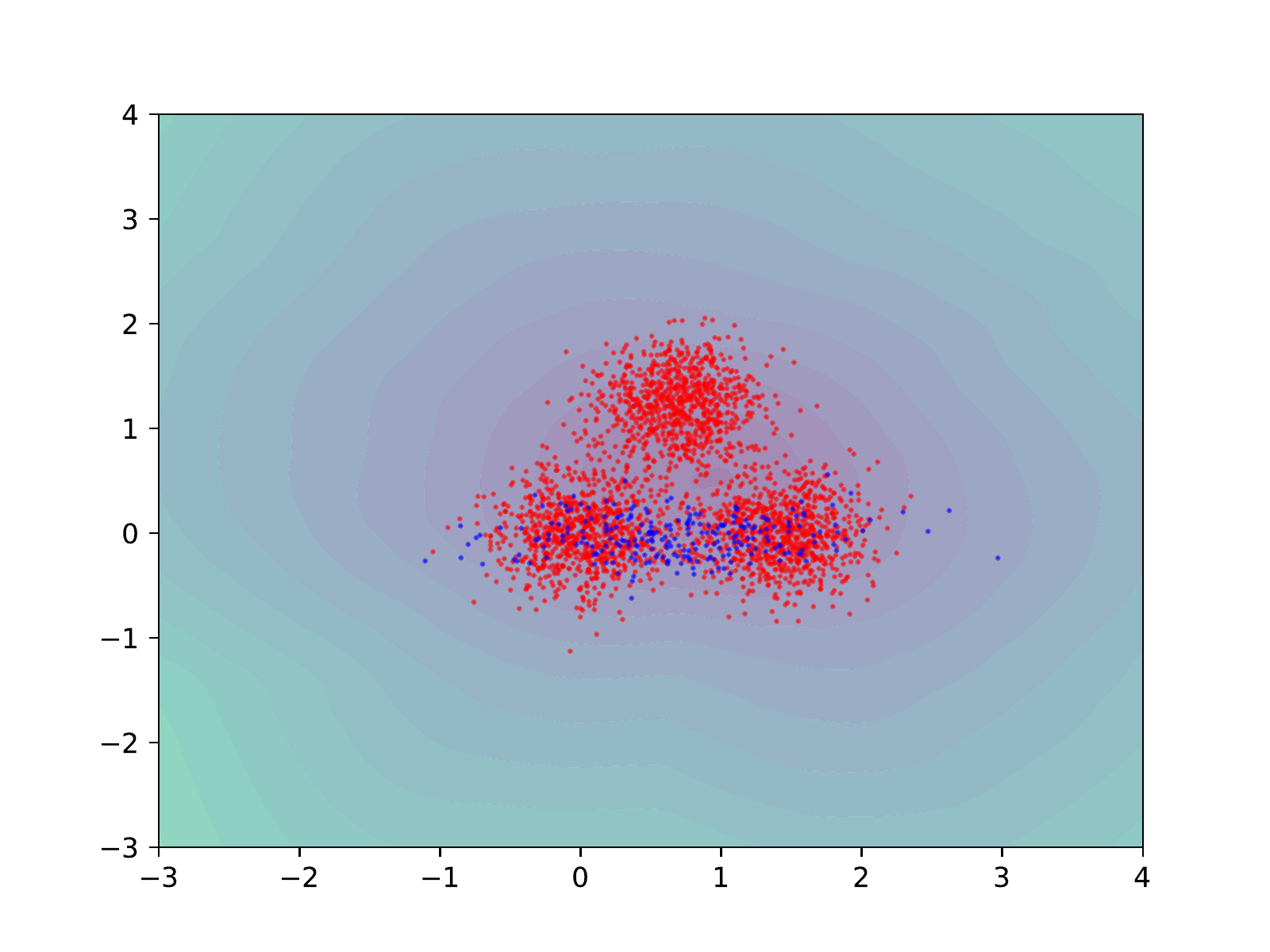}}
		\subfloat[$G_5$]{\includegraphics[width=0.3\linewidth]{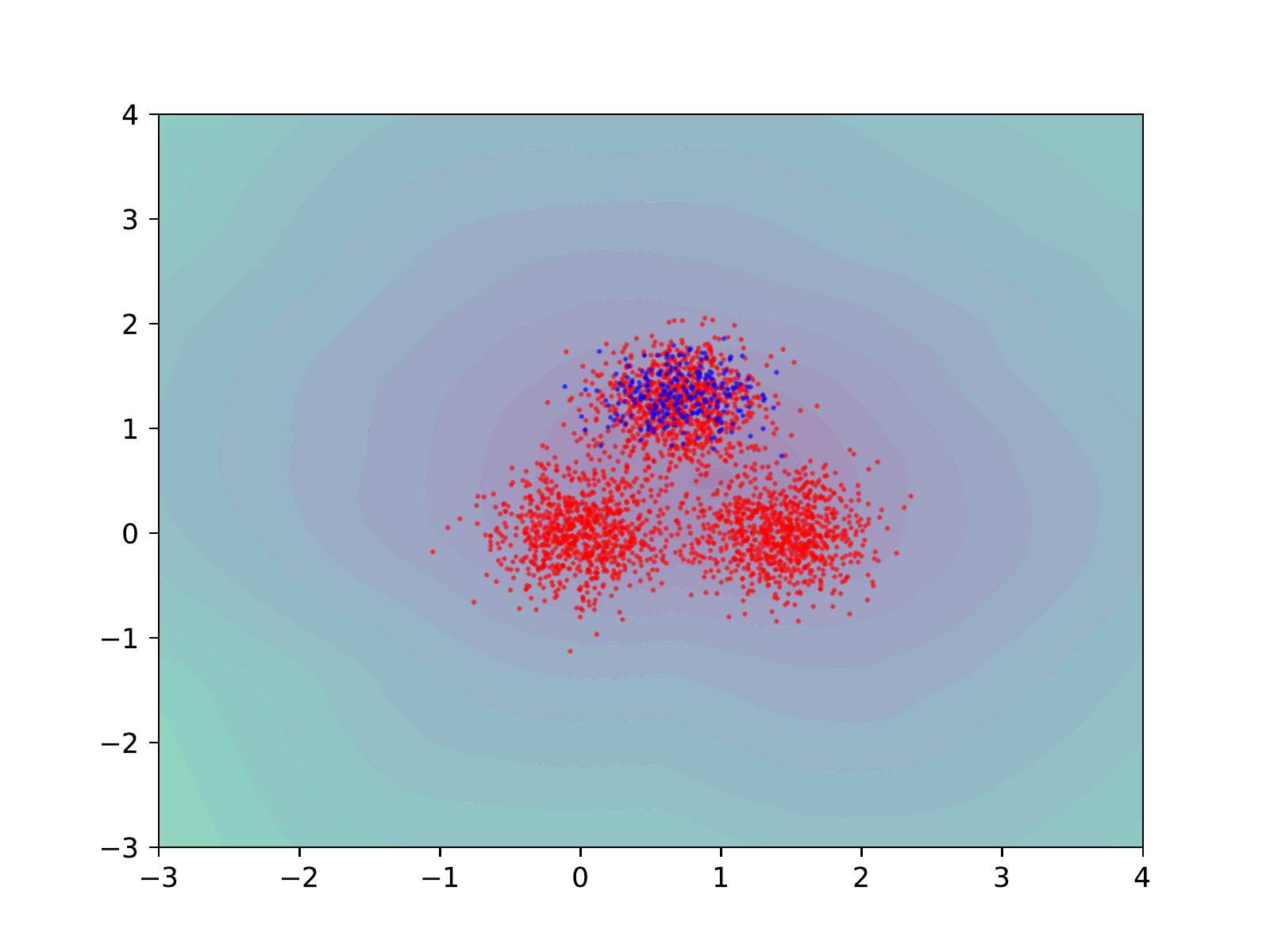}}
		\subfloat[$G_6$]{\includegraphics[width=0.3\linewidth]{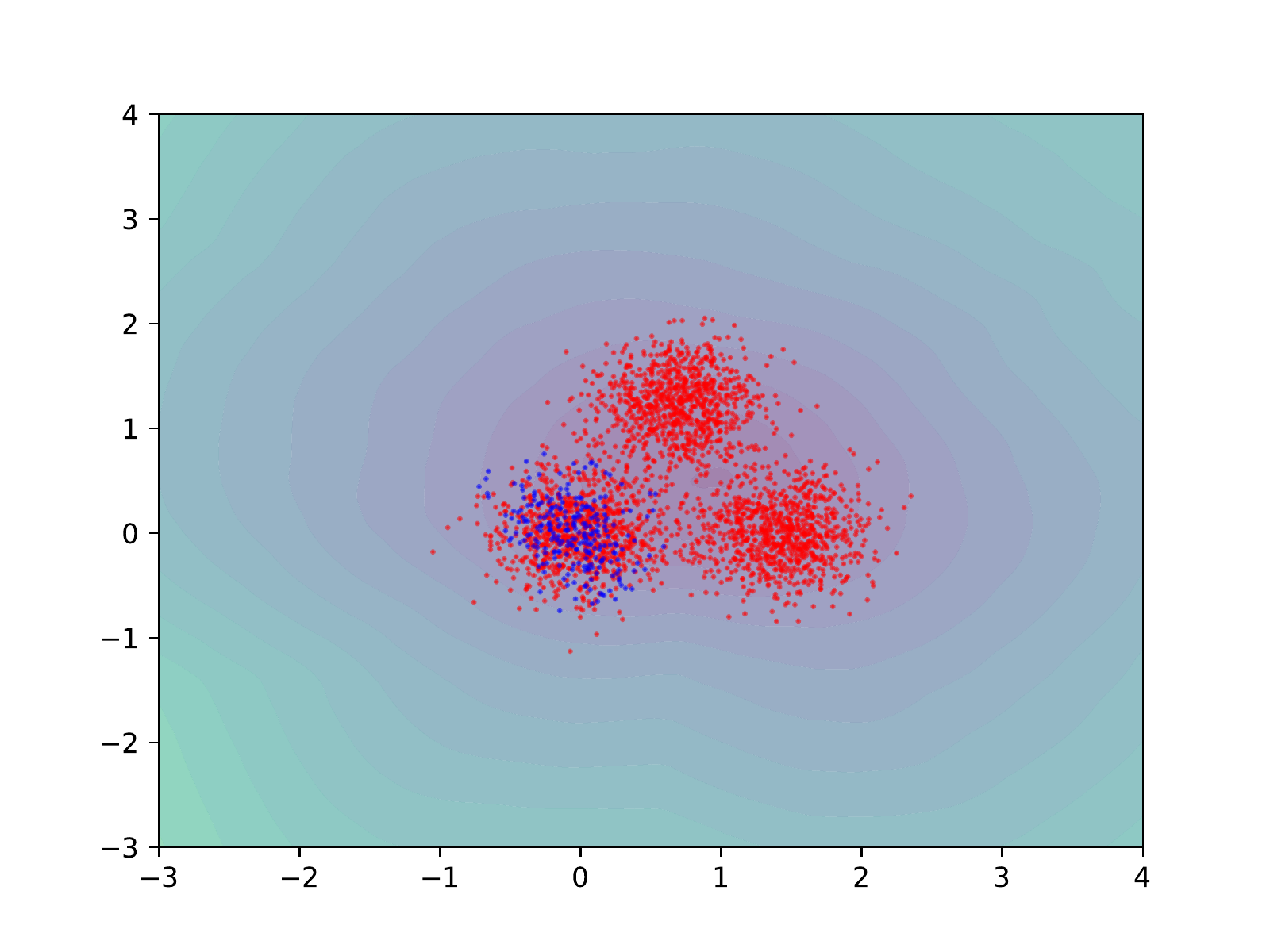}}\\
		\subfloat[$G_7$]{\includegraphics[width=0.3\linewidth]{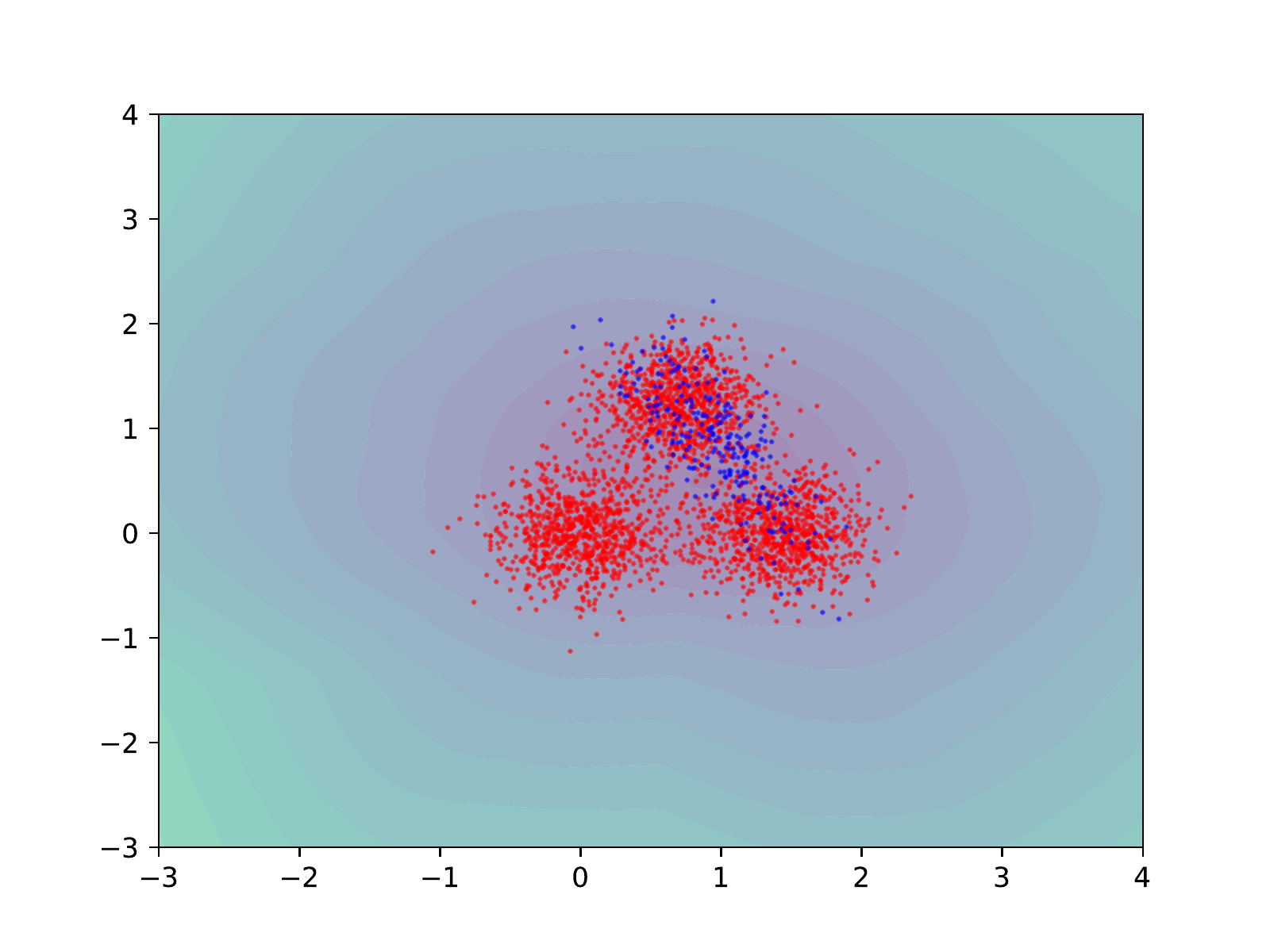}}
		\subfloat[$G_8$]{\includegraphics[width=0.3\linewidth]{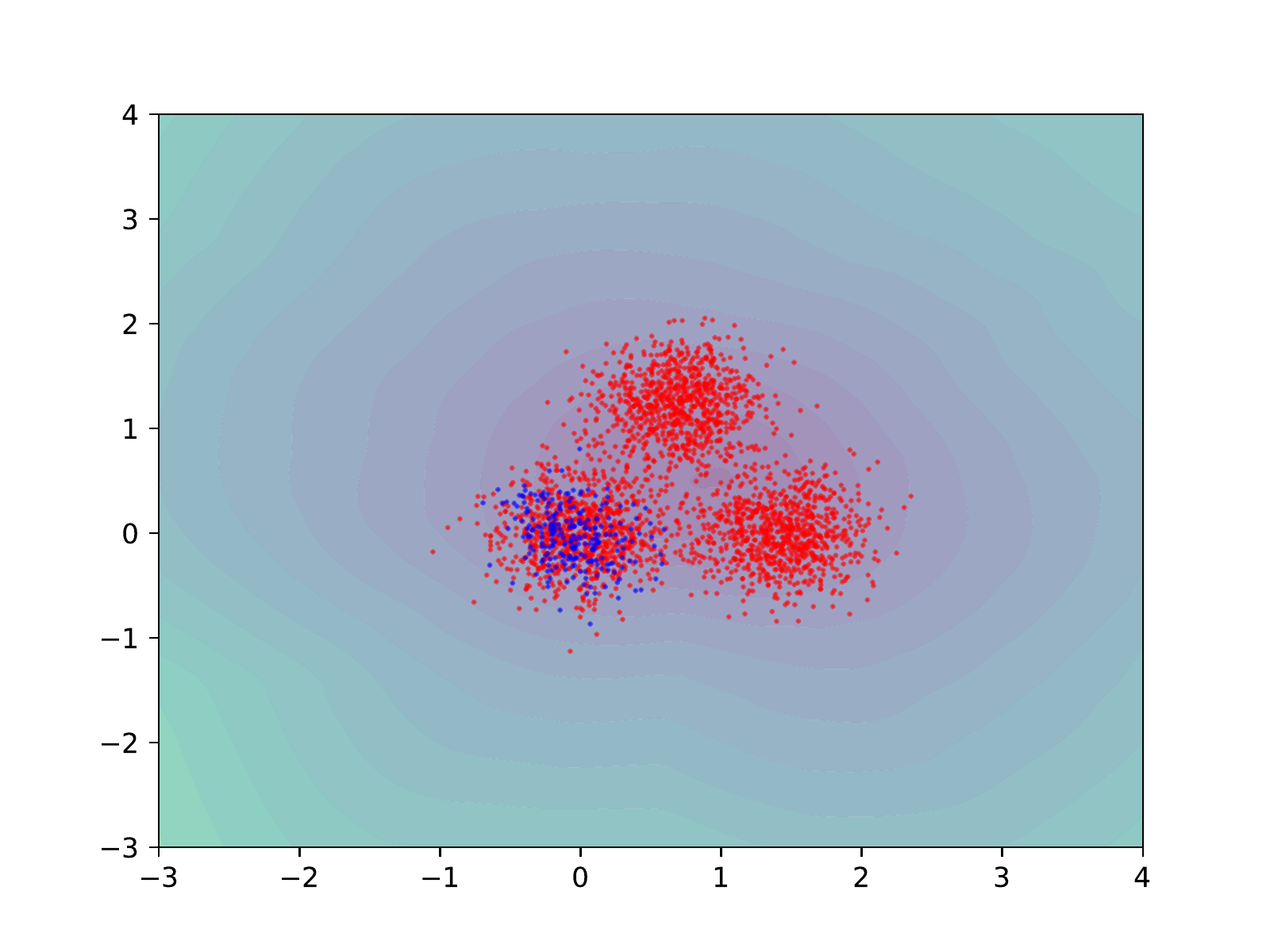}}
		\subfloat[$G_9$]{\includegraphics[width=0.3\linewidth]{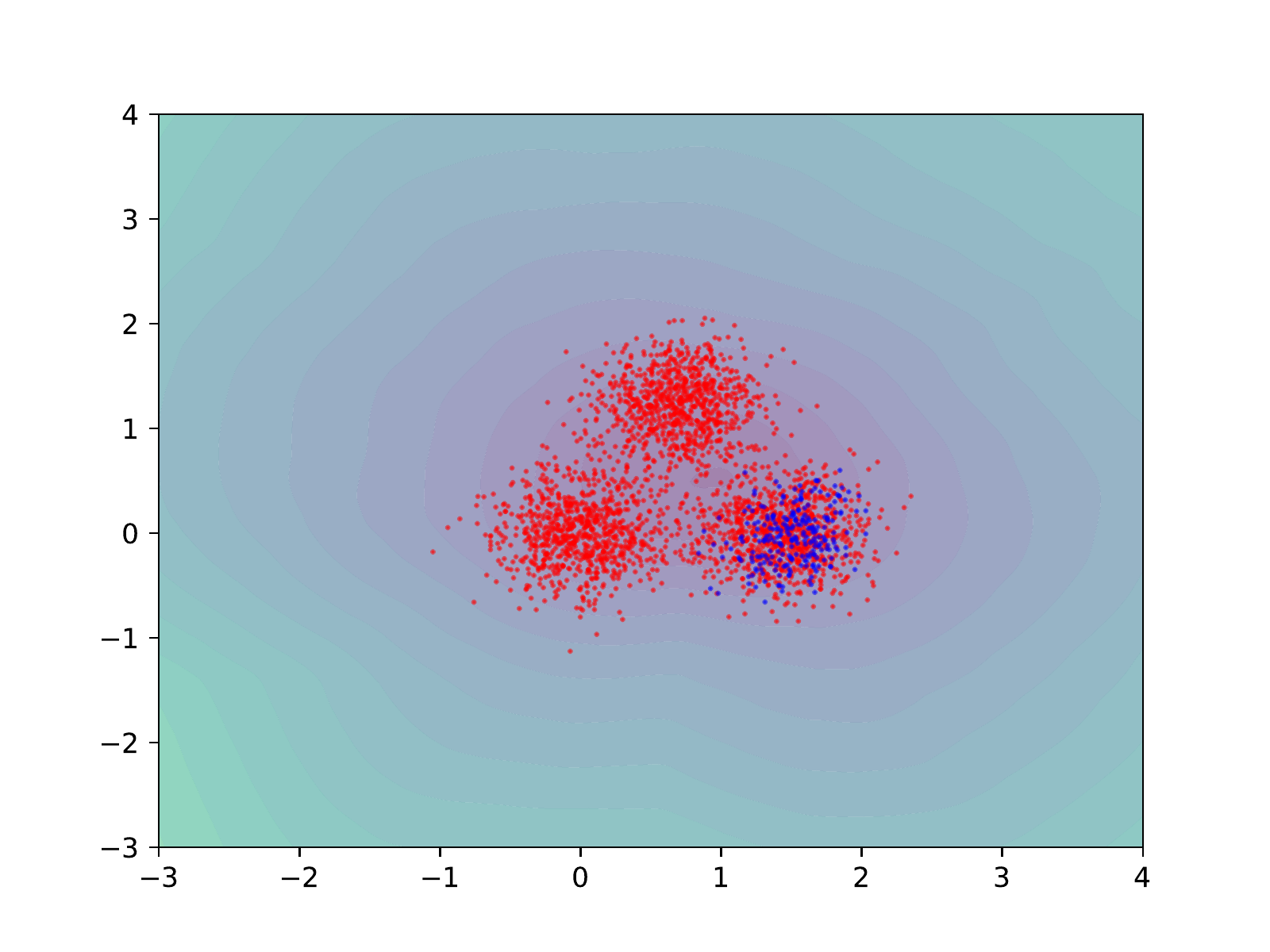}}\\
		\subfloat[$G_{10}$]{\includegraphics[width=0.3\linewidth]{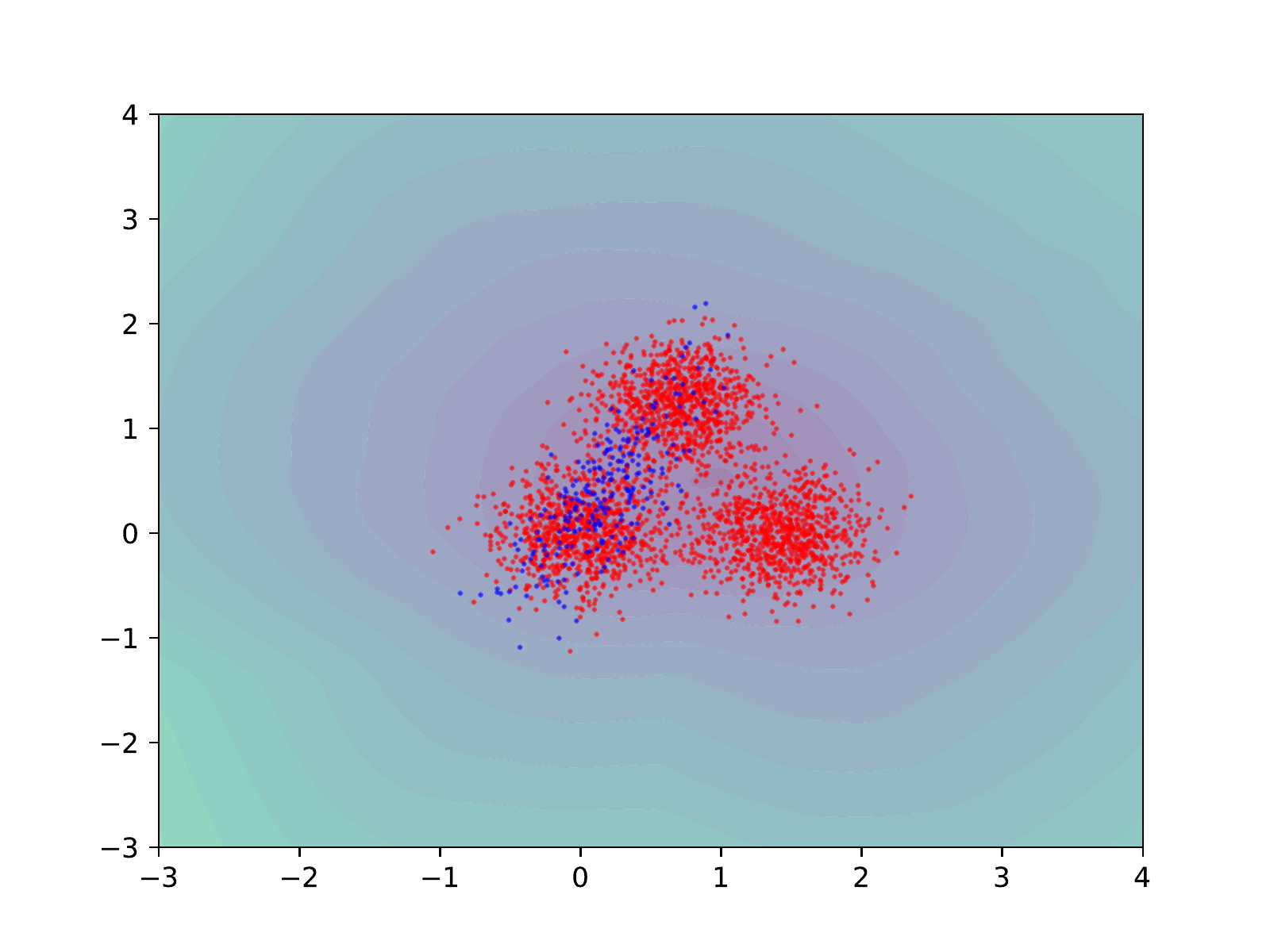}}
		\subfloat[$\bigcup_{i=1}^{10}G_i$]{\includegraphics[width=0.3\linewidth]{MIX+GAN/1010.pdf}}
	\end{center}
	\caption{Projections of real data (red dots) and samples generated by different generators (blue dots) of a MIX+GAN of 10 generators and 10 discriminators trained on the 3-Gaussians dataset.}
	\label{10G10D}
\end{figure}

An intriguing phenomenon is observed when the number of generators equals the number of Gaussian components.
In Figure \ref{2G2D}, \ref{3G3D} and \ref{4G4D}, we show different fractions of samples generated by different generators in a MIX+GAN. The results in Figure \ref{2G2D} and Figure \ref{3G3D} show that when the number of generators equals the number of Gaussian components, MIX+GAN can roughly make each generator capture one Gaussian component.
When the number of generators exceeds the number of Gaussian components, as is shown in Figure \ref{10G10D}, we can see that each generator generates a small portion of data.

The above empirical results indicate that increasing the mixture size can improve the generative distribution, partly by means of dividing the generation and discrimination tasks across multiple generators and discriminators.

\begin{figure}[!h]
	\begin{center}
		\subfloat[Fréchet distance]{\includegraphics[width=0.45\linewidth]{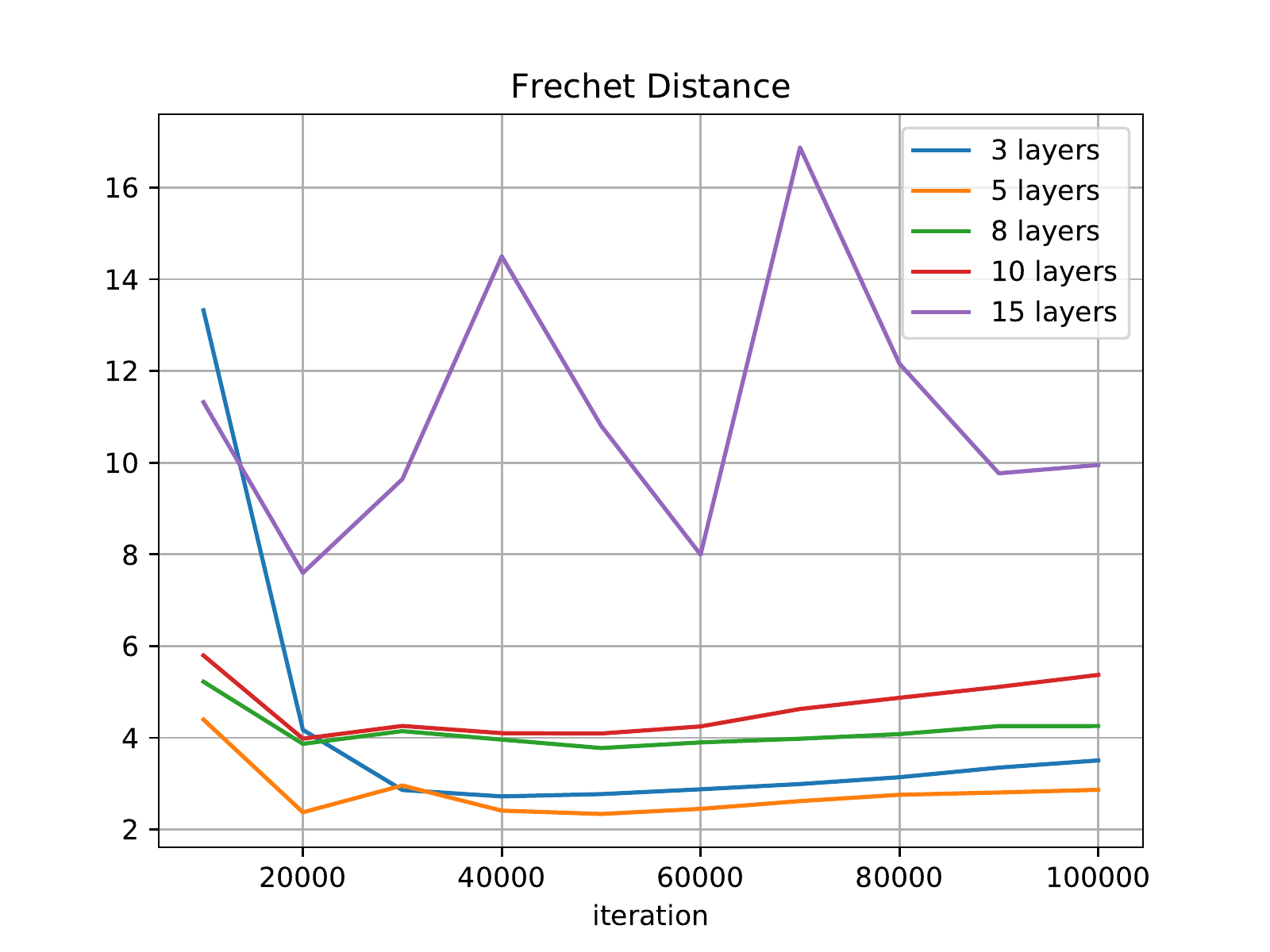}}
		\subfloat[$D(x_r)-D(x_g)$]{\includegraphics[width=0.45\linewidth]{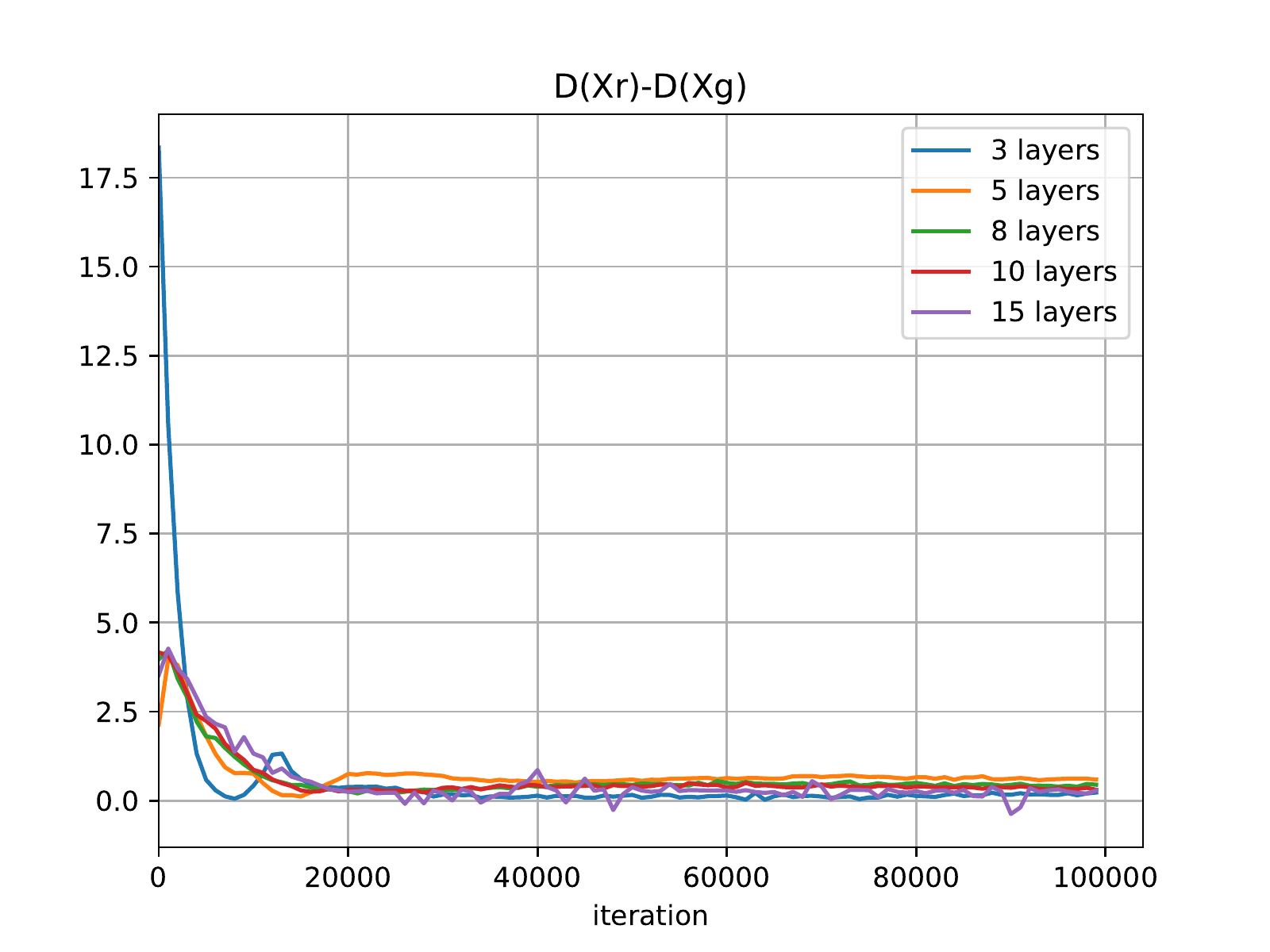}}\\
		\subfloat[Judge accuracy]{\includegraphics[width=0.45\linewidth] {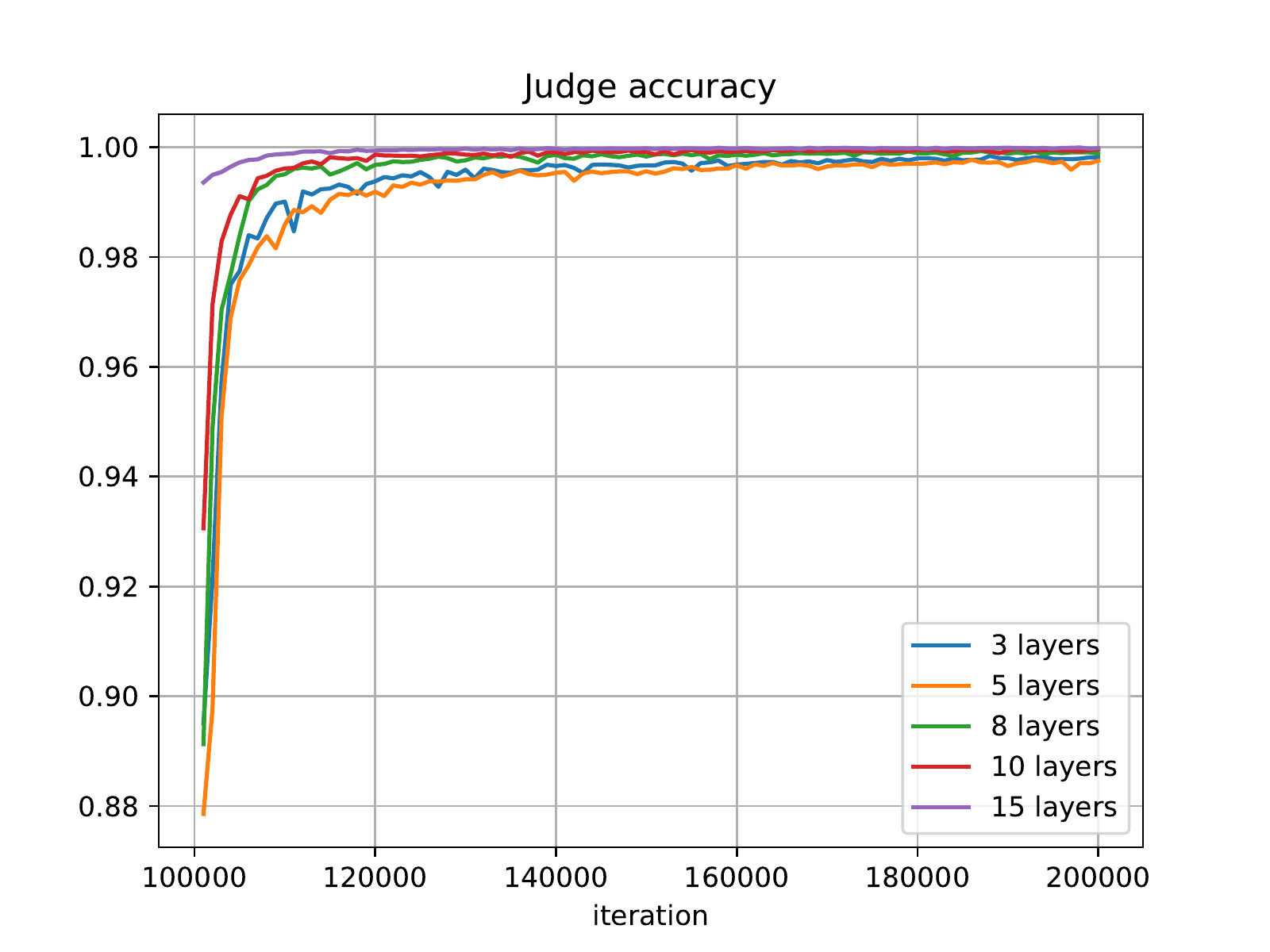}
		}
		\subfloat[Wasserstein distance]{\includegraphics[width=0.45\linewidth]{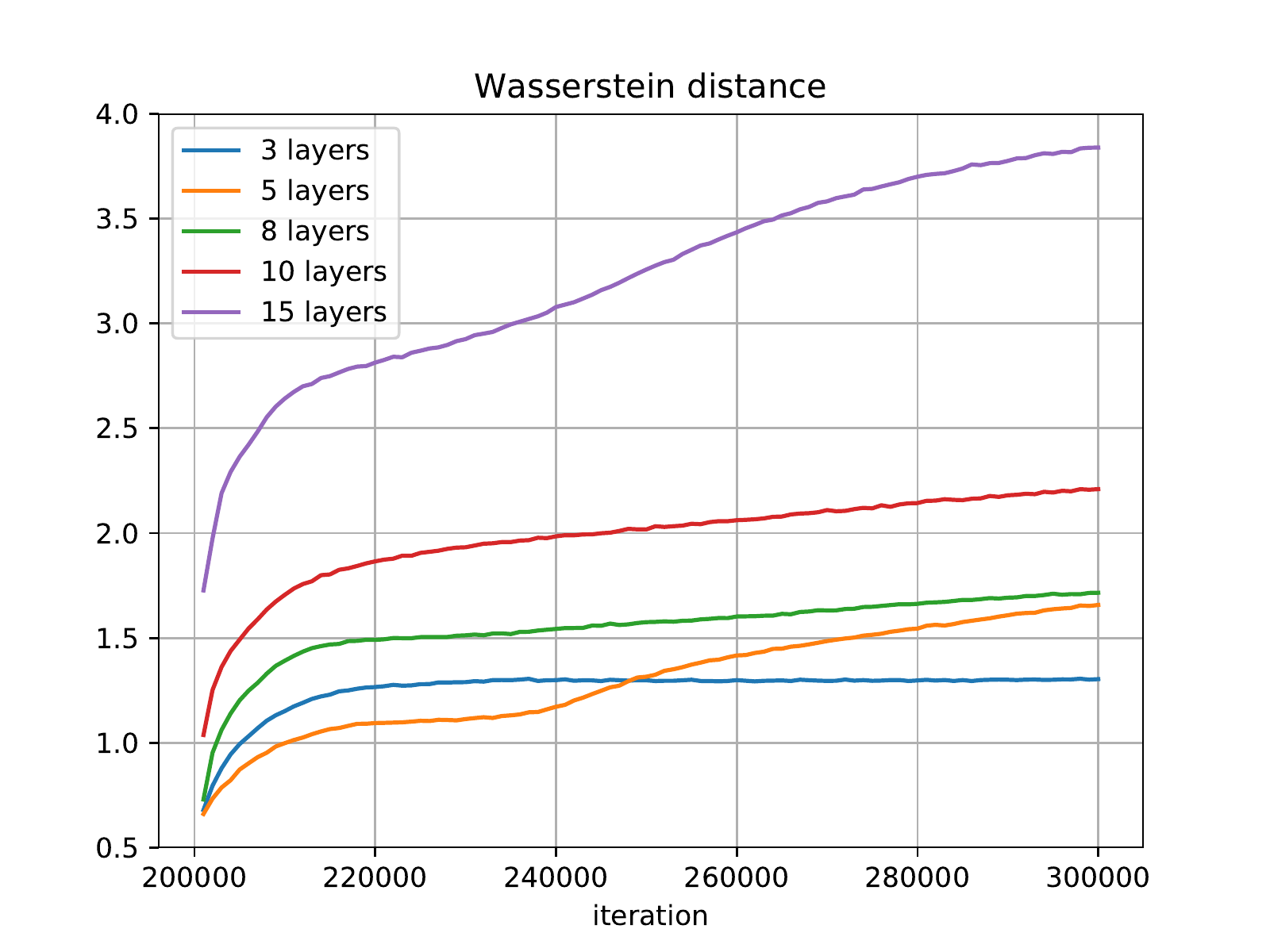}}
	\end{center}
	\caption{Quantitative results of varying the depth of the networks. "$n$ layers" indicates that there are $n$ layers in both the generator and the discriminator. For all the metrics, lower is better.}
	\label{depth_quant}
\end{figure}

\begin{figure}[!h]
	\begin{center}
		\subfloat[Fréchet distance]{\includegraphics[width=0.45\linewidth]{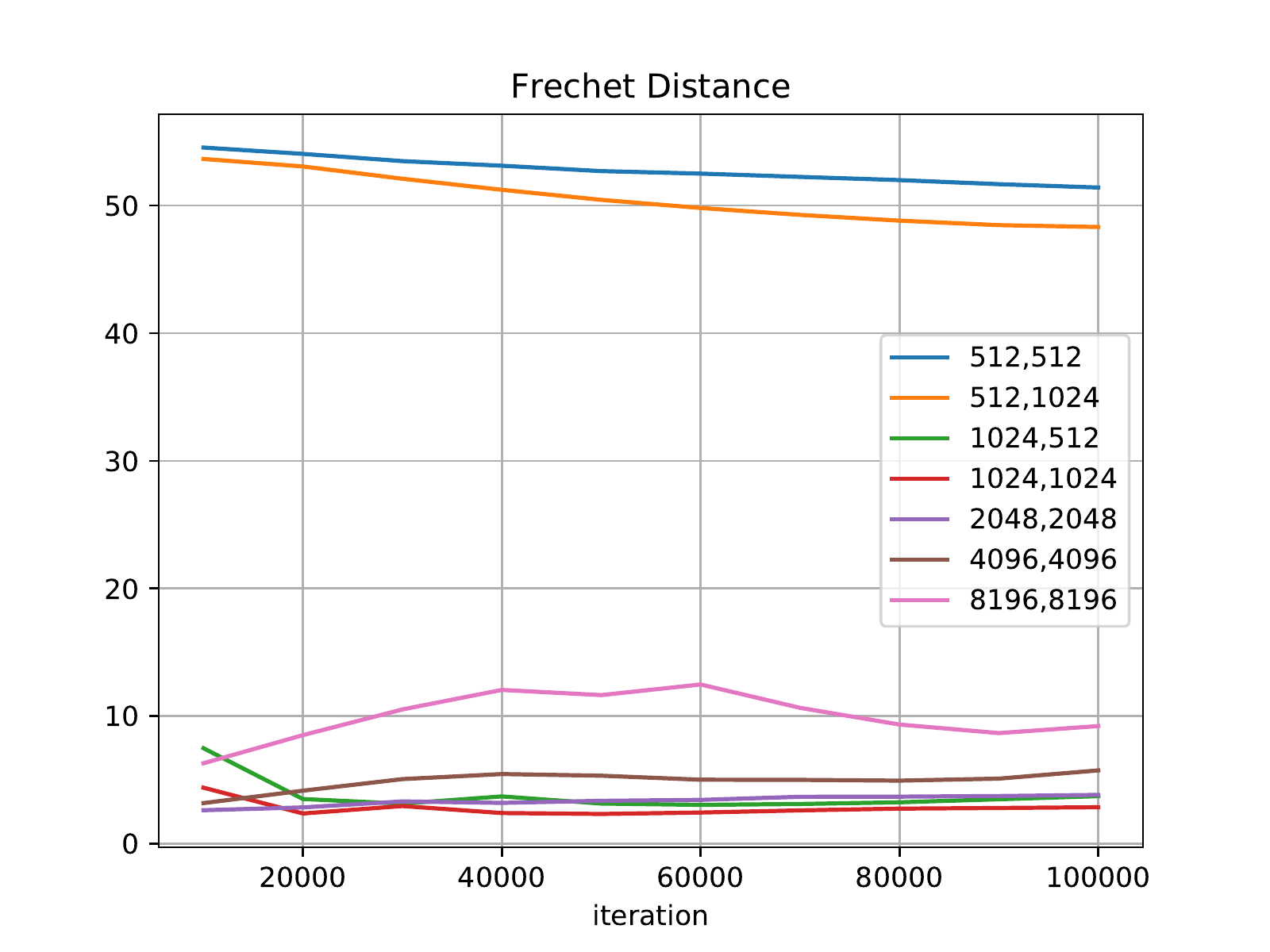}}
		\subfloat[$D(x_r)-D(x_g)$]{\includegraphics[width=0.45\linewidth]{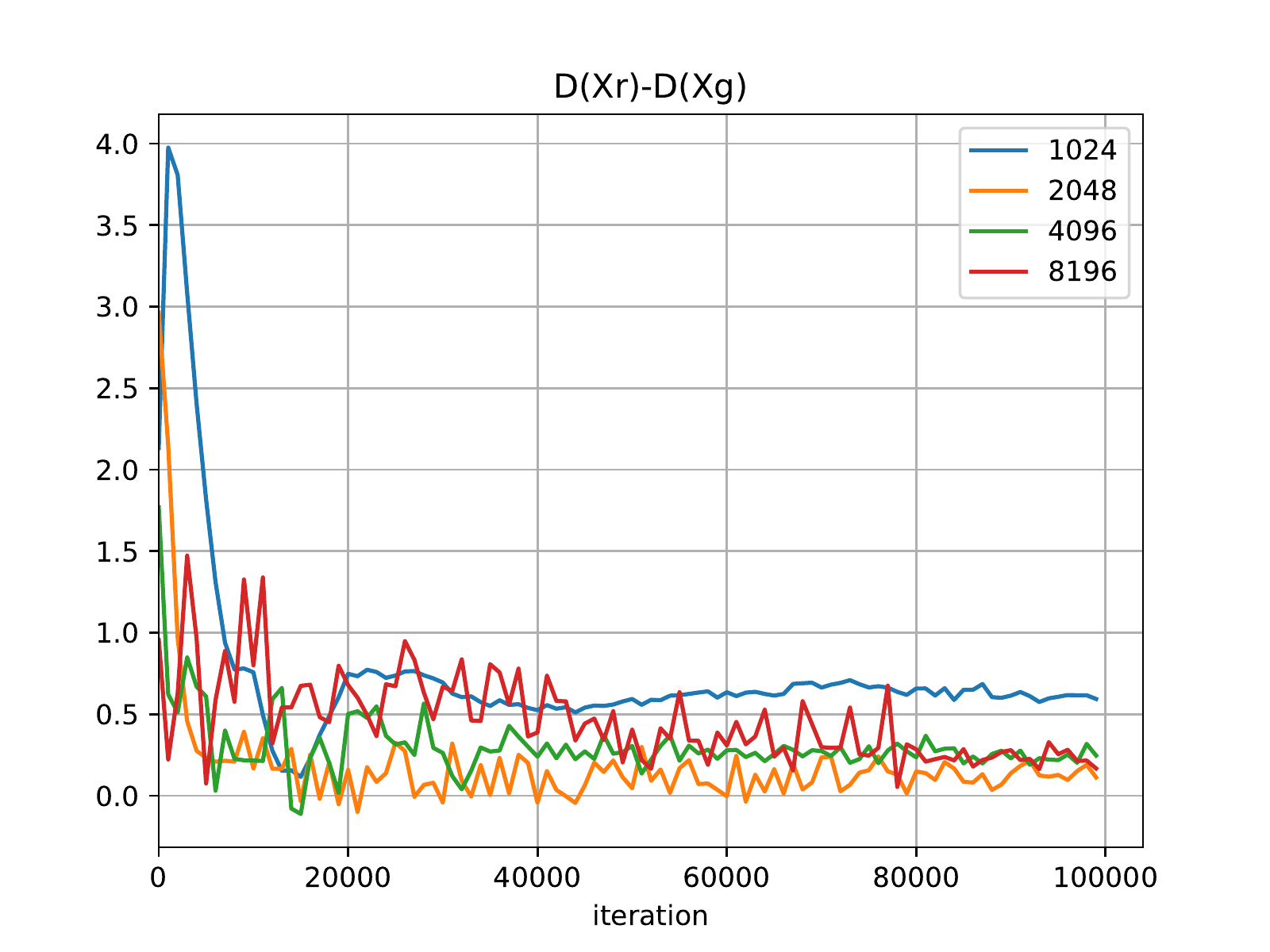}}\\
		\subfloat[Judge accuracy]{\includegraphics[width=0.45\linewidth] {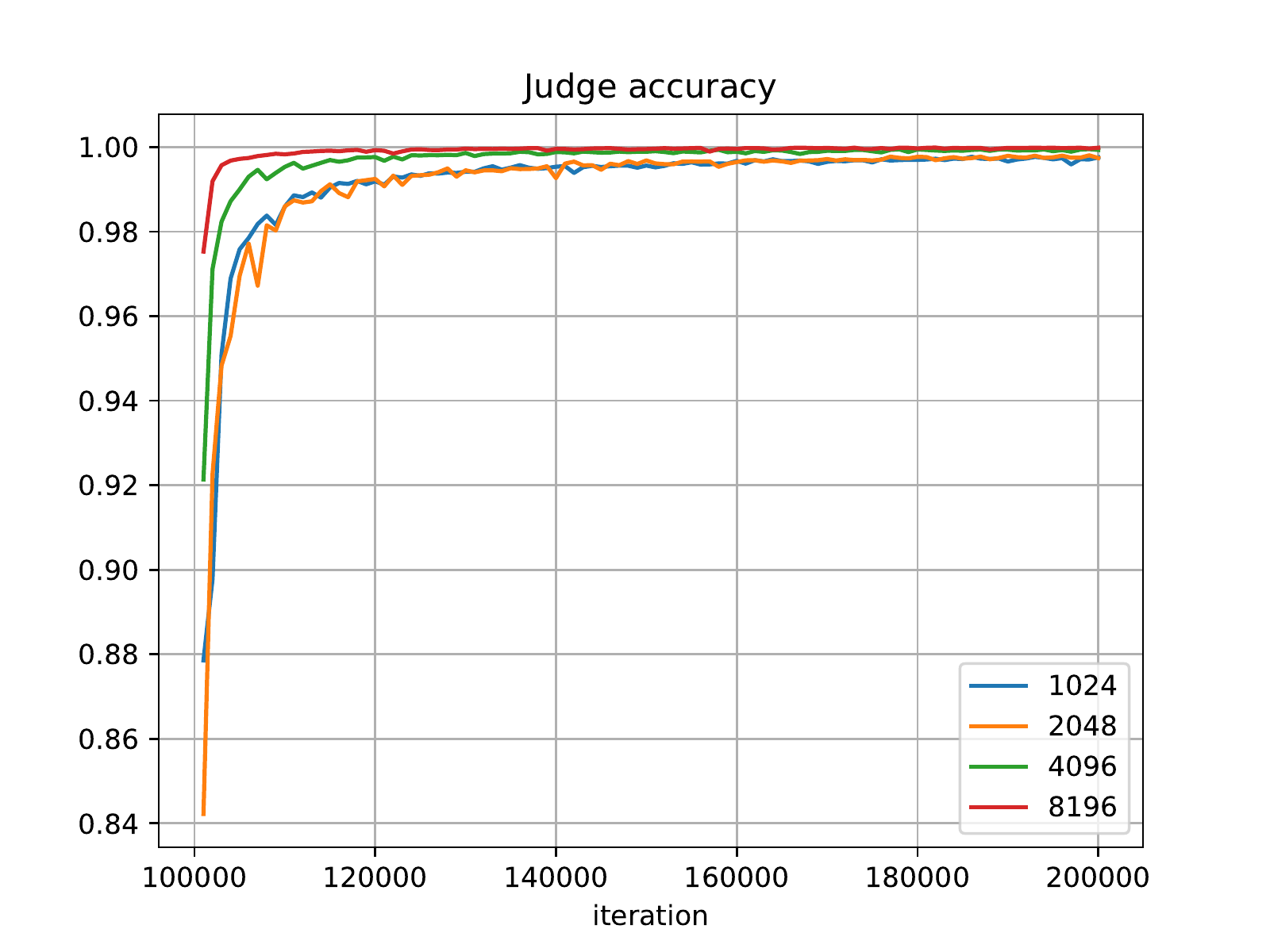}
		}
		\subfloat[Wasserstein distance]{\includegraphics[width=0.45\linewidth]{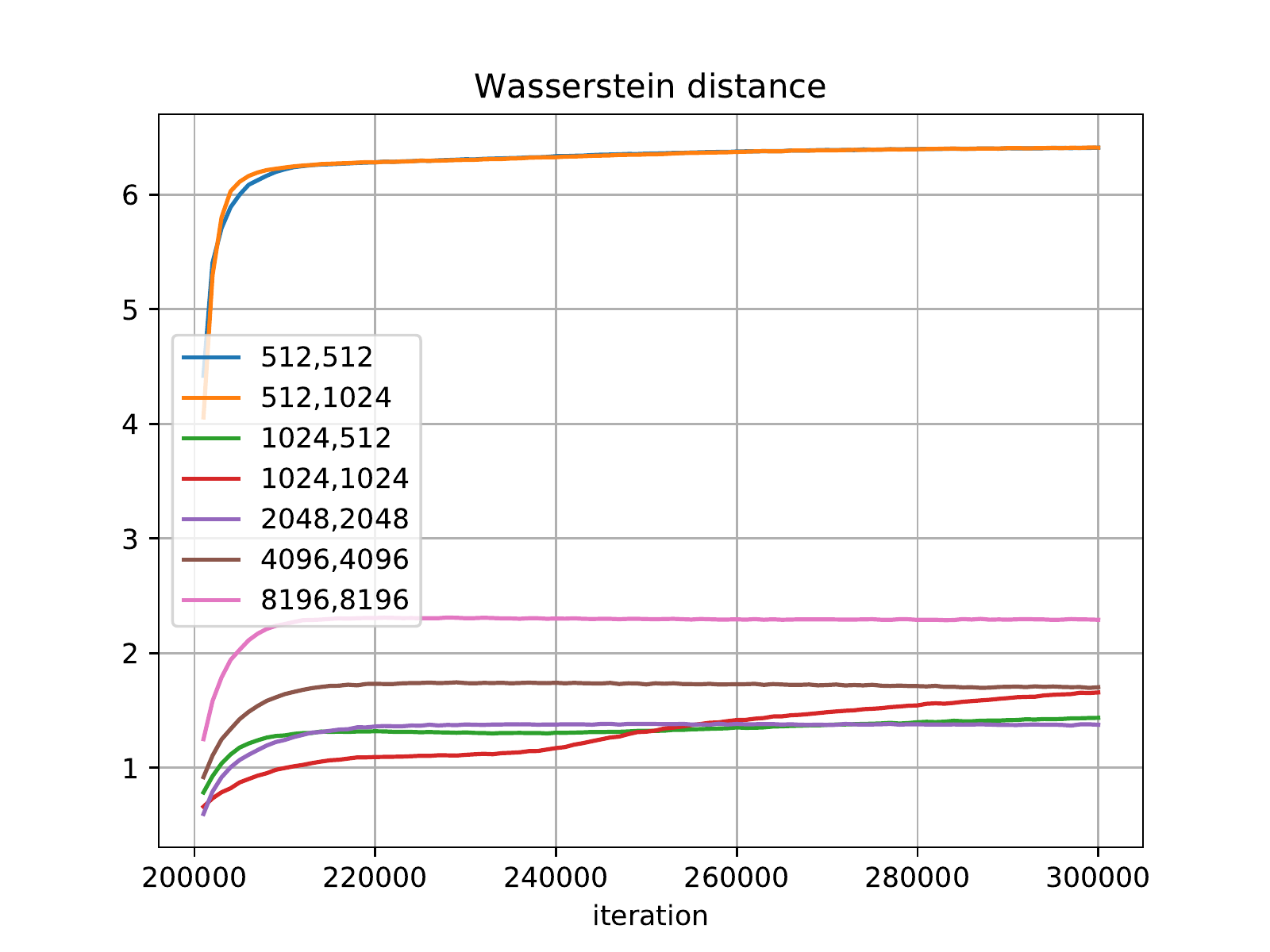}}
	\end{center}
	\caption{Quantitative results of varying the width of the networks. The numbers in the legends indicate the numbers of neurons in each hidden layer of and the discriminator. For all the metrics, lower is better.}
	\label{width_quant}
\end{figure}

In Figure \ref{depth_quant} and \ref{width_quant}, we show the quantitative results of varying the depth or width of the networks. In these experiments, we use 1 generator and 1 discriminator for generating 3 Gaussians. We do not see any significant improvement when increasing the complexity of the networks compared to increasing the number of generators and discriminators. Increasing the depth does not help might be because the dataset is too simple and more layers do not give better modeling power. Also, training deep MLPs can be unstable. Increasing the width does not help might be because hidden dimensions of 1024 can preserve information about the input and do not need to be larger.

\subsubsection{Generation of datasets defined by neural networks}
\begin{figure}[!h]
	\begin{center}
		\subfloat[Fréchet distance]{\includegraphics[width=0.45\linewidth]{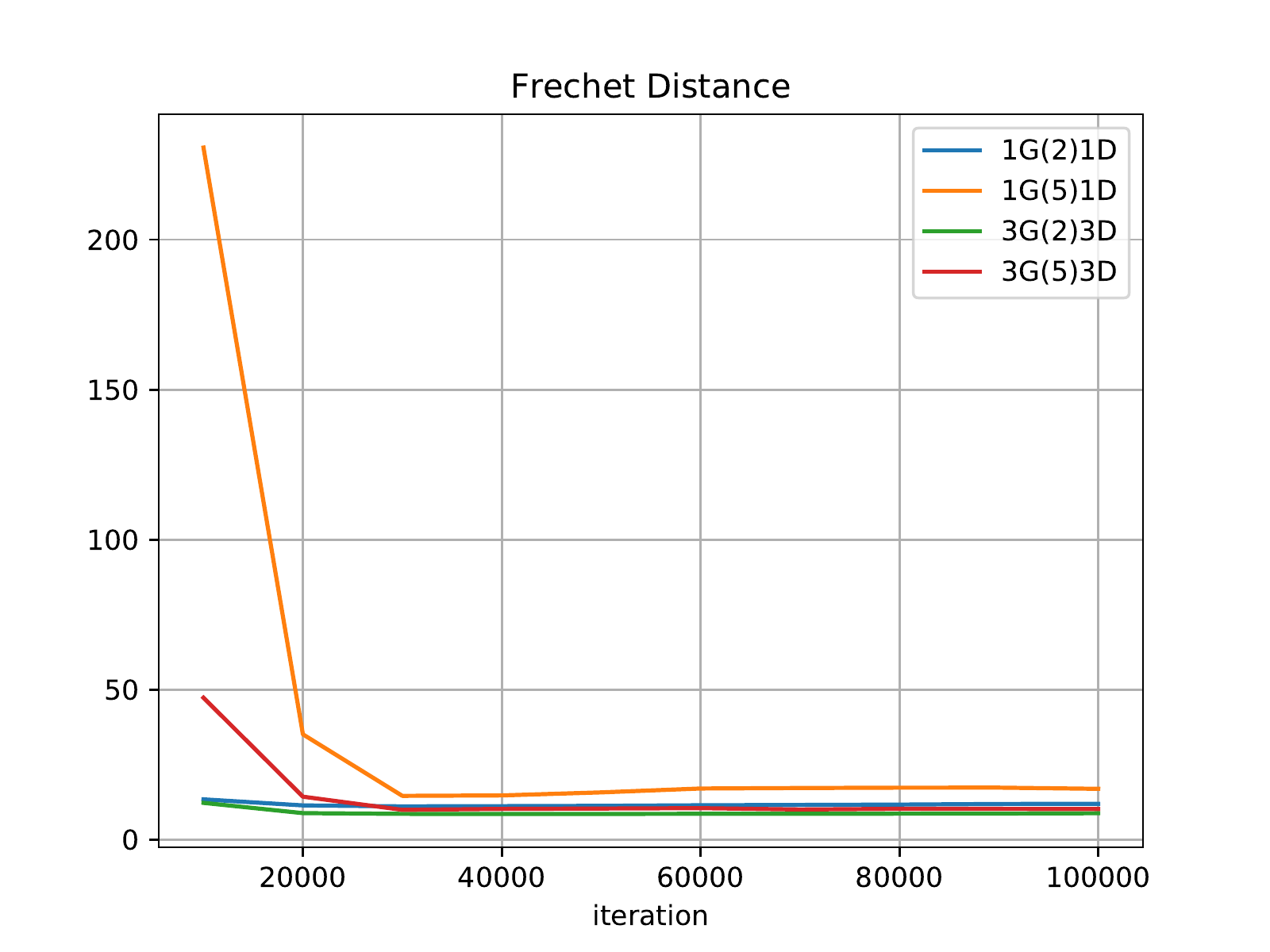}}
		\subfloat[$D(x_r)-D(x_g)$]{\includegraphics[width=0.45\linewidth]{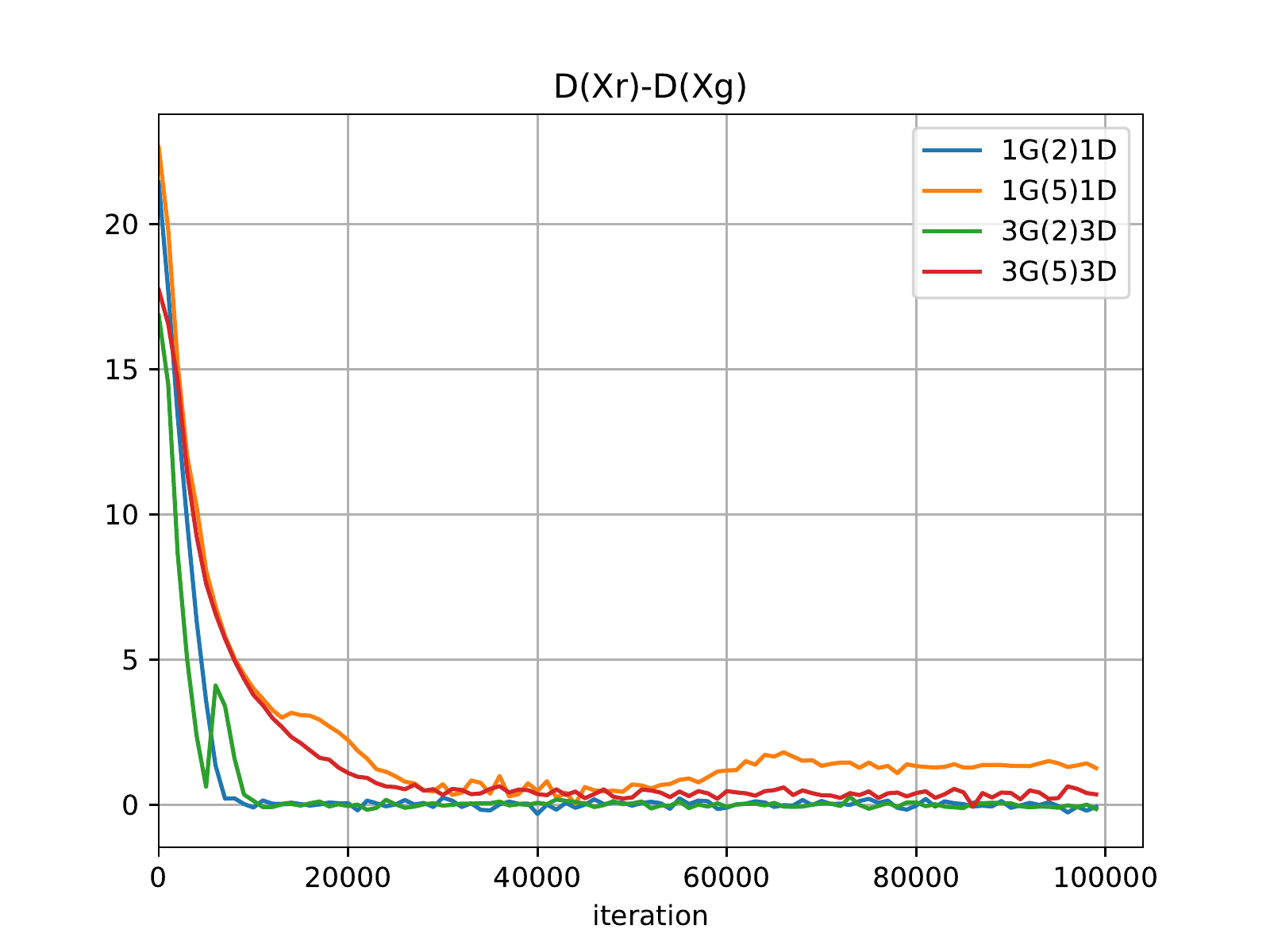}}\\
		\subfloat[Judge accuracy]{\includegraphics[width=0.45\linewidth] {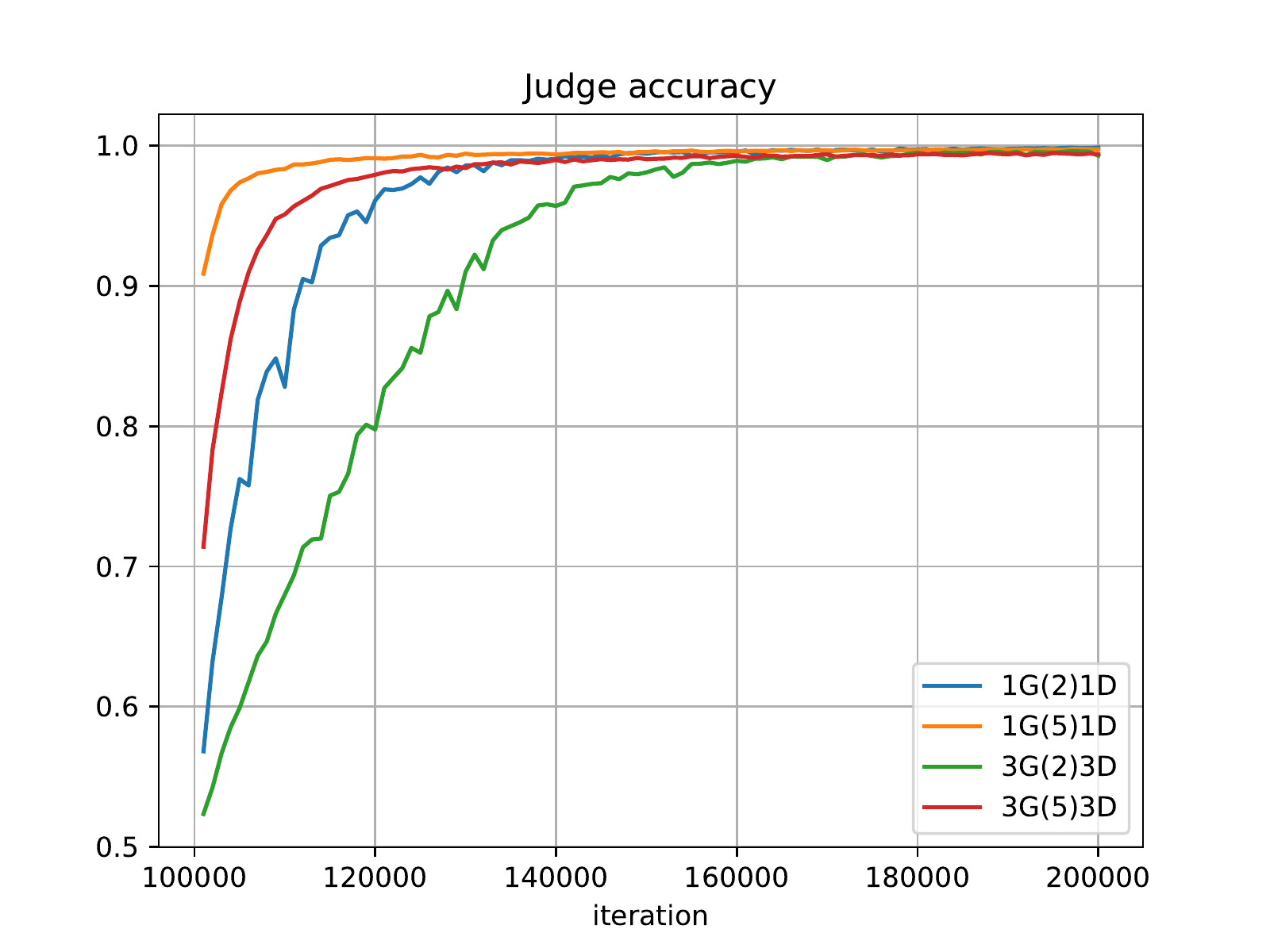}
		}
		\subfloat[Wasserstein distance]{\includegraphics[width=0.45\linewidth]{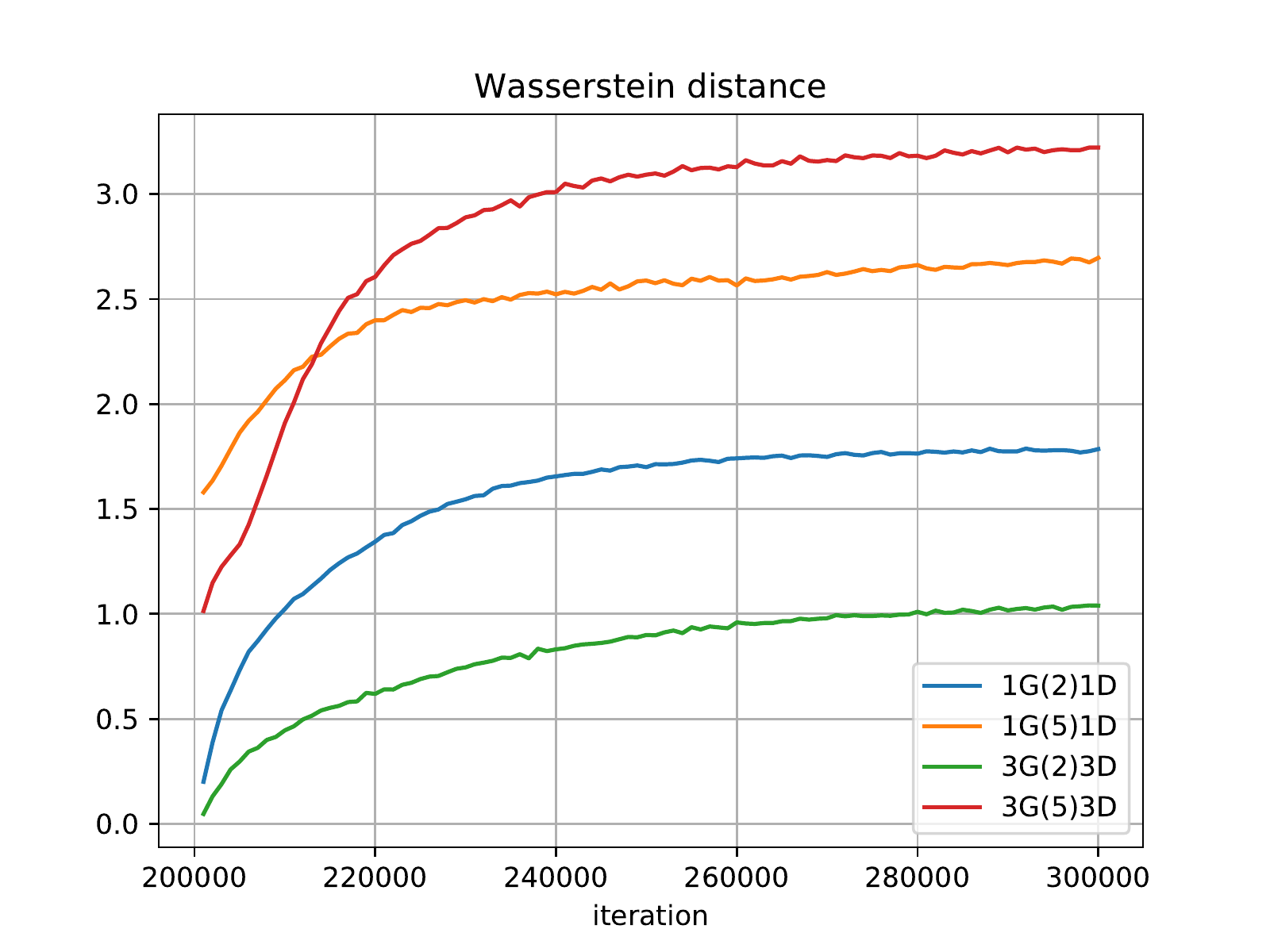}}
	\end{center}
	\caption{Results of the dataset generated by a network $R$. $nG(l)mD$ indicates that there are $n$ generators of $l$ layers and $m$ discriminators (of 5 layers by default). For all the metrics, lower is better.}
	\label{net_comp_quant}
\end{figure}

In this part, we define the real data distribution as the distribution of the output of a neural network $R$ that has the same input and architecture as the generator(s). The parameters of $R$ is randomly initialized with the Glorot uniform initializer \cite{glorot} and fixed thereafter. There are also at least two ways in which the generator(s) can win the game: either memorize a large sample of the training data according to \cite{generalization}, or learn to have the same parameters as $R$ (of course, there are other sets of parameters that enables the generator(s) to generate $\mathbb{P}_r$ due to the symmetry and complexity of neural networks). We consider the simplest situation where $R$ has only two layers, that is, it defines an affine transformation from $\mathbb{R}^{1024}$ to $\mathbb{R}^{1024}$. Therefore, this dataset is in fact a 1024-dimension Gaussian distribution with randomly initialized mean and covariance. We plot the quantitative results in Figure \ref{net_comp_quant}.
The results show that GAN training can have difficult in learning an affine transformation.

\subsubsection{Varying the training set size}
\begin{figure}[!h]
	\begin{center}
		\subfloat[$J_{acc}$ (training set)]{\includegraphics[width=0.45\linewidth] {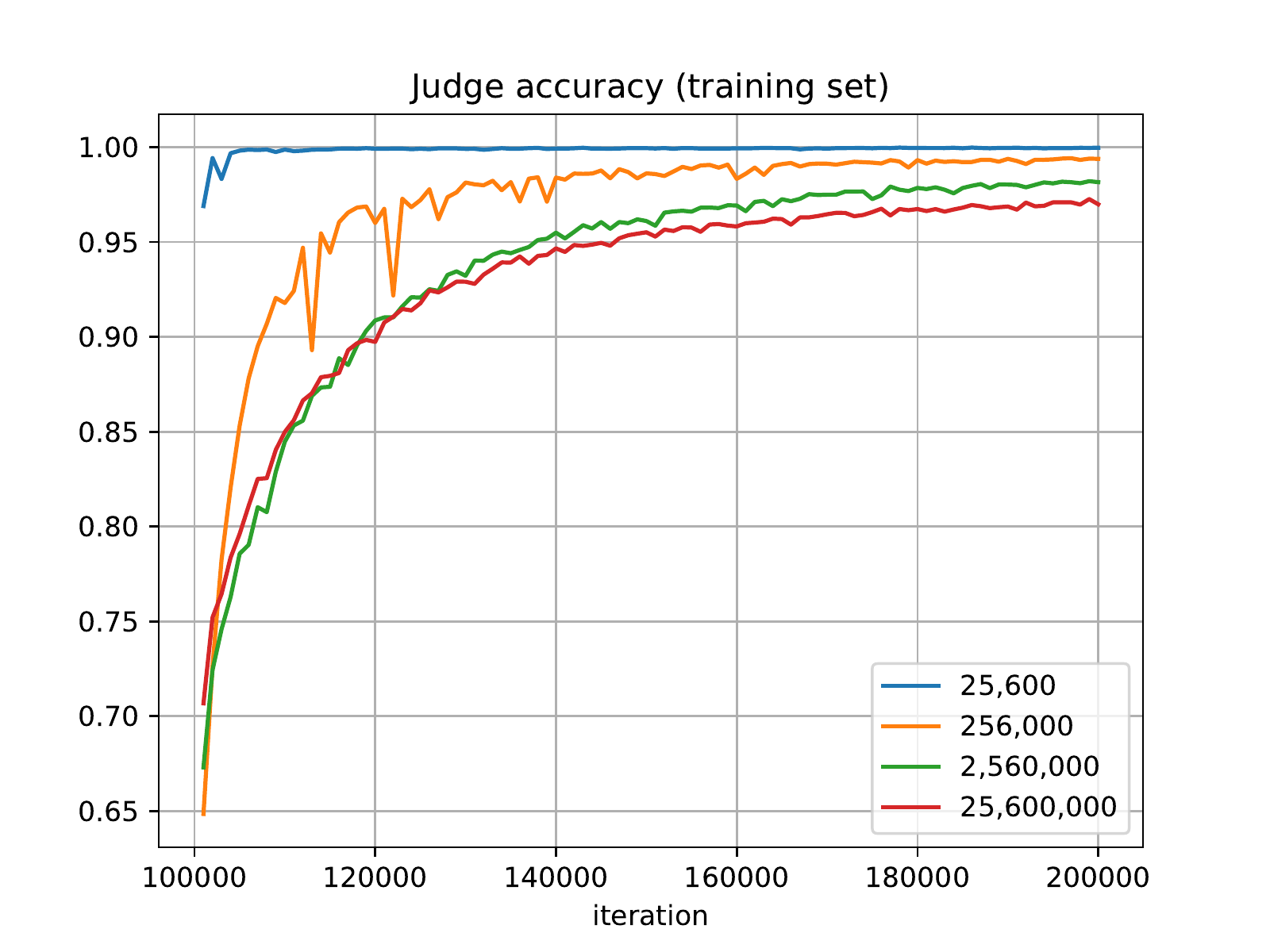}	}
		\subfloat[$J_{acc}$ (test set)]{\includegraphics[width=0.45\linewidth] {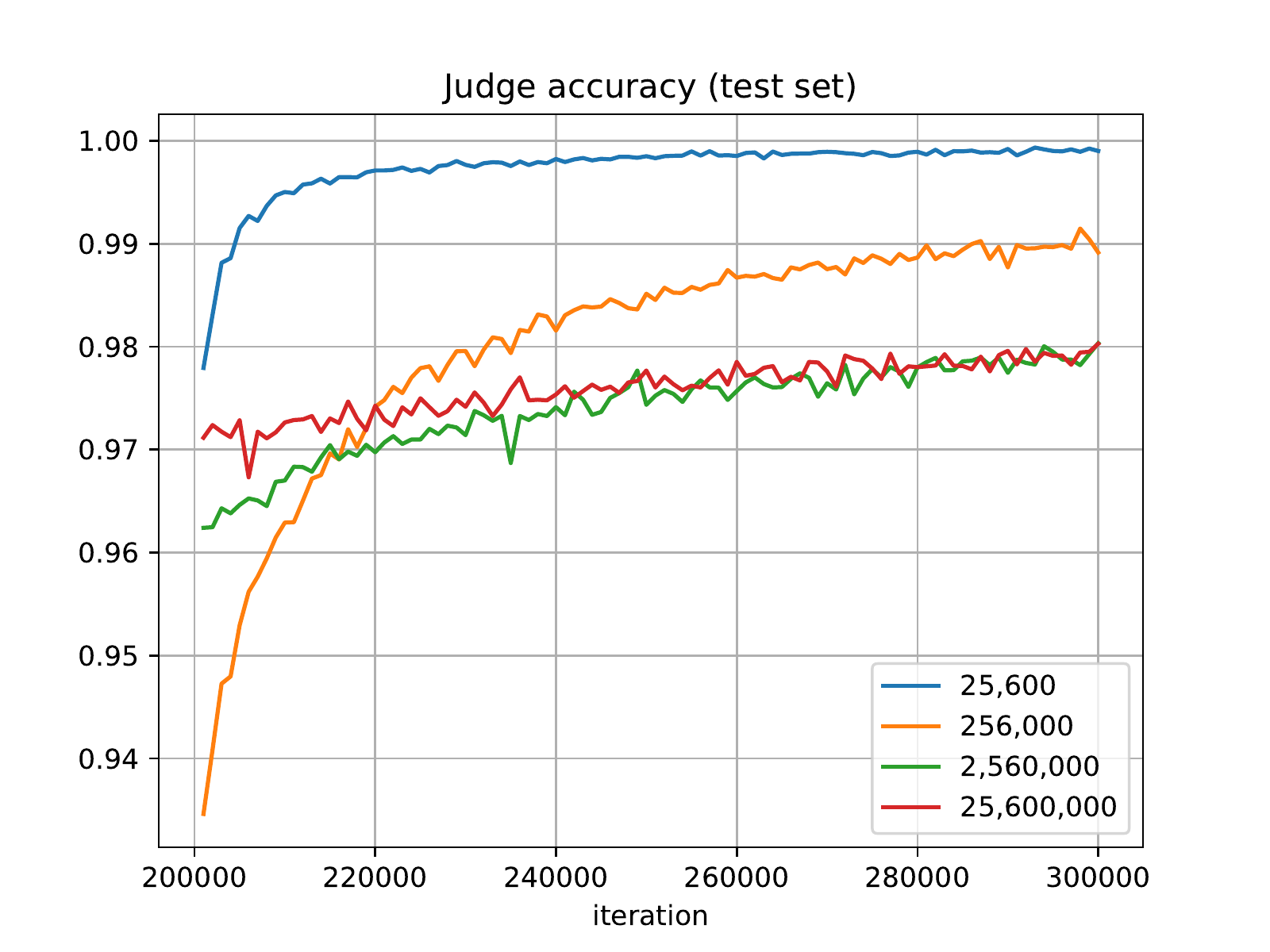}}\\
		\subfloat[W-dist (training set)]{\includegraphics[width=0.45\linewidth]{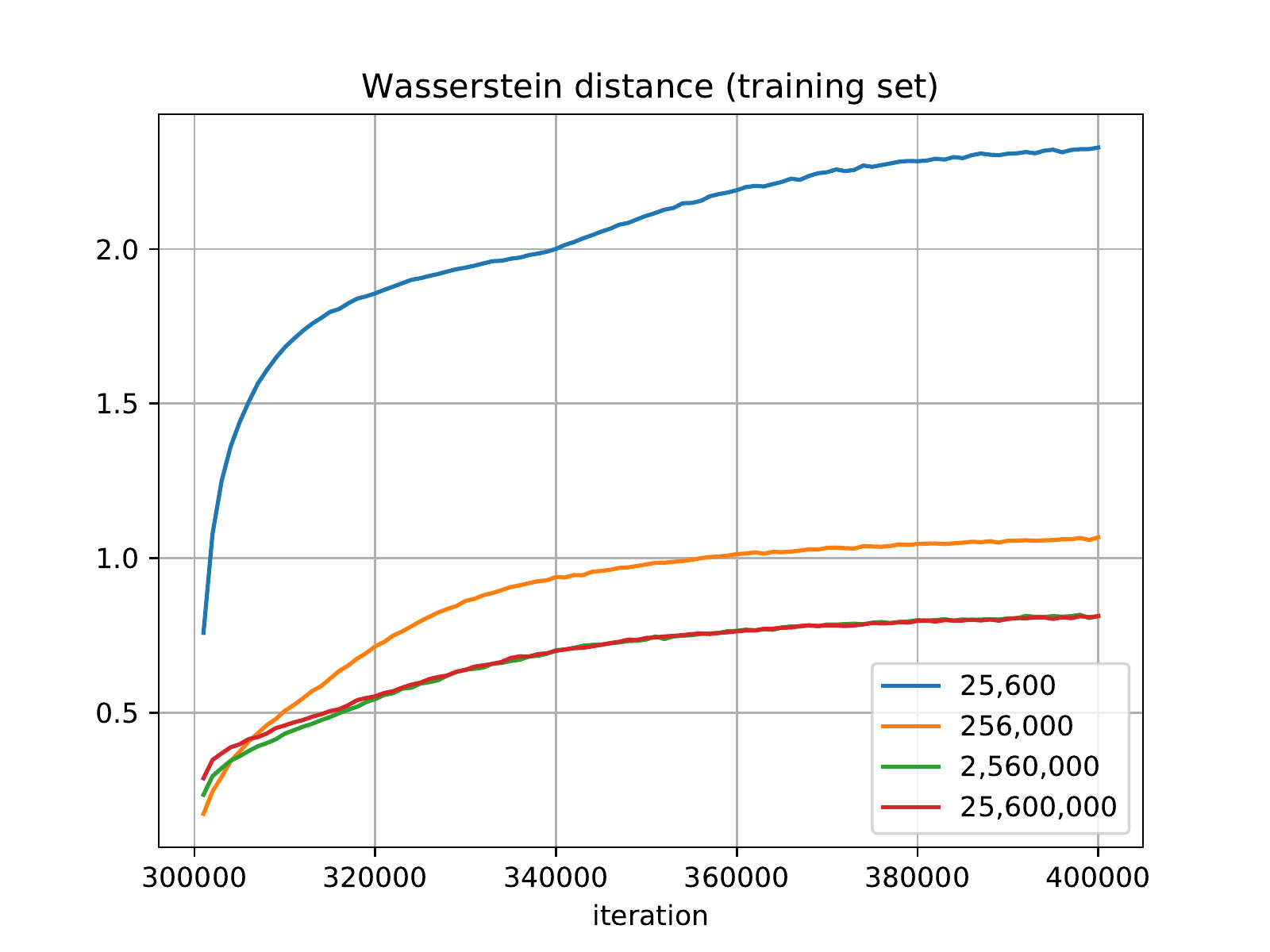}}
		\subfloat[W-dist (test set)]{\includegraphics[width=0.45\linewidth]{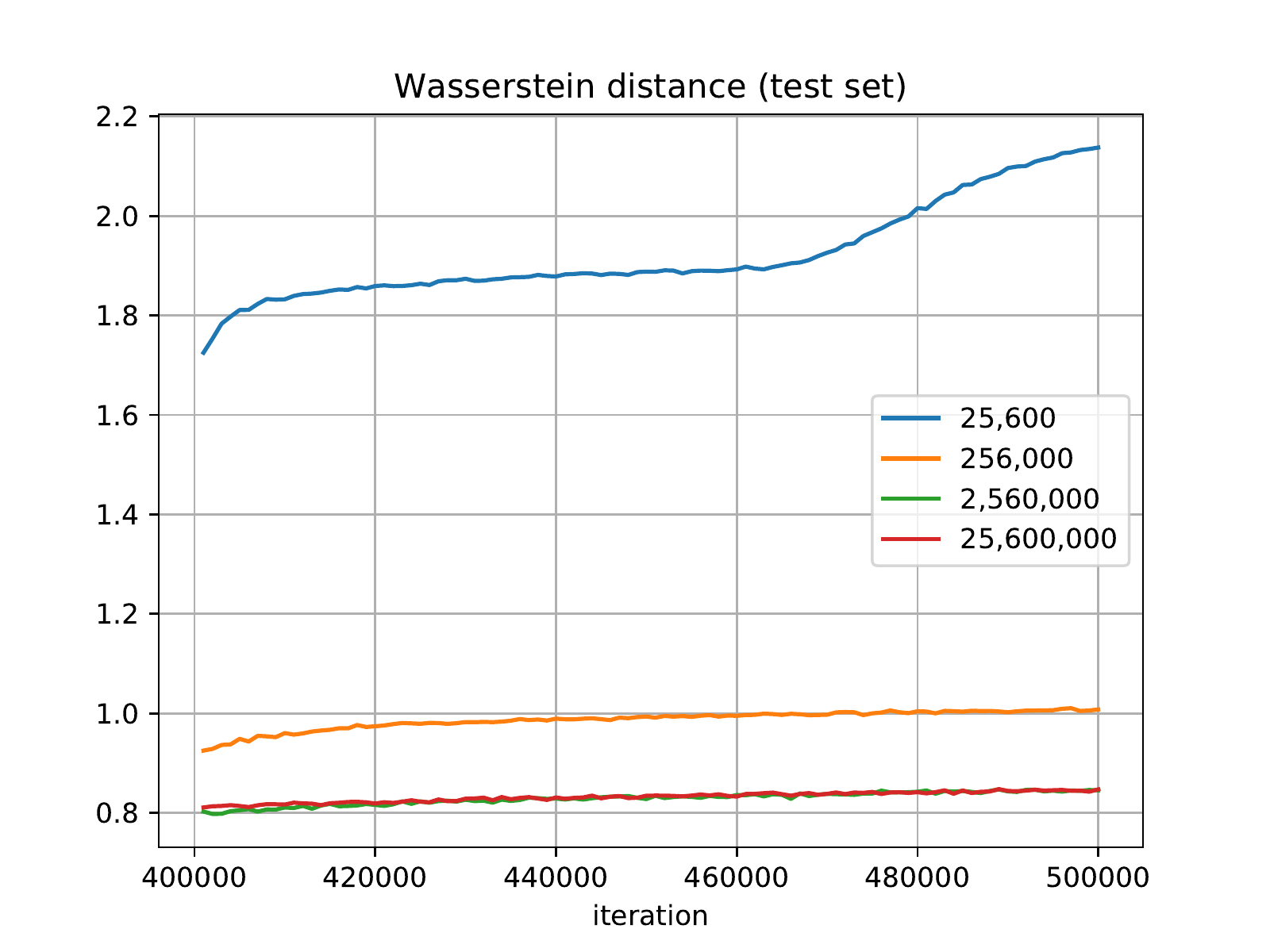}}\\
		\subfloat[$D(x_r)-D(x_g)$]{\includegraphics[width=0.45\linewidth]{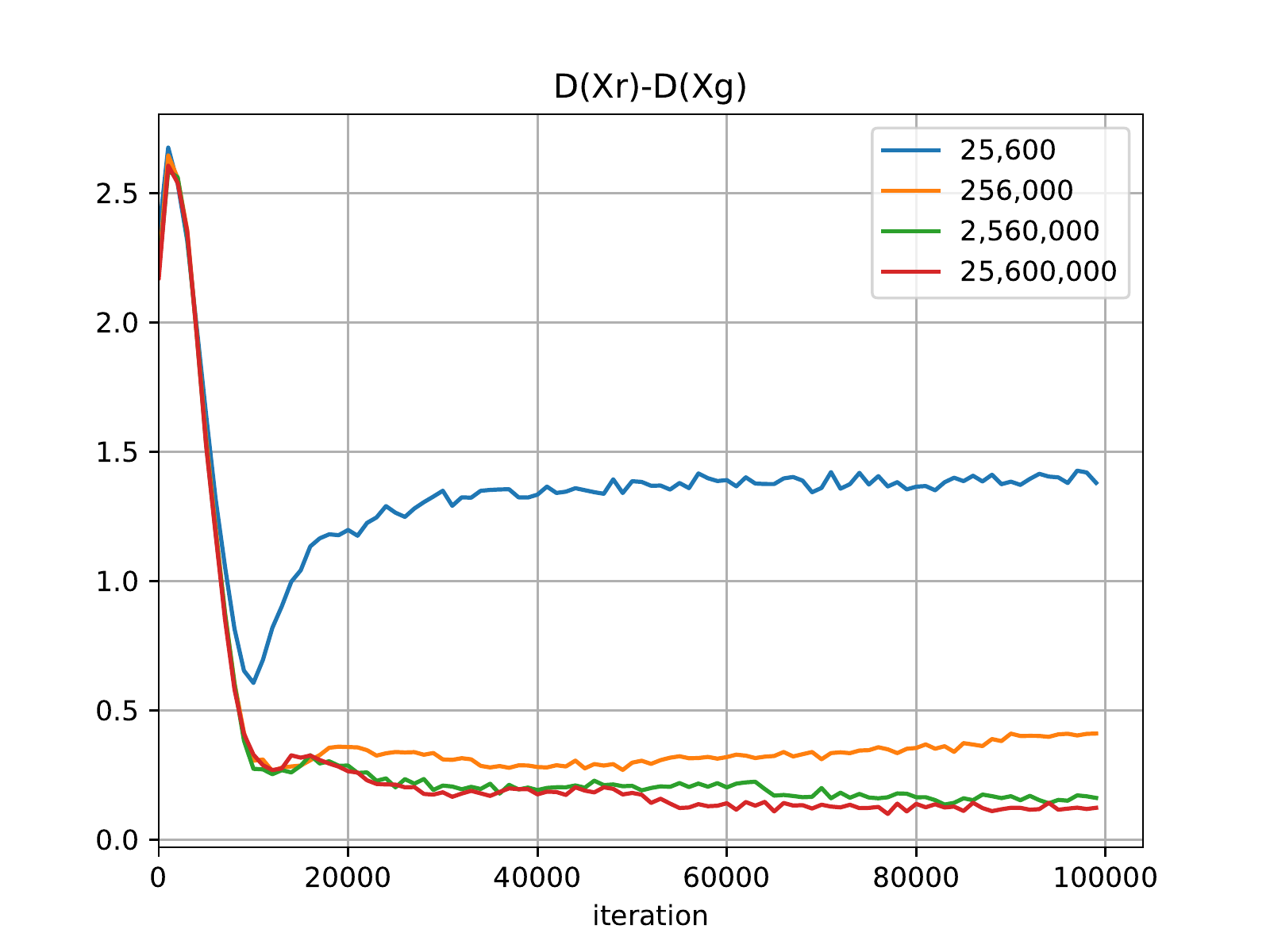}}\\
	\end{center}
	\caption{Results on the 3 Gaussians dataset when varying the training set size. The numbers in the legends indicate the training set sizes. For all the metrics, lower is better.}
	\label{setsize_comp_quant}
\end{figure}

Now that we have access to infinite training data, we are able to study the influence the training set size has on the quantitative metrics and show the results in Figure \ref{setsize_comp_quant}. In this set of experiments, we have a MIX+GAN consisting of 3 generators and 3 discriminators, each of which has 5 layers and 1024 neurons in every hidden layer. There are infinite samples in the test set. The only factor of variation is the training set size. The results show that GANs perform worse with smaller training sets. On the contrary, the GAN trained on the largest training set preforms among the best in terms of all the metrics. We can see that the distances to the training set is larger with smaller training set size. This phenomenon is not straightforward as some would believe that it is easier for the generator(s) to overfit smaller training sets. A possible explanation is that with fewer training data, the discriminator(s) can memorized the training set and reject fake samples more easily, providing less informative feedbacks to the generator. This explanation is consistent with the one in Session 4.2 of \cite{biggan}.

Note that the dimension of our data is $1024=32\times32$, which is the same as the spatial dimension of the CIFAR-10 dataset \cite{CIFAR10}. However, the CIFAR-10 dataset is more complex and consists of only 50,000 training images. Therefore, one can expect a performance boost when there are more training data for the training of GANs on CIFAR-10 or other small-scale image datasets.

\section{extended experiments on CIFAR-10}
In this section, we will show that the lessons we learned from artificial datasets apply to realistic datasets. Inspired by our empirical finding on the artificial datasets that increasing the mixture size can improve the performance of GANs, we modify MIX+GAN and train mixtures of GANs on the CIFAR-10\cite{CIFAR10} dataset.
\begin{figure}[!h]
	\begin{center}
		\subfloat{\includegraphics[width=1\linewidth]{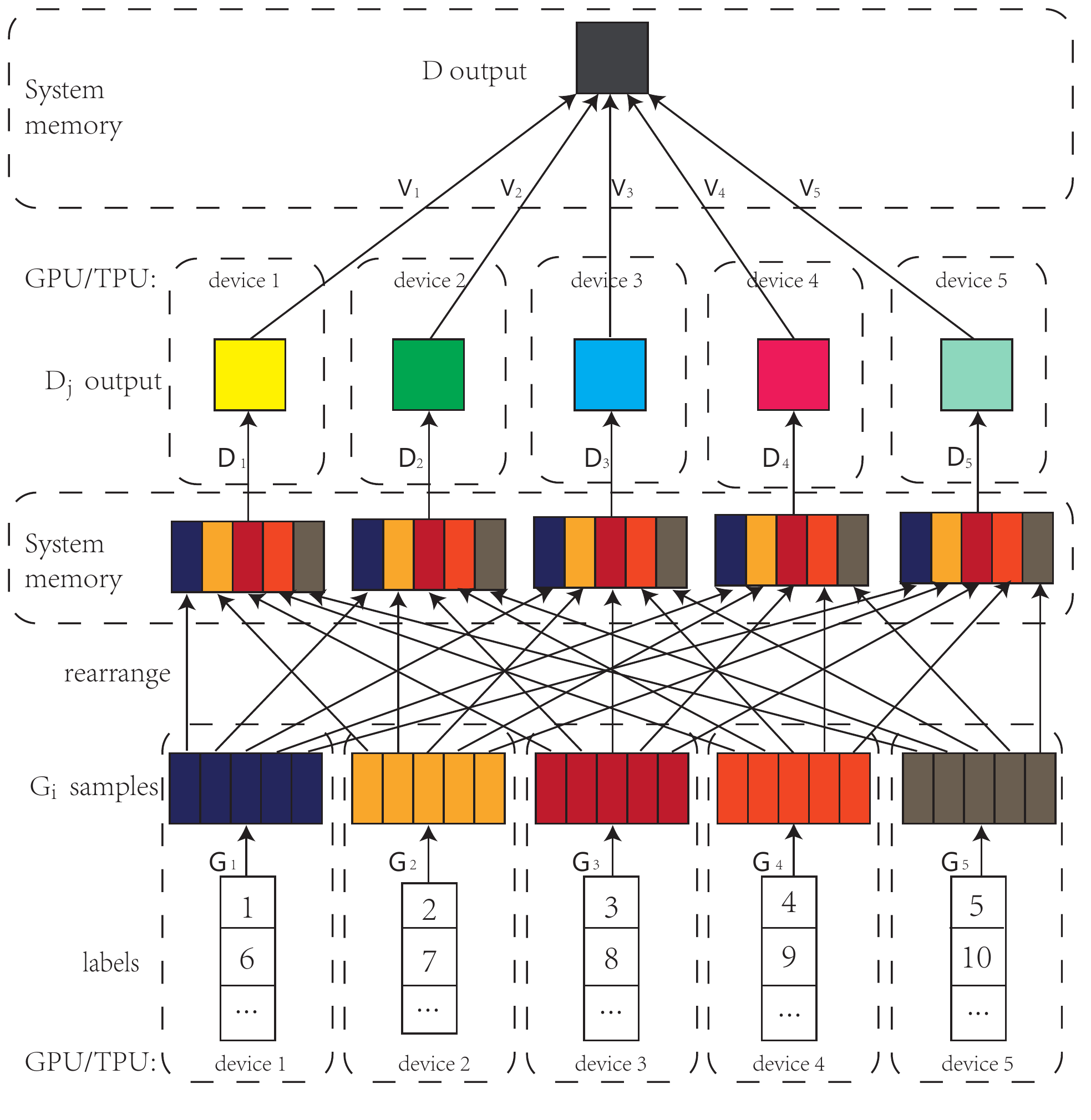}}
	\end{center}
	\caption{Illustration of the generation and discrimination of fake samples when there are 5 generators and 5 discriminators distributed across 5 devices. The input noise to each generator is omitted.}
	\label{distribute}
\end{figure}

Since the time and space complexity of MIX+GAN is $O(n_Gn_D)$, it is computationally infeasible to train very large mixtures of GANs. Thus, we propose to use a modified version of MIX+GAN. We assume that $w_i$ is uniformly distributed (which is true for the data distribution of CIFAR-10 and many other datasets). The batch generated by each $G_i$ is split into $n_D$ parts uniformly and fed to different discriminators. Therefore, the actual batch size for each generator and each discriminator remains unchanged, but each discriminator can receive samples from different generators. Inspired by the finding that different generators can capture different modes in a distribution, we do not make each generator generate samples for 10 classes, but $max\{10/n_G,1\}$ classes, which can ease the difficulty of generation for each generator. In this way, the generators can be viewed as a \textit{mixture of experts}\cite{ME} and the discriminators can be viewed as an ensemble of discriminative models. We use a model-parallelism setting where generators and discriminators are distributed across different devices. If we have $n$ GPU/TPU devices, then $G_i$ and $D_j$ are allocated to device $(i-1\bmod n)+1$ and device $(j-1 \bmod n)+1$ respectively. In this way, there is no need to synchronize parameters across different devices and load balance can be achieved if both $n_G$ and $n_D$ are divisible by $n$. Figure \ref{distribute} illustrates the flow of the generation and discrimination of fake samples when there are 5 generators and 5 discriminators distributed across 5 devices. 

For CIFAR-10, We use MHingeGAN\cite{MHGAN} as the base model. MHingeGAN is based on BigGAN\cite{biggan} but uses multi-class hinge losses. In a MHingeGAN, $D$ is a $(K+1)$-class classifier where class $0$ represents fake data and class 1 to class $K$ represent the $K$ classes in the dataset. The intuition behind the multi-class hinge loss is to make the affinity of $D$ for the target class to be by a margin of at least 1 over the other classes. The loss for $D$ is
\begin{align}
&L_D=L_{real,MH}+L_{fake,MH}\\
&=\mathbb{E}_{(x,y)\sim \mathbb{P}_r}[\max(0,1-D_y(x)+D_{\neg y}(x))]\\
&+\mathbb{E}_{(x,y)\sim \mathbb{P}_g}[\max(0,1-D_0(x)+D_{\neg 0}(x))]
\label{d_mh_loss}
\end{align}
where $D_{y}(x)$ is the $y$-th element of the output vector $D(x)$ and represents $D$'s affinity for class $y$ given input $x$, $D_{\neg y}(x)$ is $D$'s highest affinity for any class that is not $y$, \ie $D_{\neg y}(x)=\max D_{k \neq y}(x), k=0,1,2,...,K$.

The loss for $G$ is a combination of the multi-class hinge loss
\begin{align}
L_{G,MH}=\mathbb{E}_{(x,y)\sim \mathbb{P}_g}[\max(0,1-D_y(x)+D_{\neg y}(x))]
\label{g_mh_loss}
\end{align}
and a feature matching loss
\begin{align}
L_{G,FM}\!=\!\|\mathbb{E}_{{x}\sim \mathbb{P}_g}[D_{feat}(x))]\!-\!\mathbb{E}_{{x}\sim \mathbb{P}_r}[D_{feat}(x))]\|_1
\label{fm_loss}
\end{align}
where $D_{feat}(x)$ is the feature of $x$ after the last pooling layer of $D$.
Different from \cite{MHGAN}, the loss we use for $G$ is 
\begin{align}
L_G=L_{G,MH}+\lambda L_{G,FM}
\end{align}
where $\lambda=0.05$.

Our network architectures are the same as \cite{MHGAN}. We use shared embedding, hierarchical input noise, and moving average of the weights for $G$ as in BigGAN\cite{biggan}. The dimension of the input noise $z$ is 80. The batch size for each generator and each discriminator is 50. We use the Adam optimizer\cite{adam} with $\beta_1=0$, $\beta_2=0.9$ and a learning rate of $0.0002$ for all $G$s and $D$s. The proposed models are trained for $100,000$ iterations. There are 4 discriminator updates and 1 generator update per iteration. The training of a mixture of 10 generators and 10 discriminators takes 1.5 days with 5 Nvidia GTX 1080Ti GPUs and 1 day with a TPU-V3. We show the supervised and unsupervised Inception Score and FID in Table \ref{supervised_table} and Table \ref{unsupervised_table} respectively.  We refer to our method as "MIX-GAN", to distinguish from MIX+GAN. Note that the "GAN" can be substituted by the name of a specific GAN model. We evaluate the Inception Score and the FID with 50,000 samples from each distribution. Since the test set of CIFAR-10 has only 10,000 samples, it is repeated 5 times (which does not change the moments used for calculating FID). Using 10 generators and 10 discriminators, we improve the state-of-the-art IS and FID on CIFAR-10 significantly. 

\begin{table}[!h]
	\centering
	\begin{tabular}{l|r|r|r}
		$\!\!\!$Method&IS&$\!\!\!$FID(train)$\!\!\!$&$\!\!\!$FID(test)$\!\!\!$\\\hline
		$\!\!\!$ACGAN\cite{ACGAN}&8.25$\pm$0.07&-&-\\
		$\!\!\!$SGAN \cite{sgan}&8.59$\pm$0.12&-&-\\
		$\!\!\!$Splitting GAN\cite{class-splitting-gan}&8.87$\pm$0.09&-&-\\
		$\!\!\!$WGAN-GP\cite{improvedwgan}&8.42$\pm$0.10&-&-\\
		$\!\!\!$cGANs with Projection D \cite{cgans}&8.62&17.5&-\\
		$\!\!\!$CT-GAN\cite{improving}&8.81$\pm$0.13&-&-\\
		$\!\!\!$BigGAN\cite{biggan}&9.22&14.73&-\\	
		$\!\!\!$CR-BigGAN\cite{cr-biggan}&-&11.67&-\\	
		$\!\!\!$MHingeGAN\cite{MHGAN}&9.58$\pm$0.09& 7.50&-\\
		$\!\!\!$MIX-MHingeGAN, 1G1D (ours)&9.61$\pm$0.08&4.57 & 6.66\\
		$\!\!\!$MIX-MHingeGAN, 2G2D (ours)&9.83$\pm$0.12 &4.46  &6.23 \\
		$\!\!\!$MIX-MHingeGAN, 5G5D (ours)&9.98$\pm$0.09&3.95&5.78\\
		$\!\!\!$MIX-MHingeGAN,10G10D(ours)$\!\!\!$&$\!\!\!$\bf10.21$\pm$0.14&\bf3.60&\bf5.52\\
	\end{tabular}
	\caption{\label{supervised_table}Supervised Inception Scores and FIDs from $\mathbb{P}_g$ to the empirical distributions of the training set and the test set of CIFAR-10}
\end{table}

\begin{figure}[!h]
	\begin{center}
		\subfloat{\includegraphics[width=0.95\linewidth]{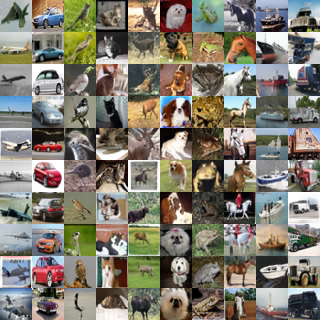}}
	\end{center}
	\caption{CIFAR-10 samples generated by our supervised model with 10 Gs and 10 Ds. Samples in different columns are generated by different generators.}
	\label{supervised}
\end{figure}

A random sample of a supervised MIX-MHingeGAN with 10 generators and 10 discriminators is shown in Figure \ref{supervised}.

Since the multi-class hinge loss is only applicable for conditional image generation, we use BigGAN\cite{biggan} as the base model for unconditional image generation. We find that the performance of the unconditional MIX-BigGAN degrades after some iterations and thus report the IS and FID at iteration $60,000$ before it degrades.
\begin{table}[!h]
	\centering
	\begin{tabular}{l|r|r|r}
		Method &IS&$\!\!\!$FID(train)$\!\!\!$&$\!\!\!$FID(test)$\!\!\!$\\\hline
		PGGAN\cite{PGGAN}&8.80$\pm$0.05&-&-\\
		SN-GAN\cite{spectralnorm}&8.22$\pm$0.05&21.7&-\\
		AutoGAN\cite{AutoGAN}&8.55&12.42& -\\
		NCSN\cite{NCSN}&8.87$\pm$0.12&25.32&-\\
		CR-GAN\cite{cr-biggan}&8.40&14.56&-\\	
		MIX-BigGAN,10G10D(ours)&\bf9.67$\pm$0.08&\bf8.17&\bf10.26\\
	\end{tabular}
	\caption{\label{unsupervised_table}Unsupervised Inception Scores and FIDs from $\mathbb{P}_g$ to the empirical distributions of the training set and the test set of CIFAR-10}
\end{table}

\begin{figure}[!h]
	\begin{center}
		\subfloat{\includegraphics[width=0.95\linewidth]{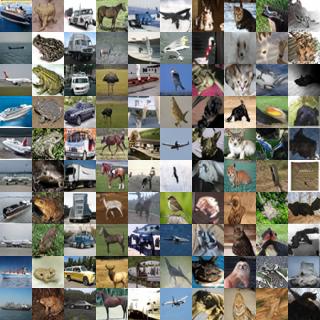}}
	\end{center}
	\caption{CIFAR-10 samples generated by our unsupervised model with 10 $G$s and 10 $D$s. Samples in different columns are generated by different generators.}
	\label{unsupervised}
\end{figure}

We show samples generated by an unsupervised MIX-BigGAN in Figure \ref{unsupervised}. To some extent, the 10 generators can learn different concepts automatically without label supervision, although they do not correspond to the 10 classes perfectly.
